# A General Framework for Robust Testing and Confidence Regions in High-Dimensional Quantile Regression

Tianqi Zhao* Mladen Kolar† Han Liu*

March 16, 2015

**Abstract**

We propose a robust inferential procedure for assessing uncertainties of parameter estimation in high-dimensional linear models, where the dimension $p$ can grow exponentially fast with the sample size $n$. Our method combines the de-biasing technique with the composite quantile function to construct an estimator that is asymptotically normal. Hence it can be used to construct valid confidence intervals and conduct hypothesis tests. Our estimator is robust and does not require the existence of first or second moment of the noise distribution. It also preserves efficiency in the sense that the worst case efficiency loss is less than 30% compared to the square-loss-based de-biased Lasso estimator. In many cases our estimator is close to or better than the latter, especially when the noise is heavy-tailed. Our de-biasing procedure does not require solving the $L_1$-penalized composite quantile regression. Instead, it allows for any first-stage estimator with desired convergence rate and empirical sparsity. The paper also provides new proof techniques for developing theoretical guarantees of inferential procedures with non-smooth loss functions. To establish the main results, we exploit the local curvature of the conditional expectation of composite quantile loss and apply empirical process theories to control the difference between empirical quantities and their conditional expectations. Our results are established under weaker assumptions compared to existing work on inference for high-dimensional quantile regression. Furthermore, we consider a high-dimensional simultaneous test for the regression parameters by applying the Gaussian approximation and multiplier bootstrap theories. We also study distributed learning and exploit the divide-and-conquer estimator to reduce computation complexity when the sample size is massive. Finally, we provide empirical results to verify the obtained theory.

## 1 Introduction

Consider a high-dimensional linear regression model:

$$Y_i = \boldsymbol{X}_i^T \boldsymbol{\beta}^* + \epsilon_i, \qquad i = 1, \ldots, n, \tag{1.1}$$

where $\boldsymbol{X}_i \in \mathbb{R}^p$ are variables independently drawn from a distribution $\mathbb{P}_X$ with $\mathbb{E}[\boldsymbol{X}_i] = \mathbf{0}$, $\epsilon_i$ are i.i.d. random variables independent of $\boldsymbol{X}_i$, $\boldsymbol{\beta}^* \in \mathbb{R}^p$ is the unknown parameter of interest and $Y_i \in \mathbb{R}$ are the responses. We consider the setting where the dimension $p$ can be potentially much larger than the sample size $n$, but the sparsity level $s = \text{card}\{i : \beta_i^* \neq 0\}$ is smaller than $n$.

*Department of Operations Research and Financial Engineering, Princeton University, Princeton, NJ 08544, USA; Email: {tianqi, hanliu}@princeton.edu

†Booth School of Business, The University of Chicago, Chicago, IL 60637, USA; Email: mkolar@chicagobooth.edu

The model (1.1) has been intensively studied. One of the most successful estimators is the Lasso, which minimizes an $L_1$-penalized square loss function (Tibshirani, 1996). Theoretical properties of the Lasso estimator, including the minimax optimal convergence rate and model selection consistency, are well understood under suitable noise and design conditions (Candés and Tao, 2005; Bickel et al., 2009; Bühlmann and van de Geer, 2011; Negahban et al., 2012). Though significant progress has been made in estimation theories of high-dimensional statistics, quantifying model uncertainties remained untouched until recently. A series of work (Zhang and Zhang, 2013; van de Geer et al., 2014; Javanmard and Montanari, 2014) proposed a method of de-biasing the Lasso estimator and constructing confidence intervals and testing hypotheses for low-dimensional coordinates in the high-dimensional linear model (1.1). It was shown that the de-biasing procedure is uniformly valid under the Gaussian noise assumption, and Javanmard and Montanari (2014) further considered an extension to sub-Gaussian settings.

When heavy-tailed noise is present, the above estimation and inference procedures based on the square loss do not work. Quantile regression is often considered as a robust alternative (Koenker, 2005). In high-dimensional settings, Belloni and Chernozhukov (2011), Wang (2013), and Wang et al. (2012) established point estimation theories for the penalized least absolute deviation and quantile regressions. Belloni et al. (2013a) and Belloni et al. (2013b) further considered the inference task under the instrumental variable framework.

In this paper, we focus on developing valid inferential tools for high-dimensional linear models when the noise distribution is unknown and possibly heavy-tailed. Our method applies the de-biasing technique coupled with the composite quantile approach considered by Zou and Yuan (2008). See also Bickel (1973), Koenker (1984), and Bradic et al. (2011) who studied weighted composite losses. The procedure is outlined as follows: we take *any* first-stage estimator that satisfies a required convergence rate and empirical sparsity conditions, and de-bias it using a one-step update by subtracting a term associated with the sub-gradient of the composite quantile loss function. Our procedure enjoys four distinctive advantages. First, the method is robust under heavy-tailed noise distributions, and is valid even when the first two moments of the noise distribution do not exist. Second, the relative efficiency of our method to the de-biased Lasso is at least 70% in the worst case, and can be arbitrarily large in the best case. When the noise is Gaussian, the relative efficiency is approximately 95%. On the other hand, inference based on quantile regression, though robust, can have arbitrarily small relative efficiency compared to the least-square approach. Third, our procedure is quite general, as it only requires two inputs: any sparse estimate of $\boldsymbol{\beta}^*$ and any estimates of $K$ quantiles of the noise distribution, that satisfy the properties outlined in Theorem 3.2. Fourth, our assumption on the design is rather weak. In particular, we do not require the sparsity assumption on the inverse covariance matrix (also known as the precision matrix) of the design, which is imposed by most existing work on inference for high-dimensional models.

Our method inherits the nice properties of the composite quantile regression (CQR) in low dimensions. On the other hand, our framework does not require solving its high-dimensional counterpart – the penalized CQR, which may be computationally challenging. In fact, our method utilizes the composite quantile function only in the de-biasing step, not necessarily in obtaining the first-stage estimator. For initial estimation, any sparse estimator with the desired rate of convergence can be used. In addition to the penalized CQR, one can use the quantile (e.g., least absolute deviation) regressions, or even the Lasso if one believes that the noise satisfies the regularity conditions that guarantee the required estimation results. All first-stage estimators will result in the same asymptotic variance after de-biasing. We prove that each coordinate of our de-biased estimator is $\sqrt{n}$-consistent, and weakly converges to a normal distribution after scaled by $\sqrt{n}$. We can then construct asymptotic confidence intervals and test hypotheses on the low-dimensional coordinates of $\boldsymbol{\beta}^*$. Moreover, we consider the simultaneous inference for the high-dimensional global parameter $\boldsymbol{\beta}^*_{\mathcal{G}}$ for $\mathcal{G} \subseteq \{1, 2, \ldots, p\}$, whose cardinality can be as large as $p$. To this end, we employ the high-dimensional Gaussian approximation theory introduced by Chernozhukov et al. (2013).

As a special case, we study estimation theories of the $L_1$-penalized CQR and show that it converges at the minimax optimal rate of $\sqrt{s(\log p)/n}$ (Raskutti et al., 2012), and the empirical sparsity is of the same order as $s$. Therefore, it can be used as the first-stage estimator. This result is technically nontrivial and is of independent theoretical interest.

Lastly, we extend our result to a special regime where the sample size and the dimension are both large, so that it causes huge computational and storage burden to conduct inference based on the entire sample. We exploit the divide-and-conquer procedure which splits the sample and aggregates the estimators obtained on each sub-sample. In particular, we allow the number of splits to grow polynomially with the entire sample size. We prove that such a divide-and-conquer estimator achieves the "free-lunch" property: on one hand, it reduces computational complexity and requires minimal communications between sub-samples. On the other hand, the test based on the divide-and-conquer estimator preserves the asymptotic power of the "oracle" test, defined as the one based on the entire sample.

In addition to practical importance of our result, this paper develops novel proof techniques that are interesting in their own right. We develop new theories for de-biasing and inference when the loss function is not second-order differentiable and does not exhibit strong convexity. First, our theory is different from that of Zou and Yuan (2008), which mainly relies on the argmax continuous mapping theorem (van der Vaart and Wellner, 1996) to prove asymptotic normality of the CQR estimator in low dimensions. In contrast, our de-biased estimator is not defined as the minimizer of some loss function. Instead, we utilize the fact that the conditional expectation of the empirical composite quantile function has a *local* quadratic curvature (in a small neighborhood around the true parameter $\boldsymbol{\beta}^*$). The first-stage estimator is required to converge at a fast enough rate to ensure that it falls into the neighborhood of the true parameter. Such a local curvature allows us to handle the loss function that is not second-order differentiable. Second, in the proof we need to control the difference between the sub-differential of the empirical composite quantile loss function and its conditional expectation, which reduces to bounding the supremum of a non-smooth, non-Lipschitz empirical process. To this end we apply a theorem in Koltchinskii (2011) and Bouquet's concentration inequality that effectively utilize the variance information to obtain a high-probability bound of the superemum. Third, unlike the de-biased Lasso, where the one-step update is related to the sub-gradient of the penalty, we do not have such a KKT characterization for our de-biased composite quantile estimator, as we allow for any first-stage procedure. Our proof relies on a more sophisticated analysis to decompose the sub-gradient of the composite quantile loss into dominating and asymptotically ignorable terms, which is different from previous work on de-biased estimators.

**Other work on high-dimensional inference**. Besides the de-biasing framework mentioned above, other methods are studied in the literature of high-dimensional inference. Wasserman and Roeder (2009) and Meinshausen et al. (2009) proposed the sample-splitting approaches to obtaining valid p-values in high-dimensional problems. Belloni et al. (2012) and Belloni et al. (2013c) considered inference in linear and logistic regression models under the optimal instrument and double-post-selection framework. Within the class of non-convex penalties, Fan and Lv (2011) and Bradic et al. (2011) established asymptotic normality for regression parameters on the support based on oracle properties and minimal signal strength conditions. A separate line of research (Lockhart et al., 2014; G'Sell et al., 2013; Taylor et al., 2013, 2014) studied significance tests conditional on the selected model. Other related inference problems and methods include the $\ell_2$- and $\ell_\infty$-confidence set construction (Nickl and van de Geer, 2013; Juditsky et al., 2012), stability selection (Meinshausen and Bühlmann, 2010; Shah and Samworth, 2013) and worst-case procedure for post-selection inference (Berk et al., 2013).

**Organization of this paper**. In Section 2 we introduce the background on composite quantile function and describe our de-biasing procedure. Section 3 contains the main inference results for the de-biased composite quantile estimator. In particular, in §3.1 - 3.4 we show the asymptotic normality of the de-biased estimator, and its usage for conducting statistical inference such as constructing confidence intervals and testing hypotheses for the low-dimensional coordinates of the regression parameter $\boldsymbol{\beta}^*$. In §3.5 we consider the global simultaneous test for $\boldsymbol{\beta}^*$ using Gaussian approximation and multiplier bootstrap. Section 4 studies the first-stage estimators. We focus on the penalized composite quantile regression and its finite-sample properties, but also mention other estimators that satisfy the requirements for valid inference. In Section 5 we study divide-and-conquer inference with the de-biasing procedure for high-dimensional quantile regression. We provide key technical proofs for main results in Section 6. We provide thorough numerical experiment results in Section 7 and conclude the paper with Section 8. More detailled proofs are deferred to the appendix.

**Notations**. Let $F_\epsilon(t) := \mathbb{P}(\epsilon < t)$ and $f_\epsilon(t)$ be the cumulative distribution function and probability density function of $\epsilon$, respectively. For positive sequences $a_n$ and $b_n$, we write $a_n \lesssim b_n$ or $a_n = O(b_n)$ if there exist some universal constant constant $c > 0$ and positive integer $N$ independent of $n$ such that $a_n \le cb_n$ for all $n \ge N$. We write $a_n = o(b_n)$ if $a_n b_n^{-1} \to 0$ as $n \to \infty$. For a sequence of random variables $A_n$, we write $A_n = o_P(1)$ if $A_n \to 0$ in probability, and $A_n \rightsquigarrow A$ for some random variable $A$ if $A_n$ converges weakly to $A$. Let $\|\cdot\|_1$, $\|\cdot\|_2$ and $\|\cdot\|_\infty$ denote the Euclidean 1-, 2- and infinity norms, respectively. For any symmetric positive definite matrix $\mathbf{A}$, define $\|\boldsymbol{x}\|_{\mathbf{A}} := (\boldsymbol{x}^T\mathbf{A}\boldsymbol{x})^{1/2}$. Let $\|\cdot\|_2$ denote the matrix operator norm and $\|\cdot\|_{\max}$ the matrix element-wise max-norm, that is, $\|\mathbf{A}\|_{\max} = \max_{i,j}|\mathbf{A}_{ij}|$ for any matrix $\mathbf{A}$. Moreover, let $\|\mathbf{A}\|_{1,1} = \sum_{ij}|\mathbf{A}_{ij}|$ and $\|\mathbf{A}\|_{1,\max} = \max_i \sum_j |\mathbf{A}_{ij}|$. Let $A_{\cdot,j}$ and $A_{j,\cdot}$ denote the $j$-th column and row of matrix $\mathbf{A}$, respectively. Lastly, define $\mathrm{MSE}(\widehat{\boldsymbol{\beta}}) := \|\mathbb{X}(\widehat{\boldsymbol{\beta}} - \boldsymbol{\beta}^*)/\sqrt{n}\|_2$, where $\mathbb{X} := [\boldsymbol{X}_1, \ldots, \boldsymbol{X}_n]^T$ is the design matrix. We use $C_1, C_2, \ldots$ to denote absolute constants whose values may change from line to line.

# 2 Method

In this section, we introduce some background on composite quantile regression and motivate our de-biasing method. We then present the general procedure for constructing the de-biased estimator.

## 2.1 Background on Composite Quantile Regression

To motivate the composite quantile regression, we first review the usual quantile regression defined as follows: for some real number $\tau \in (0,1)$, one solves the following optimization problem

$$\min_{\boldsymbol{\beta}\in\mathbb{R}^p, b\in\mathbb{R}} \frac{1}{n}\sum_{i=1}^n \phi_\tau\big(Y_i - \boldsymbol{X}_i^T\boldsymbol{\beta} - b\big), \tag{2.1}$$

where $\phi_\tau$ is the check function $\phi_\tau(t) := \tau t_+ + (1-\tau)t_- = |t|/2 + (\tau - 1/2)t$, and $t_+ = \max\{t, 0\}$ and $t_- = \max\{-t, 0\}$. The check function is an extension of the absolute value function (when $\tau = 1/2$). It is used to compute any quantile of a sequence of real numbers. Specifically, given $a_1, a_2, \ldots, a_n \in \mathbb{R}$, we can compute the $\tau$-quantile of the $n$ numbers by finding $\arg\min_x \sum_{i=1}^n \phi_\tau(a_i - x)$. This is evident by inspecting the KKT conditions. Alternatively, we provide a pictorial explanation of how the check function computes the $\tau$-quantile in Figure 1.

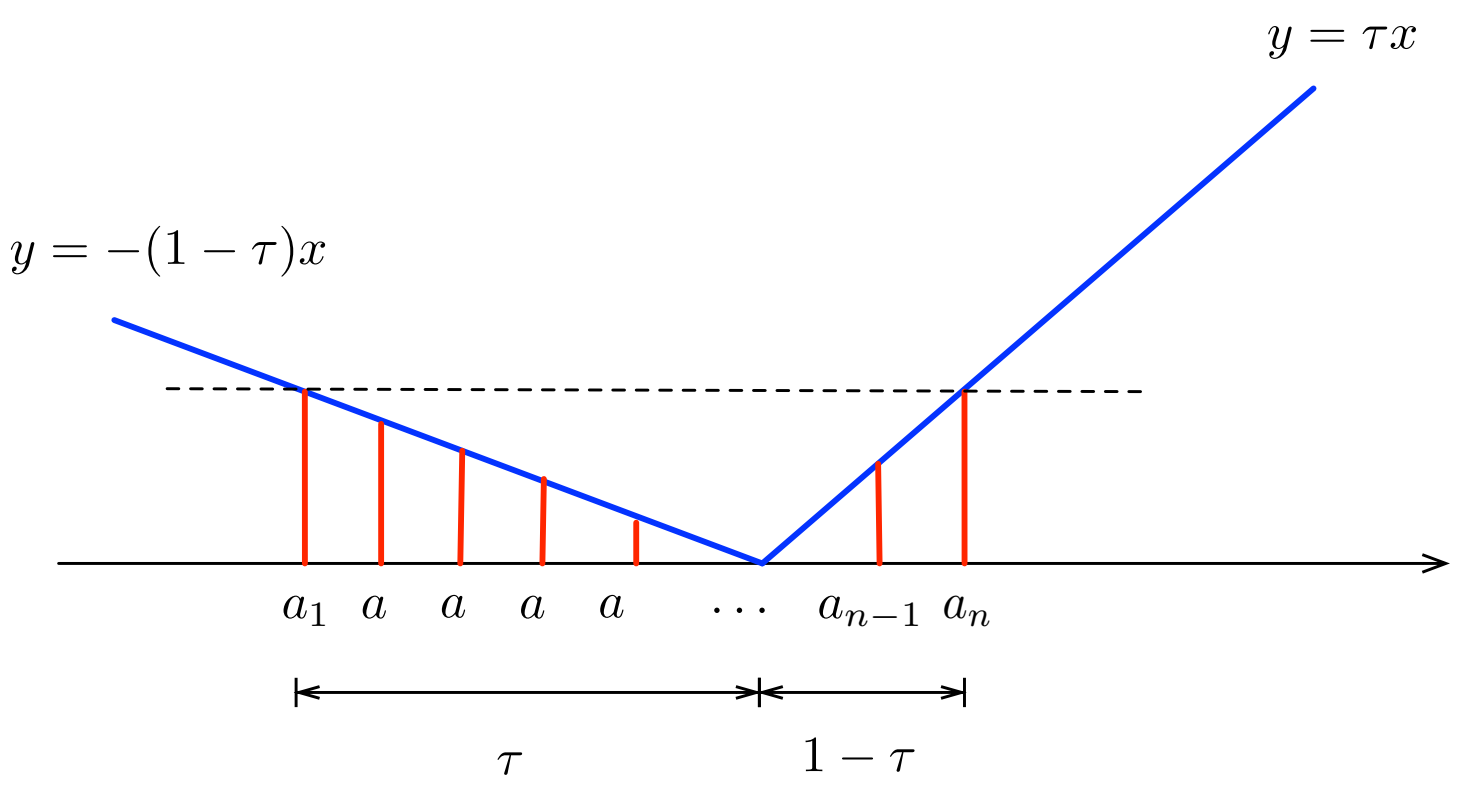

Figure 1: Illustration of the check function and how it computes the $\tau$-quantile. To minimize the function $\sum_{i=1}^n \phi_\tau(a_i - x)$, the two red bars on both ends should have equal lengths. In this scenario, the kink point of the check function is at the $\tau$-quantile of $n$ numbers.

Under the linear model (1.1), it is easy to see that the conditional $\tau$-quantile of $Y \mid \boldsymbol{X}$ is

$$\boldsymbol{X}^T\boldsymbol{\beta}^* + b_\tau^*,$$

where $b_\tau^*$ is the $\tau$-quantile of the residual term $\epsilon$. Therefore, estimating $\boldsymbol{\beta}^*$ and $b_\tau^*$ amounts to calculating the conditional quantile of $Y \mid \boldsymbol{X}$, resulting in the optimization problem (2.1).

A key observation that motivates the composite quantile regression is that for different $\tau$, the minimization problem solves for the same $\boldsymbol{\beta}^*$ and different $b_\tau^*$. Thus, we can utilize multiple quantile regressions which simultaneously solve $\boldsymbol{\beta}^*$ to boost its estimation performance (Zou and Yuan, 2008). The composite quantile loss function is defined as a sum of $K$ objective functions in (2.1) corresponding to $K$ check functions with different values of $\tau$. Each summand involves the same $\boldsymbol{\beta}$ but different $b$. Formally, for a sequence of $K$ real numbers $0 < \tau_1 < \tau_2 < \ldots < \tau_K < 1$, we solve

$$(\widetilde{\boldsymbol{\beta}}, \widetilde{b}_1, \ldots, \widetilde{b}_K) \in \operatorname*{argmin}_{\boldsymbol{\beta}\in\mathbb{R}^p, b_1,\ldots,b_K\in\mathbb{R}} \sum_{k=1}^K \frac{1}{n}\sum_{i=1}^n \phi_{\tau_k}(Y_i - \boldsymbol{X}_i^T\boldsymbol{\beta} - b_k). \tag{2.2}$$

Let $b_k^*$ denote the true $\tau_k$-quantile of $\epsilon$, and let $f_k^* := f_\epsilon(b_k^*)$, for $k = 1, \ldots, K$. Zou and Yuan (2008) showed that when $p$ is fixed and under mild regularity conditions, $\widetilde{\boldsymbol{\beta}}$ is asymptotically normal:

$$\sqrt{n}(\widetilde{\boldsymbol{\beta}} - \boldsymbol{\beta}^*) \rightsquigarrow N(\mathbf{0}, \sigma_K^2\theta_K^{-2}\boldsymbol{\Sigma}^{-1}),$$

where $\sigma_K^2 := \sum_{k,k'=1}^K \min\{\tau_k, \tau_{k'}\}(1 - \max\{\tau_k, \tau_{k'}\})$, $\theta_K := \sum_{k=1}^K f_k^*$, and $\boldsymbol{\Sigma} := \mathbb{E}[\boldsymbol{X}\boldsymbol{X}^T]$ is the population covariance matrix of the design. In high-dimensional regression, we add an $L_1$-regularization term to prevent overfitting:

$$(\widehat{\boldsymbol{\beta}}, \widehat{b}_1, \ldots, \widehat{b}_K) \in \operatorname*{argmin}_{\boldsymbol{\beta}\in\mathbb{R}^p, b_1,\ldots,b_K\in\mathbb{R}} \sum_{k=1}^K \frac{1}{n}\sum_{i=1}^n \phi_{\tau_k}(Y_i - \boldsymbol{X}_i^T\boldsymbol{\beta} - b_k) + \lambda\|\boldsymbol{\beta}\|_1, \tag{2.3}$$

where $\lambda$ is an regularization parameter. The choice of $\{\tau_k\}$ is arbitrary. For example, a simple choice is $1/(K+1), 2/(K+1), \ldots, K/(K+1)$.

### 2.2 A New De-biased Composite Quantile Estimator

Like the Lasso and other regularized estimators, the penalized CQR estimator (2.3) is biased and does not have a tractable limiting distribution. Our first contribution is to design a de-biased estimator based on composite quantile loss that allows us to construct robust confidence intervals and test hypotheses in high-dimensions. In particular, we obtain an asymptotically unbiased estimator from $\widehat{\boldsymbol{\beta}}$ by subtracting a term related to the sub-gradient of the composite quantile loss function, evaluated at the estimates $\widehat{\boldsymbol{\beta}}$ and $\widehat{b}_k$ for $k = 1, \ldots, K$:

$$\widehat{\boldsymbol{\beta}}^d = \widehat{\boldsymbol{\beta}} - \widehat{\boldsymbol{\Theta}}\widehat{\boldsymbol{\kappa}}, \tag{2.4}$$

where $\widehat{\boldsymbol{\kappa}} := \sum_{k=1}^K n^{-1} \sum_{i=1}^n \left( \mathbb{1}\{Y_i \le \boldsymbol{X}_i^T \widehat{\boldsymbol{\beta}} + \widehat{b}_k\} - \tau_k \right) \boldsymbol{X}_i$ is the sub-gradient of the objective in (2.2) and $\widehat{\boldsymbol{\Theta}}$ is a matrix constructed below. An interpretation of $\widehat{\boldsymbol{\kappa}}$ is that by the KKT condition, it is the sub-gradient of the $L_1$-penalty, which induces bias to $\widehat{\boldsymbol{\beta}}$. In the de-biasing framework studied in Zhang and Zhang (2013), van de Geer et al. (2014) and Javanmard and Montanari (2014), the role of $\widehat{\boldsymbol{\Theta}}$ is to invert the Hessian matrix of the loss function. However, in our case with composite quantile loss function, the Hessian does not exist. To handle this problem, we exploit the *local curvature* of the CQR loss function mentioned in Section 1. We show that the conditional expectation of the CQR loss given the design matrix has a local curvature $\theta_K \widehat{\boldsymbol{\Sigma}}$, and the role of $\widehat{\boldsymbol{\Theta}}$ in (2.4) is to approximately invert this rank-deficient matrix. We construct the estimator $\widehat{\boldsymbol{\Theta}}$ using a modification of the estimator given in Javanmard and Montanari (2014). More specifically, let $\mathbb{X} = (\boldsymbol{X}_1, \ldots, \boldsymbol{X}_n)^T \in \mathbb{R}^{n \times p}$ be the design matrix. For each $j$, we solve

$$\begin{aligned} \widehat{\boldsymbol{\mu}}_j \quad &= \quad \operatorname*{argmin}_{\boldsymbol{\mu}_j \in \mathbb{R}^p} \boldsymbol{\mu}_j^T \widehat{\boldsymbol{\Sigma}} \boldsymbol{\mu}_j \\ &\text{s.t.} \quad \left\| \widehat{\boldsymbol{\Sigma}} \boldsymbol{\mu}_j - \boldsymbol{e}_j \right\|_\infty \le \gamma_1, \ \left\| \mathbb{X} \boldsymbol{\mu}_j \right\|_\infty \le \gamma_2, \ \left| \frac{1}{\sqrt{n}} \sum_{i=1}^n \boldsymbol{\mu}_j^T \boldsymbol{X}_i \right| \le \gamma_3, \end{aligned} \tag{2.5}$$

for some $\gamma_1, \gamma_2, \gamma_3 > 0$. In practice, we choose $\gamma_1, \gamma_2$ and $\gamma_3$ to ensure that (2.5) is feasible. That is, we choose their values as small as possible, while still keeping the optimization problem feasible. Recall that $\theta_K := \sum_{k=1}^K f_k^*$. Let $\mathbf{M} := \left( \widehat{\boldsymbol{\mu}}_1, \ldots, \widehat{\boldsymbol{\mu}}_p \right)^T \in \mathbb{R}^{p \times p}$ and

$$\widehat{\boldsymbol{\Theta}} := \widehat{\theta}_K^{-1} \mathbf{M}, \tag{2.6}$$

where $\widehat{\theta}_K$ is a consistent estimator of $\theta_K$ satisfying the following condition for some $r_t = o(1)$:

$$\lim_{n \to \infty} \mathbb{P}\left( \left| \frac{\widehat{\theta}_K}{\theta_K} - 1 \right| \ge r_t \right) = 0. \tag{2.7}$$

To see the intuition of the definition of $\widehat{\boldsymbol{\Theta}}$, we examine the constrained minimization problem (2.5). The first constraint indicates the main role of $\mathbf{M}$ (and $\widehat{\boldsymbol{\Theta}}$), which is to "invert" the rank-deficient matrix $\widehat{\boldsymbol{\Sigma}}$. The other two constraints impose conditions on the design. The second constraint is equivalent to $\max_i \|\mathbf{M}\boldsymbol{X}_i\|_\infty \le \gamma_2$, which requires that each $\boldsymbol{X}_i$, after a linear transformation by $\mathbf{M}$, is uniformly bounded by $\gamma_2$. For example, if each row of $\mathbb{X}$ is sub-Gaussian, then the second constraint is feasible with high probability when $\gamma_2$ is of the order $\sqrt{\log(p \vee n)}$. The third condition indicates that the average of $\boldsymbol{X}_i$, after transformation by $\mathbf{M}$, converges to the expectation (which we assume to be $\mathbf{0}$) at the rate of $\gamma_3/\sqrt{n}$. This requirement can be easily met by a wide range of designs (for example, the sub-Gaussian design). In §3.1, we show that the asymptotic distribution of $\sqrt{n}(\widehat{\beta}_j^d - \beta_j^*)$ is normal with mean zero and variance $\sigma_K^2 \theta_K^{-2} \widehat{\boldsymbol{\mu}}_j^T \widehat{\boldsymbol{\Sigma}} \widehat{\boldsymbol{\mu}}_j$. Thus, the objective function in (2.5) minimizes the asymptotic variance subject to the constraints, with the aim of maximizing the estimation efficiency of the de-biased estimator.

Our second contribution is that we can replace the penalized CQR estimator $\widehat{\boldsymbol{\beta}}$ by any other first-stage estimators that is sparse and converges to the true parameter at the desired rate as the penalized CQR. It can be shown that any such estimator, after de-biasing, is first-order asymptotically equivalent. In Section 4, we discuss various choices of first-stage estimators.

Such a property greatly benefits computation. For example, if we use the penalized LAD as the first-stage estimator, the de-biasing method achieves the nice properties of CQR in high-dimensions, while only at the computational cost of solving a single penalized quantile regression. Note that to compute the penalized CQR estimator, we need to solve a linear programming that is $K$ times larger than that of the penalized LAD.

# 3 Inference Results

In this section, we state the main inference results of the paper. In §3.1 we show that each coordinate of the de-biased estimator is $\sqrt{n}$-consistent and weakly converges to a normal distribution, for any first-stage estimators $\widehat{\boldsymbol{\beta}}$ and $\widehat{\boldsymbol{b}} = (\widehat{b}_1, \ldots, \widehat{b}_K)^T$ that have the following non-asymptotic properties: $\widehat{\boldsymbol{\beta}}$ and $\widehat{\boldsymbol{b}}$'s converge to the true parameters at the $L_2$-rate $\sqrt{s(\log p)/n}$ and $\sqrt{Ks(\log p)/n}$, respectively, and $\widehat{\boldsymbol{\beta}}$ has empirical sparsity at the same order as $s$. Given any estimator satisfying the above requirements, the only condition we need for asymptotic normality is the following distributional assumption on the noise $\epsilon$.

**Assumption 3.1.** The density function $f_\epsilon$ of the random variable $\epsilon$ is uniformly bounded from above, and bounded from below at points $b_k^*$ for $k = 1, \ldots, K$, that is, $\sup_{t \in \mathbb{R}} f_\epsilon(t) < C_+$ and $\min_k f_k^* > C_-$, for some constants $C_+, C_- > 0$. Moreover, we assume that $f_\epsilon$ is differentiable with the first order derivative uniformly bounded by some constant $C'_+ > 0$.

Assumption 3.1 imposes mild assumptions on the density function of $\epsilon$ and allows $\epsilon$ to follow a wide range of distributions. For example, it includes heavy-tailed distributions without first two moments, mixture of distributions, and distributions with outliers.

## 3.1 Asymptotic Properties of the De-biased Estimator

Recall that $\widehat{\boldsymbol{b}} = (\widehat{b}_1, \ldots, \widehat{b}_K)^T \in \mathbb{R}^K$. We similarly define $\boldsymbol{b}^* = (b_1^*, \ldots, b_K^*)^T \in \mathbb{R}^K$. The following theorem provides the asymptotic property of the de-biased estimator $\widehat{\boldsymbol{\beta}}^d$.

**Theorem 3.2.** Under Assumption 3.1, suppose we have $\|\widehat{\boldsymbol{\beta}}\|_0 \le c^* s$, $\mathrm{MSE}(\widehat{\boldsymbol{\beta}}) = \|\mathbb{X}(\widehat{\boldsymbol{\beta}} - \boldsymbol{\beta}^*)/\sqrt{n}\|_2 \le C_1\sqrt{s(\log p)/n}$, $\|\widehat{\boldsymbol{\beta}} - \boldsymbol{\beta}^*\|_2 \le C_2\sqrt{s(\log p)/n}$ and $\|\widehat{\boldsymbol{b}} - \boldsymbol{b}^*\|_2 \le C_3\sqrt{Ks(\log p)/n}$ for some constants $c^*, C_1, C_2$ and $C_3$ with probability tending to 1. Then under the scaling conditions that $\gamma_1 s\sqrt{\log p} = o(1)$, $\gamma_3\sqrt{s(\log p)/n} = o(1)$, $\gamma_2 s(\log p)/\sqrt{n} = o(1)$, $r_t\gamma_2\sqrt{s\log p} = o(1)$ and $\gamma_2(s\log p)^{3/4}/n^{1/4} = o(1)$, we have

$$\sqrt{n}(\widehat{\boldsymbol{\beta}}^d - \boldsymbol{\beta}^*) = \boldsymbol{Z} + \boldsymbol{W}, \tag{3.1}$$

where $\boldsymbol{Z} = n^{-1/2}\sum_{i=1}^n \theta_K^{-1}\mathbf{M}\boldsymbol{X}_i\Psi_{i,K}$, $\Psi_{i,K} = \sum_{k=1}^K \big(\tau_k - \mathbb{1}\{\epsilon_i \le b_k^*\}\big)$, and $\|\boldsymbol{W}\|_\infty = o_P(1)$.

Theorem 3.2 shows that for each $j$, $\sqrt{n}(\widehat{\beta}_j^d - \beta_j^*)$ can be decomposed into a term which we later show is asymptotically normal, and a term that is asymptotically negligible. In particular, $\widehat{\beta}_j^d$ is a $\sqrt{n}$-consistent estimator of $\beta_j^*$ under any error distribution satisfying Assumption 3.1. The conclusions of Theorem 3.2 follow from the non-asymptotic properties (convergence rates and empirical sparsity) of the first-stage estimators $(\widehat{\boldsymbol{\beta}}, \widehat{\boldsymbol{b}})$. In Section 4, we provide several examples of such estimators.

Although Theorem 3.2 requires a number of scaling conditions, it can be shown that the last one $\gamma_2(s\log p)^{3/4}/n^{1/4} = o(1)$ usually dominates the remaining ones (see §3.2 for details). This condition requires that $s = o\big(\gamma_2^{-4/3}n^{1/3}/\log p\big)$. For well behaved design, $\gamma_2$ grows slowly with $n$. For example, $\gamma_2 = \sqrt{\log(p \vee n)}$ under the sub-Gaussian design. In previous study of high-dimensional inference for quantile and least absolute deviation regressions, Belloni et al. (2013a) and Belloni et al. (2013b) also require a similar scaling condition $(K_x^2 s^2 + s^3)\log^3 p/n = o(1)$, where $K_x = \|\mathbb{X}\|_{\max}$. It appears that such scaling conditions are unavoidable for inference with non-smooth loss functions.

We next focus on the asymptotic property of $\boldsymbol{Z} = (Z_1, \ldots, Z_p)^T \in \mathbb{R}^p$ in (3.1). The following theorem shows that each entry $Z_j$ weakly converges to a normal distribution.

**Theorem 3.3.** Suppose $\gamma_2 n^{-1/2} = o(1)$ and that $(\widehat{\boldsymbol{\mu}}_j^T\widehat{\boldsymbol{\Sigma}}\widehat{\boldsymbol{\mu}}_j)^{1/2} \ge c_n$, with probability tending to 1, where $c_n$ satisfies $\liminf_{n\to\infty} c_n = c_\infty > 0$. With no assumption on $\epsilon$, we have for any $j = 1, \ldots, p$

$$\frac{Z_j}{(\widehat{\boldsymbol{\mu}}_j^T\widehat{\boldsymbol{\Sigma}}\widehat{\boldsymbol{\mu}}_j)^{1/2}} \rightsquigarrow N(0, \theta_K^{-2}\sigma_K^2),$$

where $\sigma_K$ and $\theta_K$ are defined in §2.1.

Theorem 3.3 imposes no assumption on noise $\epsilon$ and a mild assumption on the design, which can be verified under the sub-Gaussian condition in the next section (shown by Lemma B.1 in Appendix B.3). Combining Theorems 3.2 and 3.3, we conclude that $\sqrt{n}(\widehat{\beta}_j - \beta_j^*)$ converges weakly to a Gaussian distribution. Based on this result, we can conduct various inference tasks, including construction of confidence intervals and hypothesis testing. We can also extend the coordinate-wise inference result to infer a generic form $\mathbf{Q}\boldsymbol{\beta}^*$, where $\mathbf{Q}$ projects $\boldsymbol{\beta}^*$ into a low-dimensional space. The details are discussed in §3.4.

## 3.2 Feasibility of Local Curvature Estimate and Trade-off Between Efficiency and Scaling

We now discuss the choice of the tuning parameters $\gamma_1$, $\gamma_2$ and $\gamma_3$ in the optimization problem (2.5) for calculating $\mathbf{M}$. Choosing the parameters too small will render the problem infeasible, while making them too large will impose stronger scaling conditions on $s, n$ and $p$ (See Theorem 3.2).

As explained in §2.2, $\gamma_1$ reflects the quality of inverting the sample covariance matrix, while $\gamma_2$ and $\gamma_3$ impose conditions on the design. For well-behaved designs, both requirements can be met for $\gamma_1$, $\gamma_2$ and $\gamma_3$ not too large. We consider two cases: sub-Gaussian and strongly bounded designs. Both are commonly used in the literature. We first recall the definition of sub-Gaussian random variables.

**Definition 3.4** (Sub-Gaussian Random Variables)**.** A random variable $X$ is $\sigma^2$-sub-Gaussian if for all $t > 0$, one has

$$\mathbb{P}(|X| > t) \le 2e^{-t^2/2\sigma^2}.$$

Here $\sigma^2$ is called the variance proxy of $X$. A random vector $\boldsymbol{X} \in \mathbb{R}^p$ is $\sigma^2$-sub-Gaussian if for all $\boldsymbol{u} \in \mathbb{R}^p$ with $\|\boldsymbol{u}\|_2 = 1$, we have $\boldsymbol{u}^T\boldsymbol{X}$ is $\sigma^2$-sub-Gaussian.

**Assumption 3.5.** $\boldsymbol{X}_1, \ldots, \boldsymbol{X}_n$ are $\sigma_x^2$-sub-Gaussian random vectors.

Alternatively, we assume the designs are strongly bounded after transformation by the precision matrix $\boldsymbol{\Sigma}^{-1}$.

**Assumption 3.5a.** $\boldsymbol{\Sigma}^{-1}\boldsymbol{X}_1, \ldots, \boldsymbol{\Sigma}^{-1}\boldsymbol{X}_n$ have entries uniformly bounded by a constant $C_X$.

In addition, we require the following assumption on the minimum and maximum eigenvalues of the population covariance matrix $\boldsymbol{\Sigma}$:

**Assumption 3.6.** The matrix $\boldsymbol{\Sigma}$ has eigenvalues uniformly upper bounded by $\rho_{\max}$ and lower bounded by $\rho_{\min}$.

This assumption is also standard and commonly assumed in the literature. Under this condition, Assumption 3.5a will imply Assumption 3.5 by Hoeffding's inequality. More details can be found in the proof of Proposition 3.8.

Our assumptions most resemble those in Javanmard and Montanari (2014). In other work on high-dimensional inference, the sparsity or similar conditions on the precision matrix $\boldsymbol{\Sigma}^{-1}$ are imposed, so as to obtain a consistent estimator for the asymptotic variance (see, for example, van de Geer et al., 2014; Belloni et al., 2012, 2013a). However, we do not require such assumptions as we only need to control $\|\mathbf{M}\widehat{\boldsymbol{\Sigma}} - \mathbf{I}\|_{\max}$ for the estimator $\mathbf{M}$ in order to show that the remainder term is asymptotically negligible. A more intuitive reason is that to construct a confidence interval for $\beta_j^*$, we only need to estimate $[\boldsymbol{\Sigma}^{-1}]_{jj}$ rather than the entire matrix $\boldsymbol{\Sigma}^{-1}$. Similarly, only a sub-matrix of $\boldsymbol{\Sigma}^{-1}$ is estimated in order to construct a confidence region for a set of low-dimensional coordinates of $\boldsymbol{\beta}^*$. In §3.5 where we consider simultaneous test for the whole vector $\boldsymbol{\beta}^*$, we do need to estimate the entire matrix $\boldsymbol{\Sigma}^{-1}$, hence the sparsity condition of $\boldsymbol{\Sigma}^{-1}$ is imposed in that section.

In the following, we show that under either design specified in Assumption 3.5 or 3.5a, we can choose the parameters $\gamma_1$, $\gamma_2$ and $\gamma_3$ that ensure the feasibility of (2.5), without imposing strong requirements on the scaling of $s, n$ and $p$.

**Proposition 3.7** (Feasibility under sub-Gaussian design)**.** Under Assumptions 3.5 and 3.6, if we choose $\gamma_1 = a_1\sqrt{(\log p)/n}$, $\gamma_2 = a_2\sqrt{\log(p \vee n)}$ and $\gamma_3 = a_3\sqrt{\log(p \vee n)}$, where $a_1 = 6e\sigma_x\sqrt{2\rho_{\max}/\rho_{\min}}$, $a_2 = \sqrt{3}\sigma_x/\rho_{\min}$ and $a_3 = \sqrt{2}\sigma_x/\rho_{\min}$, then with probability at least $1 - 3p^{-1}$, we have

$$\big\|\boldsymbol{\Sigma}^{-1}\widehat{\boldsymbol{\Sigma}} - \mathbf{I}_p\big\|_{\max} \le \gamma_1, \quad \|\mathbb{X}\boldsymbol{\Sigma}^{-1}\|_{\max} \le \gamma_2, \quad \text{and} \quad \Big\|\frac{1}{\sqrt{n}}\sum_{i=1}^n \boldsymbol{\Sigma}^{-1}\boldsymbol{X}_i\Big\|_\infty \le \gamma_3. \tag{3.2}$$

Hence, with at least the same probability, (2.5) has a feasible solution for all $j \in \{1, \ldots, p\}$.

Therefore, under a sub-Gaussian design and with the choices of $\gamma_1, \gamma_2$ and $\gamma_3$ specified in Theorem 3.7, the optimization problem in (2.5) has a feasible solution with high probability. The scaling conditions in Theorem 3.2 reduce to

$$s = o\big(n^{1/3}/\log^{5/3}(p \vee n)\big) \quad \text{and} \quad r_t\sqrt{s}\log(p \vee n) = o(1). \tag{3.3}$$

For strongly bounded designs, the following theorem shows that we can choose a smaller $\gamma_2$ to further weaken the scaling condition.

**Proposition 3.8** (Feasibility under strongly bounded design)**.** Under Assumptions 3.5a and 3.6, let $\gamma_1 = a_1\sqrt{(\log p)/n}$, $\gamma_2 = C_X$ and $\gamma_3 = a_3\sqrt{\log(p \vee n)}$, where $a_1 = 6eC_X\sqrt{2\rho_{\max}^3/\rho_{\min}}$ and $a_3 = \sqrt{2}C_X\rho_{\max}/\rho_{\min}$. Then with probability at least $1 - 2p^{-1}$, we have

$$\big\|\widehat{\mathbf{\Sigma}}\mathbf{\Sigma}^{-1} - \mathbf{I}_p\big\|_{\max} \leq \gamma_1, \quad \big\|\mathbb{X}\mathbf{\Sigma}^{-1}\big\|_{\max} \leq \gamma_2, \quad \text{and} \quad \Big\|\frac{1}{\sqrt{n}}\sum_{i=1}^{n}\mathbf{\Sigma}^{-1}\boldsymbol{X}_i\Big\|_{\infty} \leq \gamma_3. \tag{3.4}$$

Therefore, (2.5) has a feasible solution for all $j \in \{1, \ldots, p\}$ with the same probability .

Under the choices of $\gamma_1, \gamma_2, \gamma_3$ in Theorem 3.8, the scaling conditions in Theorem 3.2 reduce to

$$s = o\big(n^{1/3}/\log(p \vee n)\big) \quad \text{and} \quad r_t\sqrt{s\log p} = o(1). \tag{3.5}$$

Note that the optimization problem (2.5) displays a tradeoff between the efficiency of $\widehat{\boldsymbol{\beta}}^d$ and the scaling of $s, n$ and $p$. Choosing large values for $\gamma_1$, $\gamma_2$ and $\gamma_3$ decreases the optimal value of (2.5) thus the variance of $\widehat{\beta}_j^d$, but requires $s$ to grow slower with $n$.

### 3.3 Estimation of $\widehat{\theta}_K$

To calculate $\widehat{\boldsymbol{\beta}}^d$, we require an estimator $\widehat{\theta}_K$ of $\theta_K$, where recall $\theta_K = \sum_{k=1}^K f_k^*$ and $f_k^* = f_\epsilon(b_k^*)$. Therefore, we need to estimate the density function at the $K$ quantiles of $\epsilon$. In this section, we discuss how to estimate this quantity.

In the case that the density function is known, we can estimate $\theta_K = \sum_{k=1}^K f_\epsilon(b_k^*)$ by directly plugging in $\widehat{b}_k$ for $k = 1, \ldots, K$. Then under Assumption 3.1, we have

$$\left|\frac{\widehat{\theta}_K}{\theta_K} - 1\right| = |\theta_K^{-1}(\widehat{\theta}_K - \theta_K)| \leq C_-^{-1}\frac{1}{K}\sum_{k=1}^{K}|f_\epsilon(\widehat{b}_k) - f_\epsilon(b_k^*)| \leq C_-^{-1}C_+'\frac{1}{K}\|\widehat{\boldsymbol{b}} - \boldsymbol{b}^*\|_1 \leq C_-^{-1}C_+'\frac{1}{\sqrt{K}}\|\widehat{\boldsymbol{b}} - \boldsymbol{b}^*\|_2,$$

where the last inequality is by Cauchy-Schwartz. Therefore, by the convergence of $\widehat{\boldsymbol{b}}$, we have

$$\lim_{n\to\infty}\mathbb{P}\left(\left|\frac{\widehat{\theta}_K}{\theta_K} - 1\right| \geq C_3'\sqrt{\frac{s\log p}{n}}\right) = 0, \tag{3.6}$$

where $C_3' = C_-^{-1}C_+'C_3$.

In the case that the density function is unknown, we exploit the method outlined in Koenker (2005) to estimate $f_k^*$. The idea is motivated by the following observation:

$$\frac{d}{d\tau}F_\epsilon^{-1}(\tau) = \big[f_\epsilon(F_\epsilon^{-1}(\tau))\big]^{-1}.$$

Let $Q_\epsilon(\tau) := F_\epsilon^{-1}(\tau)$ be the quantile function of $\epsilon$. The above equation and the relationship $Q_\epsilon(\tau_k) = b_k^*$ yield that $Q_\epsilon'(\tau_k) = f_k^{*-1}$. Therefore, to estimate $f_k^*$, we let

$$\widehat{f}_k = \frac{2h}{\widehat{Q}(\tau_k + h) - \widehat{Q}(\tau_k - h)},$$

where $h$ is a bandwidth to be chosen later, and $\widehat{Q}(\tau)$ is an estimator of the $\tau$-quantile of $\epsilon$, obtained in the same way as $\widehat{b}_k = \widehat{Q}(\tau_k)$. The following proposition gives the estimation rate of $\widehat{f}$.

**Proposition 3.9.** Suppose Assumption 3.1 holds and that $\sup_{|\tau-\tau_k|\le h}|Q'''(\tau)| \le C_0$. Moreover, suppose that with probability tending to 1, we have $|\widehat{Q}(\tau_k+u) - Q(\tau_k+u)| \le C\sqrt{s(\log p)/n}$ for $u = \pm h$ and some constant $C$ that does not depend on $n$, $p$ and $s$. Then under the scaling condition $\sqrt{s(\log p)/h^2 n} = o(1)$ and $h = o(1)$, we have

$$|\widehat{f}_k - f_k^*| \lesssim \sqrt{\frac{s\log(n\vee p)}{h^2 n}} + h^2, \tag{3.7}$$

with probability tending to 1.

The error bound in (3.7) represents a bias-variance tradeoff. To obtain the optimal rate of convergence , we choose $h^2 = \sqrt{s\log(p\vee n)/(h^2 n)}$, which yields $h = (s\log(p\vee n)/n)^{1/6}$ and

$$|\widehat{f}_k - f_k^*| \lesssim \left(\frac{s\log(p\vee n)}{n}\right)^{1/3}.$$

Using the same argument as in the case with known density, we obtain

$$\lim_{n\to\infty}\mathbb{P}\left(\left|\frac{\theta_K}{\theta_K} - 1\right| \ge C\left(\frac{s\log(p\vee n)}{n}\right)^{1/3}\right) = 0. \tag{3.8}$$

Based on (3.6) and (3.8), we conclude that $r_t = \sqrt{(s\log p)/n}$ when $f_\epsilon$ is known and $r_t = (s\log(p\vee n)/n)^{1/3}$ when $f_\epsilon$ is unknown. In both cases, by (3.3) and (3.5), the scaling conditions in Theorem 3.2 become

$$s = o\bigg(\frac{n^{1/3}}{\log^{5/3}(p\vee n)}\bigg), \quad \text{and} \quad s = o\bigg(\frac{n^{1/3}}{\log(p\vee n)}\bigg)$$

under sub-Gaussian and strongly bounded designs, respectively.

### 3.4 Robust and Uniformly Honest Inference for Low-Dimensional Coordinates

Based on the theoretical results presented in previous sections, we now conduct statistical inference for the unknow parameters. We start with coordinate-wise inference for $\boldsymbol{\beta}^*$.

**Theorem 3.10.** Suppose Assumptions 3.1, 3.5 (or 3.5a) and 3.6 hold. Consider the de-biased estimator $\widehat{\boldsymbol{\beta}}^d$ defined by (2.4), in which $\widehat{\boldsymbol{\beta}}, \widehat{b}_1, \ldots, \widehat{b}_K$ satisfy the conditions in Theorem 3.2, and $\widehat{\boldsymbol{\Theta}}$ is obtained through (2.5) and (2.6), with $\gamma_1, \gamma_2$ and $\gamma_3$ chosen by Proposition 3.7 under Assumption 3.5 (or by Proposition 3.8 under 3.5a), and $\widehat{\theta}_K$ satisfying (2.7). Let $s = s(n)$ be the sparsity of $\boldsymbol{\beta}^*$ satisfying the scaling conditions $s = o(n^{1/3}/\log^{5/3}(p\vee n))$ and $r_t\sqrt{s}\log(p\vee n) = o(1)$ (or $s = o(n^{1/3}/\log p)$ and $r_t\sqrt{s\log p} = o(1)$). Then for any $\boldsymbol{\beta}^* \in \mathbb{R}^p$ with $\|\boldsymbol{\beta}^*\|_0 \le s$, we have

$$\frac{\sqrt{n}(\widehat{\beta}_j^d - \beta_j^*)}{(\widehat{\boldsymbol{\mu}}_j^T\widehat{\boldsymbol{\Sigma}}\widehat{\boldsymbol{\mu}}_j)^{1/2}\widehat{\theta}_K^{-1}} \rightsquigarrow N(0, \sigma_K^2). \tag{3.9}$$

Note that in order for (3.9) to hold, we only require $\epsilon$ to satisfy Assumption 3.1. Therefore, our procedure is robust and can be used to conduct inference on the linear model (1.1) under a wide range of noise distributions. In the following, we compare the inferential power of our de-biased CQR method with other alternatives:

- Javanmard and Montanari (2014) studied the de-biased Lasso estimator under similar design conditions. In order to conduct valid statical inference, strong distributional assumptions on $\epsilon$ are required, such as Gaussian distribution or existence of higher order moment. When $\epsilon$ satisfies such assumptions, one has

$$\frac{\sqrt{n}(\widehat{\beta}_j^{\text{dLasso}} - \beta_j^*)}{(\widehat{\boldsymbol{\mu}}_j^T \widehat{\boldsymbol{\Sigma}} \widehat{\boldsymbol{\mu}}_j)^{1/2}} \to N(0, \sigma^2), \tag{3.10}$$

  where $\sigma^2$ is the variance of $\epsilon$. We compare estimation efficiencies of the two methods using *asymptotic relative efficiency*, which is defined as the ratio between the asymptotic variances of de-biased Lasso (3.10) and de-biased CQR (3.9) estimators:

$$\text{ARE}_1(K, f_\epsilon) = \frac{\sigma^2 \theta_K^2}{\sigma_K^2}.$$

  The asymptotic relative efficiency is exactly the same as that of the ordinary least square to CQR estimator in the low-dimensional settings (see Equation (3.1) in Zou and Yuan, 2008). Therefore, our de-biased composite quantile estimator in high-dimensional settings inherits the inferential power of its low-dimensional counterpart. In particular, it is shown in Zou and Yuan (2008) that

$$\inf_{f_\epsilon \in \mathcal{F}} \lim_{K \to \infty} \text{ARE}_1(K, f_\epsilon) > \frac{6}{e\pi} = 0.7026,$$

  where $\mathcal{F}$ is a collection of all density functions satisfying Assumption 3.1. This shows that the efficiency loss of de-biased composite quantile estimator to de-biased Lasso is at most 30% when $K$ is large. Moreover, in many applications, the relative efficiency is close to or higher than 1. For example, under Gaussian distribution, $\text{ARE}_1(K, f_\epsilon) \to 0.955$ when $K \to \infty$. For t-distribution with degree of freedom 3, $\text{ARE}_1(K, f_\epsilon) \to 1.9$ when $K \to \infty$, showing that the CQR method has a big gain in efficiency. More examples are provided in Zou and Yuan (2008).

- When $K = 1$, our procedure becomes the de-biased quantile regression estimator, which is also robust for high-dimensional linear models. The asymptotic variance of $\sqrt{n}(\widehat{\beta}_j^d - \beta_j^*)/(\widehat{\boldsymbol{\mu}}_j^T \widehat{\boldsymbol{\Sigma}} \widehat{\boldsymbol{\mu}}_j)^{1/2}$ is $\tau(1-\tau)/f_k^*$ when $K = 1$, hence its relative efficiency to de-biased Lasso is the same as the low-dimensional counterpart of quantile regression to ordinary least square (Koenker, 2005). Therefore, the de-biased quantile regression can have arbitrarily small relative efficiency compared to the de-biased Lasso. We note that Belloni et al. (2013a) and Belloni et al. (2013b) established asymptotic normality for a single coordinate of $\boldsymbol{\beta}^*$ in high-dimensional quantile and least absolute deviation regressions under the instrumental variable framework, while requiring that the precision matrix of the design to be sparse.

As the relative efficiency is the same as in the low-dimensional settings, we in practice adopt the choice of $K$ as recommended by Zou and Yuan (2008), who showed that $\text{ARE}_1(K, f_\epsilon)$ when $K = 9$ is already close to the limit of $K \to \infty$ for various examples of $f_\epsilon$.

Based on the asymptotic distribution of the de-biased CQR estimator, we can conduct various inference tasks for a coordinate $\beta_j^*$. More generally, we consider the problem of constructing confidence regions or testing hypotheses for $\mathbf{Q}\boldsymbol{\beta}^*$, where $\mathbf{Q} \in \mathbb{R}^{q \times p}$ satisfies the following conditions: $\mathbf{Q}$ is full rank, $q < p$ and does not scale with $p$ or $n$, and $\|\mathbf{Q}\|_{1,1}$ is bounded by a constant that does not grow with $p$ or $n$. For example, $H_0 : \mathbf{Q}\boldsymbol{\beta}^* = \mathbf{0}$ where $\mathbf{Q} = (\mathbf{I}_q, \mathbf{0})$ tests the hypothesis that $\beta_j^* = 0$ for all $1 \le j \le q$, and $H_0' : \mathbf{Q}\boldsymbol{\beta}^* = \mathbf{0}$ where $\mathbf{Q} = (\mathbf{Q}_1, \mathbf{0})$ and

$$\mathbf{Q}_1 = \begin{pmatrix} 1 & -1 & 0 & \dots & 0 & 0 \\ 0 & 1 & -1 & \dots & 0 & 0 \\ \vdots & \vdots & \vdots & \ddots & \vdots & \vdots \\ 0 & 0 & 0 & \dots & 1 & -1 \end{pmatrix} \in \mathbb{R}^{q \times (q+1)}$$

tests the hypothesis that $\beta_1^* = \beta_2^* = \ldots = \beta_{q+1}^*$.

Furthermore, we extend the pointwise result in Theorem 3.10 to uniform convergence result, which enables us to construct uniformly honest confidence intervals and hypothesis tests. Define the following events

$$\mathcal{E}_1(\boldsymbol{\beta}^*) = \left\{ \|\widehat{\boldsymbol{\beta}}\|_0 \le c^* s \right\}, \qquad \mathcal{E}_2(\boldsymbol{\beta}^*) = \left\{ \mathrm{MSE}(\widehat{\boldsymbol{\beta}}) \le C_1 \sqrt{s(\log p)/n} \right\},$$

$$\mathcal{E}_3(\boldsymbol{\beta}^*) = \left\{ \|\widehat{\boldsymbol{\beta}} - \boldsymbol{\beta}^*\|_2 \le C_2 \sqrt{s(\log p)/n} \right\}, \qquad \text{and} \qquad \mathcal{E}_4(\boldsymbol{\beta}^*) = \left\{ \|\widehat{\boldsymbol{b}} - \boldsymbol{b}^*\|_2 \le C_3 \sqrt{Ks(\log p)/n} \right\}.$$

**Theorem 3.11.** Suppose Assumptions 3.1, 3.5 (or 3.5a) and 3.6 hold. Let $\mathbb{P}_*$ denote the probability measure under the true parameter $\boldsymbol{\beta}^*$. Consider the de-biased estimator $\widehat{\boldsymbol{\beta}}^d$ defined by (2.4), in which $\widehat{\boldsymbol{\beta}}, \widehat{b}_1, \ldots, \widehat{b}_K$ satisfy

$$\lim_{n \to \infty} \sup_{\boldsymbol{\beta}^* \in \mathbb{R}^p, \|\boldsymbol{\beta}^*\|_0 \le s} \mathbb{P}_* \big( \mathcal{E}_1(\boldsymbol{\beta}^*)^c \cup \mathcal{E}_2(\boldsymbol{\beta}^*)^c \cup \mathcal{E}_3(\boldsymbol{\beta}^*)^c \cup \mathcal{E}_4(\boldsymbol{\beta}^*)^c \big) = 0 \tag{3.11}$$

and $\widehat{\boldsymbol{\Theta}}$ is obtained through (2.5) and (2.6), with $\gamma_1, \gamma_2$ and $\gamma_3$ chosen by Proposition 3.7 under Assumption 3.5 (or by Proposition 3.8 under 3.5a) and $\widehat{\theta}_K$ satisfying the following condition

$$\lim_{n \to \infty} \sup_{\boldsymbol{\beta}^* \in \mathbb{R}^p, \|\boldsymbol{\beta}^*\|_0 \le s} \mathbb{P}_* \left( \left| \frac{\widehat{\theta}_K}{\theta_K} - 1 \right| \ge r_t \right) = 0. \tag{3.12}$$

Let $s = s(n)$ be the sparsity of $\boldsymbol{\beta}^*$ satisfying the scaling condition $s = o\big(n^{1/3} / \log^{5/3}(p \vee n)\big)$ and $r_t \sqrt{s} \log(p \vee n) = o(1)$ (or $s = o\big(n^{1/3}/\log p\big)$ and $r_t \sqrt{s \log p} = o(1)$). Let $\mathbf{A} := \mathbf{Q}\mathbf{M}\widehat{\boldsymbol{\Sigma}}\mathbf{M}^T\mathbf{Q}^T$. Then for any $\boldsymbol{x} \in \mathbb{R}^q$, we have

$$\lim_{n \to \infty} \sup_{\boldsymbol{\beta}^* \in \mathbb{R}^p, \|\boldsymbol{\beta}^*\|_0 \le s} \left| \mathbb{P}_* \left( \sqrt{n} \sigma_K^{-1} \widehat{\theta}_K \mathbf{A}^{-1/2} (\mathbf{Q}\widehat{\boldsymbol{\beta}}^d - \mathbf{Q}\boldsymbol{\beta}^*) \le \boldsymbol{x} \right) - \boldsymbol{\Phi}(\boldsymbol{x}) \right| = 0, \tag{3.13}$$

where $\boldsymbol{\Phi}(\boldsymbol{x}) = \Phi(x_1)\Phi(x_2)\ldots\Phi(x_q)$, with $\Phi$ denoting the cumulative distribution function of the standard normal random variable, and the inequality in (3.13) is considered elementwise.

Based on Theorem 3.11, we can construct a uniformly valid confidence region for the general form $\mathbf{Q}\boldsymbol{\beta}^*$ as follows: let $C_\alpha$ be a set in $\mathbb{R}^q$ such that $\mathbb{P}(\boldsymbol{R} \in C_\alpha) = 1 - \alpha$ for some random vector $\boldsymbol{R} \sim N(\mathbf{0}, \mathbf{I}_q)$. The confidence region at $1 - \alpha$ level for $\mathbf{Q}\boldsymbol{\beta}^*$ is

$$\mathbf{Q}\widehat{\boldsymbol{\beta}}^d + n^{-1/2} \sigma_K \widehat{\theta}_K^{-1} \mathbf{A}^{1/2} C_\alpha := \big\{ \mathbf{Q}\widehat{\boldsymbol{\beta}}^d + n^{-1/2} \sigma_K \widehat{\theta}_K^{-1} \mathbf{A}^{1/2} \boldsymbol{x} \,|\, \boldsymbol{x} \in C_\alpha \big\},$$

where $\mathbf{A}$ is defined in Theorem 3.11. This confidence region is uniformly valid, since by Theorem 3.11, we have

$$\lim_{n \to \infty} \sup_{\boldsymbol{\beta}^* \in \mathbb{R}^p, \|\boldsymbol{\beta}^*\|_0 \le s} \left| \mathbb{P} \left( \mathbf{Q}\boldsymbol{\beta}^* \notin \mathbf{Q}\widehat{\boldsymbol{\beta}}^d + n^{-1/2} \sigma_K \widehat{\theta}_K^{-1} \mathbf{A}^{1/2} C_\alpha \right) - \alpha \right| = 0.$$

Next, we test the simple hypothesis $H_0 : \mathbf{Q}\boldsymbol{\beta}^* = \boldsymbol{\omega}$ versus $H_1 : \mathbf{Q}\boldsymbol{\beta}^* \neq \boldsymbol{\omega}$. Our level-$\alpha$ Wald test is constructed as follows:

$$\Psi_\alpha = \mathbb{1}\left\{\boldsymbol{\omega} \notin \mathbf{Q}\widehat{\boldsymbol{\beta}}^d + n^{-1/2}\sigma_K\widehat{\theta}_K^{-1}\mathbf{A}^{1/2}C_\alpha\right\}.$$

The consistency of the above test is guaranteed by Theorem 3.11, which implies $\lim_{n\to\infty}\mathbb{P}_0(\Psi_\alpha = 1) = \alpha$, where $\mathbb{P}_0$ is the probability under the null hypothesis $H_0$.

More generally, we consider the following composite null and alternative hypotheses. For simplicity we only study the case with one coordinate. The test is formulated as follows:

$$H_0 : \boldsymbol{\beta}^* \in \mathcal{B}_0 \qquad \text{v.s.} \qquad H_1 : \boldsymbol{\beta}^* \in \mathcal{B}_a, \tag{3.14}$$

where $\mathcal{B}_0 = \{\boldsymbol{\beta} \in \mathbb{R}^p : \|\boldsymbol{\beta}\|_0 \le s, \beta_j = 0\}$ and $\mathcal{B}_a = \{\boldsymbol{\beta} \in \mathbb{R}^p : \|\boldsymbol{\beta}\|_0 \le s, \beta_j = an^{-\gamma}\}$, and $a$ and $\gamma$ are some known constants. The level-*alpha* test is constructed as

$$\Psi_j = \begin{cases} 0, & \text{if } |\widehat{\beta}_j^d| \le \Phi^{-1}(1-\alpha/2)\sigma_K(\widehat{\boldsymbol{\mu}}_j^T\widehat{\boldsymbol{\Sigma}}\widehat{\boldsymbol{\mu}}_j)^{1/2}/(\sqrt{n}\widehat{\theta}_K); \\ 1, & \text{otherwise.} \end{cases}$$

The following theorem provides uniform level and power analysis of the above test.

**Theorem 3.12.** Suppose conditions in Theorem 3.11 hold. Then we have

$$\lim_{n\to\infty}\sup_{\boldsymbol{\beta}^*\in\mathcal{B}_0}\mathbb{P}(\Psi_j = 1) = \alpha \qquad \text{and} \qquad \liminf_{n\to\infty}\frac{\sup_{\boldsymbol{\beta}^*\in\mathcal{B}_a}\mathbb{P}(\Psi_j = 1)}{G_n(\alpha,\gamma)} \ge 1, \tag{3.15}$$

where

$$G_n(\alpha,\gamma) = 1 - \Phi\left(-an^{1/2-\gamma}\theta_K\sigma_K^{-1}[\boldsymbol{\Sigma}^{-1}]_{j,j}^{-1/2} - t_{\alpha/2}\right) - \Phi\left(-an^{1/2-\gamma}\theta_K\sigma_K^{-1}[\boldsymbol{\Sigma}^{-1}]_{j,j}^{-1/2} + t_{\alpha/2}\right),$$

and $t_{\alpha/2} = \Phi^{-1}(1-\alpha/2)$.

The proof uses the result in Theorem 3.11 and follows similarly as that of Theorem 3.5 in Javanmard and Montanari (2014), hence is omitted here. The first inequality in (3.15) shows that the type-one error of the test $\Psi_j$ is controlled by $\alpha$ uniformly over the composite null hypothesis. The second inequality in (3.15) establishes a lower bound on the power of $\Psi_j$. Note that when $\gamma < 1/2$, $\lim_{n\to\infty} G_n(\alpha,\gamma) = \alpha$, which is the trivial power obtained by randomly rejecting the null hypothesis. Therefore, the null and alternative hypotheses are indistinguishable. When $\gamma > 1/2$, $\lim_{n\to\infty} G_n(\alpha,\gamma) = 1$, that is, the power goes to one as the sample size tends to infinity. When $\gamma = 1/2$, the power goes to a constant between $\alpha$ and 1 and can be computed by

$$\lim_{n\to\infty} G_n(\alpha,\gamma) = 1 - \Phi\left(-a\theta_K\sigma_K^{-1}\Omega_j^{-1/2} - t_{\alpha/2}\right) - \Phi\left(-a\theta_K\sigma_K^{-1}\Omega_j^{-1/2} + t_{\alpha/2}\right),$$

where $\Omega_j = \lim_{p\to\infty}[\boldsymbol{\Sigma}^{-1}]_{j,j}$.

## 3.5 Simultaneous Test with Multiplier Bootstrap

The framework in the previous section is only for inference with a fixed subset of coordinates. In this section, we consider a *high-dimensional* simultaneous test formulated as follows:

$$H_0 : \boldsymbol{\beta}_{\mathcal{G}}^* = \boldsymbol{\beta}_{0,\mathcal{G}} \qquad \text{v.s.} \qquad H_1 : \boldsymbol{\beta}_{\mathcal{G}}^* \neq \boldsymbol{\beta}_{0,\mathcal{G}}, \tag{3.16}$$

where $\mathcal{G} \subset \{1, \ldots, p\}$, and $|\mathcal{G}|$ is allowed to grow exponentially fast with $n$ (or even as large as $p$). For this test, we employ the theory of Gaussian approximation and the multiplier bootstrap. As mentioned in §3.2, to simultaneously test the entire high-dimensional vector $\boldsymbol{\beta}_{0,\mathcal{G}}$, we need to estimate the whole precision matrix $\boldsymbol{\Sigma}^{-1}$. Hence we need to modify the estimator $\mathbf{M}$. For $j = 1, \ldots, p$, we solve

$$\widehat{\boldsymbol{\mu}}'_j = \underset{\boldsymbol{\mu}_j \in \mathbb{R}^p}{\operatorname{argmin}} \|\boldsymbol{\mu}_j\|_1 \tag{3.17}$$
$$\text{s.t.} \quad \big\|\widehat{\boldsymbol{\Sigma}}\boldsymbol{\mu}_j - \boldsymbol{e}_j\big\|_\infty \leq \gamma_1, \ \big\|\mathbb{X}\boldsymbol{\mu}_j\big\|_\infty \leq \gamma_2, \ \Big|\frac{1}{\sqrt{n}}\sum_{i=1}^n \boldsymbol{\mu}_j^T \boldsymbol{X}_i\Big| \leq \gamma_3.$$

and let $\mathbf{M}' = [\widehat{\boldsymbol{\mu}}'_1, \ldots, \widehat{\boldsymbol{\mu}}'_p]^T$. This is a modification of the CLIME estimator defined in Cai et al. (2011), with the addition of the second and third constraints as extra conditions on the design. Let

$$\widetilde{\boldsymbol{\beta}}^d = \widehat{\boldsymbol{\beta}} - \widetilde{\boldsymbol{\Theta}}\widehat{\boldsymbol{\kappa}}, \qquad \text{where} \qquad \widetilde{\boldsymbol{\Theta}} = \widehat{\boldsymbol{\theta}}_K^{-1}\mathbf{M}', \tag{3.18}$$

and define $T_{\mathcal{G}} := \max_{j \in \mathcal{G}} \sqrt{n}\left(\widetilde{\beta}^d_j - \beta_{0,j}\right)$. We approximate the distribution of $T_{\mathcal{G}}$ using the multiplier bootstrap. Let $\{g_i\}_{i=1}^n$ be a sequence of i.i.d. $N(0,1)$ random variables. Let $U_{\mathcal{G}} := \max_{j \in \mathcal{G}} \frac{1}{\sqrt{n}} \sum_{i=1}^n \sigma_K \widehat{\theta}_K^{-1} \widehat{\boldsymbol{\mu}}_j'^T \boldsymbol{X}_i g_i$, and $c_\alpha := \inf\{t \in \mathbb{R} : \mathbb{P}(U_{\mathcal{G}} > t \,|\, \mathbb{X}) \leq \alpha\}$. Define the test

$$\Psi_h = \mathbb{1}\{T_{\mathcal{G}} > c_\alpha\}.$$

In the following theorem, let $d := |\mathcal{G}|$ be the cardinality of the set $\mathcal{G}$ containing the indices of coordinates to be tested.

**Theorem 3.13.** Suppose Assumptions 3.1, 3.5, 3.6 hold. Moreover, suppose $\|\boldsymbol{\Sigma}^{-1}\|_{1,\max} \leq R$ and $s_1 := \max_j \|[\boldsymbol{\Sigma}^{-1}]_{j,\cdot}\|_0$. Consider the de-biased estimator $\widetilde{\boldsymbol{\beta}}^d$ defined by (3.18), where $\widehat{\boldsymbol{\beta}}, \widehat{b}_1, \ldots, \widehat{b}_K$ satisfy the conditions in Theorem 3.2, and $\widetilde{\boldsymbol{\Theta}} = \widehat{\theta}_K \mathbf{M}'$, for some $\widehat{\theta}_K$ satisfying (2.7) and $\mathbf{M}'$ solved by (3.17). We choose $\gamma_1 = a_1 R\sqrt{(\log p)/n}$ for some constant $a_1 = 2\max\{16, 36\sigma_x^4\}(5 + 4e^2 \max\{4, 6\sigma_x^2\})^2$, and let $\gamma_2$ and $\gamma_3$ be chosen as in Proposition 3.7. Then under scaling conditions $s = o(n^{1/3}/\log^{5/3}(p \vee n))$, $r_t\sqrt{s}\log(p \vee n) = o(1)$, $s_1 = o(n^{1/2}/\log^{5/2}(p \vee n))$ and $(\log(dn))^7/n \leq C_1 n^{-c_1}$ for some constants $c_1, C_1 > 0$, we have

$$\sup_{\alpha \in (0,1)} \big|\mathbb{P}(T_{\mathcal{G}} > c_\alpha) - \alpha\big| = o(1).$$

Theorem 3.13 shows that the quantiles of $T_{\mathcal{G}}$ can be well-approximated by the conditional quantiles of the bootstrapped statistic $U_{\mathcal{G}}$, hence the test $\Psi_h$ is valid. To prove this result, we need to impose a sparsity condition and a bound on the $L_1$-norm of each row of $\boldsymbol{\Sigma}^{-1}$, so that we can consistently estimate $\boldsymbol{\Sigma}^{-1}$ under the regime where $p$ grows much faster than $n$.

# 4 Choices of the First-Stage Estimators

As stated in Theorem 3.2, we require the first-stage estimators to possess desirable non-asymptotic properties. In this section, we provide several estimators that satisfy this requirement.

The most popular estimator for high-dimensional linear model is the Lasso estimator. Indeed, the Lasso has desired convergence rate and empirical sparsity. However, the conditions required to achieve such results include stringent assumptions on the noise term, for example, sub-Gaussianity or existence of higher order moments. If one believes that such assumptions are satisfied, then Lasso can be chosen as the first-stage estimator for the de-biasing procedure with composite quantile loss. In this section we focus our attention on robust estimators that do not require strong distributional assumptions on the noise term.

### 4.1 High-dimensional Penalized Composite Quantile Regression Estimators

Under our de-biasing framework, the most natural first-stage estimators are the ones solved by the penalized composite quantile regression $\widehat{\boldsymbol{\beta}}, \widehat{b}_1, \ldots, \widehat{b}_K$ defined in (2.3). In this subsection, we establish finite-sample results for these estimators. In particular, Theorem 4.3 shows that with a properly chosen tuning parameter $\lambda$, $\|\widehat{\boldsymbol{\beta}} - \boldsymbol{\beta}^*\|_2$ converges to 0 at the minimax optimal rate $\sqrt{s(\log p)/n}$. In addition, $\|\widehat{\boldsymbol{b}} - \boldsymbol{b}^*\|_2$ converges at the rate $\sqrt{sK(\log p)/n}$. Theorem 4.5 shows that the empirical sparsity $\widehat{s} = \|\widehat{\boldsymbol{\beta}}\|_0$ is of the same order as the true sparsity level $s$.

The finite-sample results demonstrate that the high-dimensional penalized CQR estimator has comparable estimation performance as the Lasso, but requires much weaker assumptions on the noise distribution. Another advantage of the penalized CQR estimator is that it simultaneously estimates all the quantiles $b_k^*$'s for $k = 1, \ldots, K$.

In order to establish the $L_2$ convergence rate of the estimators $\widehat{\boldsymbol{\beta}}$ and $\widehat{\boldsymbol{b}}$, we require the following two general assumptions on the design matrix. We later verify that these conditions are satisfied under the same conditions given in previous sections. We first define the following notations:

$$\mathbf{S} := \sum_{k=1}^{K} f_k^* \begin{pmatrix} \boldsymbol{\Sigma} & \mathbf{0} \\ \mathbf{0} & \mathrm{diag}(\mathbf{e}_k) \end{pmatrix} \in \mathbb{R}^{(p+K)\times(p+K)},$$

where $\mathbf{e}_k \in \mathbb{R}^K$ is a vector with all zero entries except the $k$-th entry is 1. By the definition of $\mathbf{S}$, we have that for any $\underset{\sim}{\boldsymbol{\beta}} := (\boldsymbol{\beta}^T, \boldsymbol{b}^T)^T \in \mathbb{R}^{p+K}$, there must be $\|\underset{\sim}{\boldsymbol{\beta}}\|_{\mathbf{S}}^2 = \sum_{k=1}^K f_k^*(\boldsymbol{\beta}^T\boldsymbol{\Sigma}\boldsymbol{\beta} + b_k^2)$. In addition, we denote $\mathcal{T} := \{j \in (1, \ldots, p) : \beta_j^* \neq 0\}$ to be the true support of $\boldsymbol{\beta}^*$, and

$$\mathcal{A} := \left\{ \underset{\sim}{\boldsymbol{\beta}} = (\boldsymbol{\beta}^T, \boldsymbol{b}^T)^T \in \mathbb{R}^{p+K} : \boldsymbol{\beta} \in \mathbb{R}^p, \boldsymbol{b} \in \mathbb{R}^K, \|\boldsymbol{\beta}_{\mathcal{T}^c}\|_1 \le 3\|\boldsymbol{\beta}_{\mathcal{T}}\|_1 + \|\boldsymbol{b}\|_1/K \right\}.$$

**Assumption 4.1.** There exists a constant $m_0$ possibly depending on $K$ such that for all $\underset{\sim}{\boldsymbol{\beta}} = (\boldsymbol{\beta}^T, b_1, \ldots, b_K)^T \in \mathbb{R}^{p+K}, \underset{\sim}{\boldsymbol{\beta}} \in \mathcal{A}$, we have

$$\sum_{k=1}^{K} \mathbb{E}\left[|\boldsymbol{X}^T\boldsymbol{\beta} + b_k|^3\right] \le m_0 \|\underset{\sim}{\boldsymbol{\beta}}\|_{\mathbf{S}}^3.$$

When Assumptions 3.1 and 3.6 are satisfied, a sufficient condition for Assumption 4.1 is the boundedness of the third moment of $\boldsymbol{u}^T\boldsymbol{X}$ for any unit vector $\boldsymbol{u} \in \mathbb{R}^p$. For example, the sub-Gaussian design satisfies this condition (See Proposition 4.7).

Under Assumption 4.1, we show that the expectation of the empirical composite quantile loss function can be lower bounded by a quadratic form when the estimator is close to the true parameter, that is, it has a local curvature, as mentioned in Section 1. This enables us to obtain the optimal convergence rate for high-dimensional CQR estimator, even though the objective function is not strongly convex. For complete technical details, see the proof of Theorem 4.3 in Appendix C.

**Assumption 4.2.** With probability at least $1 - \delta_n$, where $\delta_n \to 0$ as $n \to 0$, we have

$$\max_{1 \le j \le p} \frac{1}{n} \sum_{i=1}^{n} X_{ij}^2 \le \widetilde{\sigma}_x^2.$$

Assumption 4.2 requires the covariates of $\boldsymbol{X}$ to be well-behaved, in the sense that the empirical second moment is bounded with high probability.

We are now ready to present the result on the rates of convergence of $\widehat{\boldsymbol{\beta}}$ and $\widehat{\boldsymbol{b}}$.

**Theorem 4.3.** Suppose Assumptions 3.1, 3.6, 4.1 and 4.2 hold. Let $\widehat{\underset{\sim}{\boldsymbol{\beta}}} = (\widehat{\boldsymbol{\beta}}^T, \widehat{\boldsymbol{b}}^T)^T \in \mathbb{R}^{p+K}$ be the solution to the penalized minimization problem (2.3). We choose $\lambda$ such that

$$\lambda = 4K\zeta \max\{\widetilde{\sigma}_x, 1\}\sqrt{(\log p)/n}$$

for some $\zeta \ge 1$. If $n, p, s$ satisfy the scaling condition:

$$\sqrt{\frac{s \log p}{n}} < \frac{3}{8C'_+ m_0 C_0 \sqrt{K}}, \tag{4.1}$$

where $m_0$ is specified in Assumption 4.1, and $C_0 = \max\left\{\sqrt{6}, 32\widetilde{\sigma}_x\right\}(4\rho_{\min}^{-1/2} + 2)C_-^{-1/2}(1 + 4\zeta)$, then with probability at least $1 - 4Kp^{-7} - 6p^{-3} - 3\delta_n$, we have

$$\|\widehat{\underset{\sim}{\boldsymbol{\beta}}} - \underset{\sim}{\boldsymbol{\beta}}^*\|_{\mathbf{S}} \le 4C_0\sqrt{K}\sqrt{\frac{s \log p}{n}},$$

which further implies that

$$\|\widehat{\boldsymbol{\beta}} - \boldsymbol{\beta}^*\|_2 \le \frac{4C_0}{\sqrt{C_- \rho_{\min}}}\sqrt{\frac{s \log p}{n}}, \qquad \text{and} \qquad \|\widehat{\boldsymbol{b}} - \boldsymbol{b}^*\|_2 \le \frac{4C_0}{\sqrt{C_-}}\sqrt{K}\sqrt{\frac{s \log p}{n}}.$$

Theorem 4.3 shows that the penalized CQR estimator $\widehat{\boldsymbol{\beta}}$ achieves the minimax optimal rate, under the choice of $\lambda = 4K\zeta \max\{\widetilde{\sigma}_x, 1\}\sqrt{(\log p)/n}$. The constant $\zeta$ does not affect the convergence rate, and only alters the constant $C_0$. The smallest $C_0$ is achieved when $\zeta = 1$, that is, choosing $\lambda = 4K \max\{\widetilde{\sigma}_x, 1\}\sqrt{(\log p)/n}$. However, to simultaneously obtain empirical sparsity of $\widehat{\boldsymbol{\beta}}$ at the same order of $s$, Theorem 4.5 below requires taking a constant $\zeta > 1$. Note that the $L_2$-convergence of $\widehat{\boldsymbol{b}}$ is slower than that of $\widehat{\boldsymbol{\beta}}$ by approximately a $\sqrt{K}$ factor. This is intuitive as we expect each of $\widehat{b}_k$ to converge at the same rate as $\widehat{\boldsymbol{\beta}}$.

To establish the sparsity and mean square error rate results, we require one more condition on the design. Define

$$\psi(q) = \sup_{\|\boldsymbol{\beta}\|_0 \le q} \frac{\boldsymbol{\beta}^T \widehat{\boldsymbol{\Sigma}} \boldsymbol{\beta}}{\|\boldsymbol{\beta}\|_2^2}. \tag{4.2}$$

**Assumption 4.4.** There exists a constant $\psi_0$ such that $\psi(n/\log(p \vee n)) \le \psi_0$ with probability at least $1 - d_n$, where $d_n \to 0$ as $n \to \infty$.

This is the the sparse eigenvalue (SE) condition. We present the empirical sparsity result in the following theorem.

**Theorem 4.5.** Suppose Assumptions 3.1, 3.6, 4.1, 4.2 and 4.4 and the scaling condition (4.1) hold. Let $C_S = \widetilde{\sigma}_x + \psi_0^{1/2} + c_0\psi_0^{1/2} \geq \widetilde{\sigma}_x$ where $c_0$ is a universal constant. We choose $\lambda = 4KC_S\sqrt{(\log p)/n}$, then with probability at least $1 - 6Kp^{-7} - 8p^{-3} - 4\delta_n - d_n$, we have

$$\widehat{s} \leq \frac{48\rho_{\max}C_+C_0^2}{C_-C_S^2}s,$$

where $C_0$ is the constant specified in Theorem 4.3.

By the sparse eigenvalue condition, the mean square error rate of $\widehat{\boldsymbol{\beta}}$ follows from the $L_2$ convergence rate and empirical sparsity.

**Corollary 4.6.** Suppose conditions in Theorem 4.3 and 4.5 hold. In addition, suppose

$$\frac{48\rho_{\max}C_+C_0^2}{C_-C_S^2}s \leq \frac{n}{\log(p \vee n)}.$$

Then with probability at least $1 - 6Kp^{-7} - 8p^{-3} - 4\delta_n - d_n$, we have

$$\mathrm{MSE}(\widehat{\boldsymbol{\beta}}) = \|\mathbb{X}(\widehat{\boldsymbol{\beta}} - \boldsymbol{\beta}^*)/\sqrt{n}\|_2 \leq \frac{4\psi_0^{1/2}C_0}{\sqrt{C_-\rho_{\min}}}\sqrt{\frac{s\log p}{n}}.$$

Theorems 4.3, 4.5 and Corollary 4.6 show that the estimation of $\boldsymbol{\beta}^*$ using penalized CQR is robust and minimax optimal. In particular, $\widehat{\boldsymbol{\beta}}$ and $\widehat{\boldsymbol{b}}$ achieve the desired finite-sample requirements in order to establish the asymptotic conclusions in Theorems 3.10 and 3.11, and to construct uniformly valid confidence intervals and test hypotheses for $\boldsymbol{\beta}^*$ based on the de-biased estimator $\widehat{\boldsymbol{\beta}}^d$.

In the rest of the section, we show that the general Assumptions 4.1, 4.2 and 4.4 are satisfied under conditions stated in the previous sections. In particular, the sub-Gaussian design satisfies all three assumptions.

**Proposition 4.7.** Suppose Assumptions 3.1, 3.5 and 3.6 hold. Then we have

(i) Assumption 4.1 holds with $m_0 = C_-^{3/2}\min\{\rho_{\min}^{3/2}, 1\}/(3\max\{8^{5/4}\sigma_x^{3/2}, 1\})$;

(ii) When $\log p \leq n/2$, Assumption 4.2 holds with $\delta_n = 2\exp(-n/2)$ and $\widetilde{\sigma}_x = c\sigma_x$ for some universal constant $c$;

(ii) Assumption 4.4 holds with $d_n = \exp(-c_1 n)$ and $\psi_0, c_1$ being some constants that only depends on $\sigma_x$.

Therefore, applying Theorem 4.3, 4.5 and Corollary 4.6, we obtain the following finite-sample conclusions for the high-dimensional CQR estimator $\widehat{\boldsymbol{\beta}}$ and $\widehat{\boldsymbol{b}}$ under the sub-Gaussian design.

**Corollary 4.8.** Suppose Assumptions 3.1, 3.5 and 3.6 hold. Let $\widehat{\boldsymbol{\beta}}$ be the solution to the minimization problem (2.3) with the choice of $\lambda$ satisfying conditions in Theorems 4.3 and 4.5. Let $s = s(n)$ satisfy the scaling condition $s = o(n/\log(p \vee n))$ and let $\widehat{s} = \|\widehat{\boldsymbol{\beta}}\|_0$. Then we have

$$\lim_{n\to\infty} \sup_{\boldsymbol{\beta}^* \in \mathbb{R}^p, \|\boldsymbol{\beta}^*\|_0 \leq s} \mathbb{P}\big(\mathcal{E}_1(\boldsymbol{\beta}^*)^c \cup \mathcal{E}_2(\boldsymbol{\beta}^*)^c \cup \mathcal{E}_3(\boldsymbol{\beta}^*)^c \cup \mathcal{E}_4(\boldsymbol{\beta}^*)^c\big) = 0$$

where $\mathcal{E}_1(\boldsymbol{\beta}^*), \ldots, \mathcal{E}_4(\boldsymbol{\beta}^*)$ are defined in §3.4.

The conclusions in Corollary 4.8 exactly meet the finite-sample requirements in Theorems 3.10 and 3.11, hence the confidence regions constructed by de-biased penalized composite quantile regression estimator is uniformly valid for high-dimensional linear models.

## 4.2 Single Quantile Regression

There are two ways to obtain the first-stage estimator using single quantile regression. The most direct method is to solve

$$(\widehat{\boldsymbol{\beta}}_k, \widehat{\boldsymbol{b}}_k) \in \underset{\boldsymbol{\beta}\in\mathbb{R}^p, b\in\mathbb{R}}{\operatorname{argmin}} \frac{1}{n}\sum_{i=1}^n \phi_{\tau_k}(Y_i - \boldsymbol{X}_i^T\boldsymbol{\beta} - b) + \lambda\|\boldsymbol{\beta}\|_1, \tag{4.3}$$

for $k = 1, \ldots, K$. The final $\widehat{\boldsymbol{\beta}}$ can be chosen from any of the estimated $\widehat{\boldsymbol{\beta}}_k$'s, or we can take the average of all the $\widehat{\boldsymbol{\beta}}_k$'s, that is, $\widehat{\boldsymbol{\beta}} = K^{-1}\sum_{k=1}^K \widehat{\boldsymbol{\beta}}_k$. The second method requires less computation: we pick a single $\tau$ and minimize the penalized $\tau$-quantile regression. For example, when $\tau = 1/2$, it becomes the penalized least absolute deviation regression. If in addition we have the prior knowledge that the median of $\epsilon$ is zero, we solve

$$\widehat{\boldsymbol{\beta}} \in \underset{\boldsymbol{\beta}\in\mathbb{R}^p}{\operatorname{argmin}} \frac{1}{2n}\sum_{i=1}^n |Y_i - \boldsymbol{X}_i^T\boldsymbol{\beta}| + \lambda\|\boldsymbol{\beta}\|_1, \tag{4.4}$$

and the $\widehat{b}_k$'s can be obtained by finding $b_k$ such that $\tau_k = \sum_{i=1}^n \mathbb{1}\{Y_i \le \boldsymbol{X}_i^T\widehat{\boldsymbol{\beta}} + b_k\}$, that is, the the $\tau_k$-quantile of $\{\widehat{\epsilon}_i := Y_i - \boldsymbol{X}_i^T\widehat{\boldsymbol{\beta}}\}_{i=1}^n$. Note that in the first method, we have $\tau_k = \sum_{i=1}^n \mathbb{1}\{Y_i \le \boldsymbol{X}_i^T\widehat{\boldsymbol{\beta}}_k + \widehat{b}_k\}$ by the first order condition. The second method essentially replaces $\widehat{\boldsymbol{\beta}}_k$ in (4.3) for each $k$ by the single $\widehat{\boldsymbol{\beta}}$ in (4.4) in calculating $\widehat{b}_k$, hence reduces computations.

Belloni and Chernozhukov (2011) studied the finite-sample convergence rate and empirical sparsity of the general penalized quantile regression in high dimensions, where the conditional $u$-quantile of $Y \mid X$ has the form $\boldsymbol{X}^T\boldsymbol{\beta}(u)$. Applying their result, it can be verified that under our Assumptions 3.1, 3.5 and 3.6, the penalized quantile regression estimators in (4.3) have desired finite-sample properties required by Theorems 3.10 and 3.11. Thus it can be used as the first-stage estimator in our de-biasing framework.

## 4.3 Truncated Single or Composite Quantile Regression

There are two concerns for obtaining empirical sparsity using the estimator $\widehat{\boldsymbol{\beta}}$ in §4.1. The first is that $\widehat{s}$ is only equal to $s$ up to a constant. If the constant is large, then $\widehat{\boldsymbol{\beta}}$ may not be very sparse. The second is that we require a delicate choice of the tuning parameter $\lambda$ for obtaining empirical sparsity, as shown in Theorem 4.5. On the contrary, to achieve the desired convergence rate of the penalized CQR estimator, the $\lambda$ in (2.3) can be tuning-free (for example, if we normalize $\mathbb{X}$ so that the column $\|\cdot\|_2$-norms are less than $\sqrt{n}$). The same consideration applies to the penalized quantile regression (4.3) in §4.2 (Belloni and Chernozhukov, 2011).

In this section, we describe a truncation procedure for obtaining a valid first-stage estimator. It can be shown that the convergence bound of the truncated estimator is enlarged by a small multiplicative constant compared to that of the estimator before truncation, while obtaining an exact sparsity of $s$ for all $s \ge \|\boldsymbol{\beta}^*\|_0$.

Specifically, let $\widehat{\boldsymbol{\beta}}'$ be the penalized single or composite quantile regression estimator. We define $\widehat{\boldsymbol{\beta}}$ by taking the $s$-largest entries of $\widehat{\boldsymbol{\beta}}'$ and letting all other entries to be 0. The $L_2$-convergence rate of $\widehat{\boldsymbol{\beta}}$ can be simply controlled as follows: using triangular inequality, we have $\|\widehat{\boldsymbol{\beta}} - \boldsymbol{\beta}^*\|_2 \le \|\widehat{\boldsymbol{\beta}} - \widehat{\boldsymbol{\beta}}'\|_2 + \|\widehat{\boldsymbol{\beta}}' - \boldsymbol{\beta}^*\|_2$. By the definition of $\widehat{\boldsymbol{\beta}}$, we have $\|\widehat{\boldsymbol{\beta}} - \widehat{\boldsymbol{\beta}}'\|_2 = \min_{\|\boldsymbol{\beta}\|_0 \le s} \|\boldsymbol{\beta} - \widehat{\boldsymbol{\beta}}'\|_2 \le \|\boldsymbol{\beta}^* - \widehat{\boldsymbol{\beta}}'\|_2$, so we conclude $\|\widehat{\boldsymbol{\beta}} - \boldsymbol{\beta}^*\|_2 \le 2\|\widehat{\boldsymbol{\beta}}' - \boldsymbol{\beta}^*\|_2$. Now that the $L_2$-error can be controlled, we have

$\mathrm{MSE}(\widehat{\boldsymbol{\beta}}) \le \sqrt{\psi(s)}\|\widehat{\boldsymbol{\beta}} - \boldsymbol{\beta}^*\|_2 \le 2\sqrt{\psi(s)}\|\widehat{\boldsymbol{\beta}}' - \boldsymbol{\beta}^*\|_2$. Therefore, the non-asymptotic properties of $\widehat{\boldsymbol{\beta}}$ satisfy the requirements in Theorem 3.2. In particular, the sparsity of $\widehat{\boldsymbol{\beta}}$ is exactly $s$. We summarize the results for truncated CQR estimator in the following corollary.

**Corollary 4.9.** Suppose Assumptions 3.1, 3.5 and 3.6 hold. Let $\widehat{\boldsymbol{\beta}}$ be the estimator obtained by truncating the $s$ largest entries of $\widehat{\boldsymbol{\beta}}'$, where $\widehat{\boldsymbol{\beta}}'$ is solved by (2.3), with $\lambda = 4K\max\{\widetilde{\sigma}_x, 1\}\sqrt{(\log p)/n}$. Then with probability at least $1 - 4Kp^{-7} - 6p^{-3} - 3\exp(-cn)$, we have

$$\|\widehat{\boldsymbol{\beta}} - \boldsymbol{\beta}^*\|_2 = O\Big(\sqrt{\frac{s\log p}{n}}\Big), \quad \|\widehat{\boldsymbol{b}} - \boldsymbol{b}^*\|_2 = O\Big(\sqrt{K}\sqrt{\frac{s\log p}{n}}\Big), \quad \text{and} \quad \mathrm{MSE}(\widehat{\boldsymbol{\beta}}) = O\Big(\sqrt{\frac{s\log p}{n}}\Big).$$

Since $s$ is usually unknown, it is a tuning parameter. In practice, it can be chosen using various methods, for example, the stability-based approach StARS (Meinshausen and Bühlmann, 2010; Liu et al., 2010; Shah and Samworth, 2013).

### 4.4 Post-Selection Estimator

Many other procedures can be used in obtaining the first stage-estimator. For example, we can use the post-selection estimators that first recover the support using the penalized quantile regression, and then fit the low-dimensional quantile regression estimator on the estimated support. Belloni and Chernozhukov (2011) show that the post-$L_1$-quantile regression estimator has the same rate as the penalized quantile regression estimator and under stronger conditions improves on the rate. Nonconvex penalties can be used as well. For example, Wang et al. (2012) studied oracle properties of the estimator with SCAD penalty in high-dimensional quantile regression.

## 5 Massive Sample Size: the Divide-and-Conquer Algorithm

When the dimension $p$ and the sample size $n$ both get extremely large, the computational cost to obtain the estimators $\widehat{\boldsymbol{\beta}}$ and $\mathbf{M}$ becomes intractable a single computer. Sometimes it is even not feasible to store all the data on one computer, or at least keep them in memory. Thus, we need computationally more efficient algorithms which not only reduce the computation time and complexity, but also allow for distributed computing and minimal communication. We introduce a divide-and-conquer procedure for de-biased CQR inference under the linear model.

Suppose we are given data $(Y_i, \boldsymbol{X}_i)$ with size $N$ and input dimension $p$. Both quantities could potentially be extremely large. The proposed divide-and-conquer algorithm is described as follows:

- Randomly split the sample into $m$ disjoint subsets $\mathcal{D}_1, \ldots, \mathcal{D}_m$. We assume that the all the splits are of the same size $n$, that is, $N = nm$.
- On subsample $\mathcal{D}_\ell$, obtain estimates $\widehat{\boldsymbol{\beta}}^{(\ell)}$ and $\widehat{b}_k^{(\ell)}$ for $k = 1, \ldots, K$. In addition, solve $\mathbf{M}^{(\ell)}$ by (2.5) and (2.6), and obtain $\widehat{\theta}_K^{(\ell)}$ that satisfies (2.7). Let

$$\widehat{\boldsymbol{\beta}}^d(\ell) = \widehat{\boldsymbol{\beta}}^{(\ell)} + \widehat{\boldsymbol{\Theta}}^{(\ell)}\widehat{\boldsymbol{\kappa}}^{(\ell)} \tag{5.1}$$

where $\widehat{\boldsymbol{\Theta}}^{(\ell)} = \widehat{\theta}_K^{(\ell)}\mathbf{M}^{(\ell)}$, and $\widehat{\boldsymbol{\kappa}}^{(\ell)} = \sum_{k=1}^K n^{-1}\sum_{i\in\mathcal{D}_\ell}\big(\mathbb{1}\{Y_i \le \boldsymbol{X}_i^T\widehat{\boldsymbol{\beta}}^{(\ell)} + \widehat{b}_k^{(\ell)}\} - \tau_k\big)\boldsymbol{X}_i$.

- Compute the aggregated estimator

$$\bar{\boldsymbol{\beta}}^d = \frac{1}{m}\sum_{\ell=1}^{m}\widehat{\boldsymbol{\beta}}^d(\ell). \tag{5.2}$$

We allow the number of subsets $m$ to grow with the sample size $N$. In the following, we show that the divide-and-conquer procedure preserves the asymptotic properties of the de-biased estimator when $m$ does not grow too fast. We first show that the optimization problem (2.5) is feasible on all $m$ data splits for suitable choices of $\gamma_1$, $\gamma_2$ and $\gamma_3$.

**Proposition 5.1** (Feasibility for the divide-and-conquer estimator)**.** Suppose Assumptions 3.5 and 3.6 hold. Take $\gamma_1 = a_1\sqrt{(\log p)/n}$, $\gamma_2 = a_2\sqrt{\log(p \vee n)}$ and $\gamma_3 = a_3\sqrt{\log(p \vee n)}$ for constants $a_1 = 12e\sigma_x\sqrt{2\rho_{\max}/\rho_{\min}}$, $a_2 = 2\sqrt{3}\sigma_x/\rho_{\min}$ and $a_3 = 2\sqrt{2}\sigma_x/\rho_{\min}$. Let $\mathbb{X}^{(\ell)}$ and $\widehat{\boldsymbol{\Sigma}}^{(\ell)}$ be the design and sample covariance matrix of $\boldsymbol{X}$ obtained based on data split $\mathcal{D}_\ell$. Then with probability at least $1 - m(p \vee n)^{-7}$, we have

$$\max_{1\le\ell\le m}\big\|\widehat{\boldsymbol{\Sigma}}^{(\ell)}\boldsymbol{\Sigma}^{-1} - \mathbf{I}_p\big\|_{\max} \le \gamma_1, \quad \max_{1\le\ell\le m}\big\|\mathbb{X}^{(\ell)}\boldsymbol{\Sigma}^{-1}\big\|_{\max} \le \gamma_2, \quad \max_{1\le\ell\le m}\Big\|\frac{1}{\sqrt{n}}\sum_{i\in\mathcal{D}_\ell}\boldsymbol{\Sigma}^{-1}\boldsymbol{X}_i\Big\|_\infty \le \gamma_3,$$

that is, $\boldsymbol{\Sigma}^{-1}$ is a feasible solution to (2.5) for all $j \in \{1,\ldots,p\}$ and $\ell \in \{1,\ldots,m\}$.

The proof follows similarly as that of Proposition 3.7 coupled with the application of union bound, so is omitted. For simplicity of analysis, we consider the truncated composite quantile regression estimator discussed in §4.3. This estimator has nice properties that it has exact sparsity of $s$ and desired convergence rate as shown in Corollary 4.9. The following theorem shows that when the number of subsets $m$ does not grow too fast with sample size $N$, the divide-and-conquer estimator is asymptotically normal.

**Theorem 5.2.** Suppose Assumptions 3.1, 3.5 and 3.6 hold. Consider the divide-and-conquer estimator $\bar{\boldsymbol{\beta}}^d$ defined as per (5.2), where $\widehat{\boldsymbol{\beta}}^d(\ell)$ is the de-biased estimator solved on $\mathcal{D}_\ell$, and $\gamma_1, \gamma_2$ and $\gamma_3$ in (2.5) are chosen as the ones in Theorem 5.1. When the number of subsets satisfy

$$m = o\left(\frac{N^{1/3}}{s\log^{5/3}(p \vee N)}\right) \quad \text{and} \quad m = o(p^3), \tag{5.3}$$

we have

$$\frac{\sqrt{N}(\bar{\beta}^d_j - \beta^*_j)}{\big(m^{-1}\sum_{\ell=1}^m \widehat{\boldsymbol{\mu}}_j^{(\ell)T}\widehat{\boldsymbol{\Sigma}}^{(\ell)}\widehat{\boldsymbol{\mu}}_j^{(\ell)}\big)^{1/2}} \to N(0, \sigma_K^2\theta_K^{-2}).$$

Theorem 5.2 shows that the divide-and-conquer estimator remains asymptotically normal, even when the number of subsets $m$ is allowed to grow with the sample size $N$. In particular the growth rate of $m$ needs to satisfy the conditions specified in (5.3). Therefore, we can construct confidence intervals and test hypotheses based on the divide-and-conquer procedure. The divide-and-conquer estimator has significant computational advantage: we only need to solve $m$ regression problems with dimension $p \times n$, which is much faster than solving the original $p \times N$ problem, since the complexity of solving the penalized regression is often $O(N^2)$ or higher. Moreover, when parallel computing is available, we can distribute the computations to $m$ machines and further reduce the computation time. In particular, it requires only a single communication to each machine for averaging the estimators obtained on each data split.

A more interesting question is what is the trade-off in asymptotic variance as the result of computational gains. The following proposition shows that the divide-and-conquer estimator achieves the same asymptotic variance as the low-dimensional CQR regression.

**Proposition 5.3.** Suppose conditions in Theorem 5.2 hold. Then we have for any $j = 1, \ldots, p$,

$$\limsup_{N\to\infty} \frac{1}{m}\sum_{\ell=1}^{m} \widehat{\boldsymbol{\mu}}_j^{(\ell)T}\widehat{\boldsymbol{\Sigma}}^{(\ell)}\widehat{\boldsymbol{\mu}}_j^{(\ell)} - [\boldsymbol{\Sigma}^{-1}]_{jj} \le 0.$$

We next turn to the hypothesis test (3.15). The test statistic based on the divide-and-conquer estimator is defined as follows:

$$\bar{\Psi}_j = \begin{cases} 0, & \text{if } |\bar{\beta}_j^d| \le \Phi^{-1}(1-\alpha/2)\sigma_K\big(m^{-1}\sum_{\ell=1}^m \widehat{\boldsymbol{\mu}}_j^{(\ell)T}\widehat{\boldsymbol{\Sigma}}^{(\ell)}\widehat{\boldsymbol{\mu}}_j^{(\ell)}\big)^{1/2}/(\sqrt{n}\widehat{\theta}_K^{(\ell)}); \\ 1, & \text{otherwise.}\end{cases}$$

The following theorem shows that the divide-and-conquer estimator $\bar{\boldsymbol{\beta}}^d$ possesses the same asymptotic power as the estimator based on the entire sample. This is an "oracle" property, as the divide-and-conquer estimator is able to reduce computation and allows for distributed inference, while still preserving the asymptotic power of the oracle test constructed on the entire data set.

**Theorem 5.4.** Suppose conditions in Theorem 5.2 hold. Then we have

$$\lim_{n\to\infty}\sup_{\boldsymbol{\beta}^*\in\mathcal{B}_0}\mathbb{P}(\bar{\Psi}_j = 1)\le\alpha \qquad \text{and} \qquad \liminf_{n\to\infty}\frac{\sup_{\boldsymbol{\beta}^*\in\mathcal{B}_a}\mathbb{P}(\bar{\Psi}_j=1)}{G_n(\alpha,\gamma)}\ge 1, \tag{5.4}$$

where $G_n(\alpha,\gamma)$ is defined in Theorem 3.12 and $\mathcal{B}_0$ and $\mathcal{B}_a$ are defined as per (3.14).

# 6 Proof of Main Results

In this section, we provide the proof of the main inference results in Section 3, including Theorems 3.2 and 3.3 for establishing asymptotic normality of the de-biased estimator. More technical details and proofs of remaining results are deferred to the supplementary appendix. We first define the following notations: for a pair $(\boldsymbol{\beta}, \boldsymbol{b})$, where $\boldsymbol{\beta}\in\mathbb{R}^p, \boldsymbol{b}\in\mathbb{R}^K$, recall that $\underset{\sim}{\boldsymbol{\beta}} := (\boldsymbol{\beta}^T, \boldsymbol{b}^T)^T\in\mathbb{R}^{p+K}$ as defined in §4.1. We furthermore let

$$\underset{\sim}{\boldsymbol{\beta}}_k := (\boldsymbol{\beta}^T, b_k)^T\in\mathbb{R}^{p+1}, \qquad \text{for } k = 1,\ldots,K.$$

Following these rules, we can define $\underset{\sim}{\widehat{\boldsymbol{\beta}}}, \underset{\sim}{\boldsymbol{\beta}}^*$ and $\underset{\sim}{\widehat{\boldsymbol{\beta}}}_k, \underset{\sim}{\boldsymbol{\beta}}_k^*$ for $k = 1,\ldots,K$. Without causing confusions, we use $\underset{\sim}{\boldsymbol{X}}$ to denote both the $\mathbb{R}^{p+1}$ vector $(\boldsymbol{X}^T, 1)^T$ and the $\mathbb{R}^{p+K}$ vector $(\boldsymbol{X}^T, \mathbf{1}_K^T)^T$ , where $\mathbf{1}_K = (1, 1, \ldots, 1)^T\in\mathbb{R}^K$.

## 6.1 Proof of Theorem 3.2

We first sketch the main idea and then provide detailed proof. The analysis of the de-biasing factor $\widehat{\boldsymbol{\kappa}}$ lies at the heart of the proof of Theorem 3.2. Define $\underset{\sim}{\widehat{\boldsymbol{\Delta}}}_k := \underset{\sim}{\widehat{\boldsymbol{\beta}}}_k - \underset{\sim}{\boldsymbol{\beta}}_k^*$ and $\widehat{\boldsymbol{\Delta}} := \widehat{\boldsymbol{\beta}} - \boldsymbol{\beta}^*$. As $\{Y_i, \boldsymbol{X}_i\}$ satisfies the linear model, we have $Y_i \le \boldsymbol{X}_i \widehat{\boldsymbol{\beta}} + \widehat{b}_k$ is equivalent to $\epsilon_i \le \underset{\sim}{\boldsymbol{X}}_i^T \underset{\sim}{\widehat{\boldsymbol{\Delta}}}_k + b_k^*$. Therefore, by the definition of $\widehat{\boldsymbol{\kappa}}$, we have

$$\widehat{\boldsymbol{\kappa}} = \underbrace{\sum_{k=1}^{K} \frac{1}{n} \sum_{i=1}^{n} \big( \mathbb{1}\{\epsilon_i \le b_k^*\} - \tau_k \big) \boldsymbol{X}_i}_{\text{(I)}} + \underbrace{\sum_{k=1}^{K} \frac{1}{n} \sum_{i=1}^{n} \big( \mathbb{1}\{\epsilon_i \le \underset{\sim}{\boldsymbol{X}}_i^T \underset{\sim}{\widehat{\boldsymbol{\Delta}}}_k + b_k^*\} - \mathbb{1}\{\epsilon_i \le b_k^*\} \big) \boldsymbol{X}_i}_{\text{(II)}} . \tag{6.1}$$

The term (I) is a sum of i.i.d random vectors with zero mean, as $\mathbb{E}[\mathbb{1}\{\epsilon_i \le b_k^*\}] = \mathbb{P}(\epsilon_i \le b_k^*) = \tau_k$ by definition. This is the dominating term that weakly converges to a normal distribution after scaling by $\sqrt{n}$ (Theorem 3.3). Moreover, notice that $\mathbb{1}\{\epsilon_i \le b_k^*\} \overset{d}{=} \text{Bernoulli}(\tau_k)$, so (I) is a linear combination of i.i.d. Bernoulli random variables.

The analysis of term (II) is more challenging. The main difficulty is that the indicator function is not differentiable. To solve this problem, we resort to its expectation conditioned on the design $\mathbb{X}$. As $\epsilon_i$ are independent of $\mathbb{X}$, it holds that $\mathbb{E}\big[\, \mathbb{1}\{\epsilon_i \le \underset{\sim}{\boldsymbol{X}}_i^T \underset{\sim}{\boldsymbol{\Delta}}_k + b_k^*\} \boldsymbol{X}_i \,|\, \mathbb{X}\big] = \mathbb{P}\big(\epsilon_i \le \underset{\sim}{\boldsymbol{X}}_i^T \underset{\sim}{\boldsymbol{\Delta}}_k + b_k^* \,|\, \mathbb{X}\big) \boldsymbol{X}_i = F_\epsilon\big(\underset{\sim}{\boldsymbol{X}}_i^T \underset{\sim}{\boldsymbol{\Delta}}_k + b_k^*\big) \boldsymbol{X}_i$ for any $\underset{\sim}{\boldsymbol{\Delta}}_k \in \mathbb{R}^{p+1}$, and similarly $\mathbb{E}\big[\, \mathbb{1}\{\epsilon_i \le b_k^*\} \boldsymbol{X}_i \,|\, \mathbb{X}\big] = F_\epsilon(b_k^*) \boldsymbol{X}_i$. By Taylor expansion, we have

$$F_\epsilon\big(\underset{\sim}{\boldsymbol{X}}_i^T \underset{\sim}{\widehat{\boldsymbol{\Delta}}}_k + b_k^*\big) - F_\epsilon(b_k^*) = f_\epsilon(b_k^*) \underset{\sim}{\boldsymbol{X}}_i^T \underset{\sim}{\widehat{\boldsymbol{\Delta}}}_k + f_\epsilon'(\xi_{ik}) \big(\underset{\sim}{\boldsymbol{X}}_i^T \underset{\sim}{\widehat{\boldsymbol{\Delta}}}_k\big)^2$$

for some $\xi_{ik}$ between $b_k^*$ and $\underset{\sim}{\boldsymbol{X}}_i^T \underset{\sim}{\widehat{\boldsymbol{\Delta}}}_k + b_k^*$. Therefore, it holds that

$$\begin{aligned} &\sum_{k=1}^{K} \frac{1}{n} \sum_{i=1}^{n} \Big\{ F_\epsilon\big(\underset{\sim}{\boldsymbol{X}}_i^T \underset{\sim}{\widehat{\boldsymbol{\Delta}}}_k + b_k^*\big) \boldsymbol{X}_i - F_\epsilon(b_k^*) \boldsymbol{X}_i \Big\} \\ &\qquad = \underbrace{\theta_K \widehat{\boldsymbol{\Sigma}} \widehat{\boldsymbol{\Delta}}}_{\text{(i)}} + \underbrace{\sum_{k=1}^{K} f_k^* (\widehat{b}_k - b_k^*) \frac{1}{n} \sum_{i=1}^{n} \boldsymbol{X}_i}_{\text{(ii)}} + \underbrace{\sum_{k=1}^{K} \frac{1}{n} \sum_{i=1}^{n} f_\epsilon'(\xi_{ik}) \big(\underset{\sim}{\boldsymbol{X}}_i^T \underset{\sim}{\widehat{\boldsymbol{\Delta}}}_k\big)^2 \boldsymbol{X}_i}_{\text{(iii)}}, \end{aligned} \tag{6.2}$$

where in the derivation of (i) we use the fact that $n^{-1} \sum_{i=1}^{n} \boldsymbol{X}_i \boldsymbol{X}_i^T = \widehat{\boldsymbol{\Sigma}}$. From (i), we can "invert" $\theta_K \widehat{\boldsymbol{\Sigma}}$ by multiplying $\theta_K^{-1} \mathbf{M}$ to recover $\widehat{\boldsymbol{\Delta}} = \widehat{\boldsymbol{\beta}} - \boldsymbol{\beta}^*$. After this operation, (ii) is of the order $o(n^{-1/2})$ under the scaling conditions stated in the theorem. Indeed, for term (ii), we have $\sum_{k=1}^{K} f_k^* (\widehat{b}_k - b_k^*) = O(K\sqrt{s(\log p)/n})$ with probability tending to 1, and $\|n^{-1} \sum_{i=1}^{n} \mathbf{M} \boldsymbol{X}_i\|_\infty = n^{-1/2} \gamma_3$ by the concentration property of $n^{-1} \sum_{i=1}^{n} \mathbf{M} \boldsymbol{X}_i$ to its expectation (which is zero). (iii) is a higher order term in the Taylor expansion and can be shown to be of the order $o(n^{-1/2})$ as well after multiplying by $\mathbf{M}$.

Lastly, we bound the difference between (II) and its conditional expectation. Define

$$\boldsymbol{V} := \sum_{k=1}^{K} \frac{1}{n} \sum_{i=1}^{n} \Big\{ g_{ik}\big(\underset{\sim}{\boldsymbol{X}}_i^T \underset{\sim}{\widehat{\boldsymbol{\Delta}}}_k\big) - g_{ik}(0) \Big\} \boldsymbol{X}_i, \tag{6.3}$$

where $g_{ik}(t) := \mathbb{1}\{\varepsilon_i \le t + b_k^*\} - \mathbb{P}(\varepsilon_i \le t + b_k^*)$. We have the following lemma that bounds this term:

**Lemma 6.1.** Suppose Assumption 3.1 holds, and $\|\widehat{\boldsymbol{\beta}}\|_0 \le c^* s$, $\text{MSE}(\widehat{\boldsymbol{\beta}}) = \|\mathbb{X}(\widehat{\boldsymbol{\beta}} - \boldsymbol{\beta}^*)/n\|_2 \le C_1 \sqrt{s(\log p)/n}$, $\|\widehat{\boldsymbol{\beta}} - \boldsymbol{\beta}^*\|_2 \le C_2 \sqrt{s(\log p)/n}$ and $\|\widehat{\boldsymbol{b}} - \boldsymbol{b}^*\|_2 \le C_3 \sqrt{Ks(\log p)/n}$ for some constants $c^*, C_1, C_2$ and $C_3$ with probability tending to 1. Moreover, suppose $s(\log p)/n = o(1)$. Then

$$\sqrt{n} \|\mathbf{M}\boldsymbol{V}\|_\infty \lesssim K \gamma_2 (s \log p)^{3/4} / n^{1/4},$$

with probability tending to 1.

With the above discussions and Lemma 6.1, we are ready to prove Theorem 3.2. We combine (6.1), (6.2) and (6.3) to obtain

$$\widehat{\boldsymbol{\kappa}} = -\boldsymbol{U} + \boldsymbol{V} + \theta_K \widehat{\boldsymbol{\Sigma}}\widehat{\boldsymbol{\Delta}} + \boldsymbol{E} + \boldsymbol{D},$$

where $\boldsymbol{U} := \sum_{k=1}^K n^{-1}\sum_{i=1}^n \big(\tau_k - \mathbb{1}\{\epsilon_i \le b_k^*\}\big)\boldsymbol{X}_i$ denotes the negative of term (I) in (6.1), and $\boldsymbol{E} := \sum_{k=1}^K f_k^*(\widehat{b}_k - b_k^*) n^{-1}\sum_{i=1}^n \boldsymbol{X}_i$, $\boldsymbol{D} := \sum_{k=1}^K n^{-1}\sum_{i=1}^n f_\epsilon'(\xi_{ik})(\underset{\sim}{\boldsymbol{X}}_i^T \underset{\sim}{\widehat{\boldsymbol{\Delta}}}_k)^2 \boldsymbol{X}_i$ denote terms (ii) and (iii) in (6.2). This implies that $\widehat{\boldsymbol{\Sigma}}\widehat{\boldsymbol{\Delta}} - \theta_K^{-1}\widehat{\boldsymbol{\kappa}} = \theta_K^{-1}\boldsymbol{U} - \theta_K^{-1}\boldsymbol{V} - \theta_K^{-1}\boldsymbol{E} - \theta_K^{-1}\boldsymbol{D}$. Multiplying $\mathbf{M}$ and adding $\widehat{\boldsymbol{\Delta}}$ on both sides of of equation, we have

$$\widehat{\boldsymbol{\Delta}} - \theta_K^{-1}\mathbf{M}\widehat{\boldsymbol{\kappa}} = \theta_K^{-1}\mathbf{M}\boldsymbol{U} - \theta_K^{-1}\mathbf{M}\boldsymbol{V} - \theta_K^{-1}\mathbf{M}\boldsymbol{E} - \theta_K^{-1}\mathbf{M}\boldsymbol{D} - (\mathbf{M}\widehat{\boldsymbol{\Sigma}} - \mathbf{I})\widehat{\boldsymbol{\Delta}}.$$

Since $\widehat{\boldsymbol{\beta}}^d = \widehat{\boldsymbol{\beta}} - \widehat{\boldsymbol{\Theta}}\widehat{\boldsymbol{\kappa}}$ where $\widehat{\boldsymbol{\Theta}} = \widehat{\theta}_K^{-1}\mathbf{M}$, it holds that

$$\begin{aligned}\sqrt{n}(\widehat{\boldsymbol{\beta}}^d - \boldsymbol{\beta}^*) = \sqrt{n}\theta_K^{-1}\mathbf{M}\boldsymbol{U} - \sqrt{n}\theta_K^{-1}\mathbf{M}\boldsymbol{V} - \sqrt{n}\theta_K^{-1}\mathbf{M}\boldsymbol{E} \\ - \sqrt{n}\theta_K^{-1}\mathbf{M}\boldsymbol{D} - \sqrt{n}(\mathbf{M}\widehat{\boldsymbol{\Sigma}} - \mathbf{I})\widehat{\boldsymbol{\Delta}} - \sqrt{n}(\widehat{\theta}_K^{-1} - \theta_K^{-1})\mathbf{M}\widehat{\boldsymbol{\kappa}}.\end{aligned}$$

Let $\boldsymbol{Z} := \sqrt{n}\theta_K^{-1}\mathbf{M}\boldsymbol{U}$. We have by definition of $\boldsymbol{U}$ that $\boldsymbol{Z} = n^{-1/2}\sum_{i=1}^n \theta_K^{-1}\mathbf{M}\boldsymbol{X}_i\Psi_{i,K}$, where $\Psi_{i,K} = \sum_{k=1}^K \big(\tau_k - \mathbb{1}\{\epsilon_i \le b_k^*\}\big)$. This is the dominating term which weakly converges to a normal distribution. The other terms can be shown to be asymptotically ignorable. In the following we bound each term separately.

**The term** $\sqrt{n}(\mathbf{M}\widehat{\boldsymbol{\Sigma}} - \mathbf{I})\widehat{\boldsymbol{\Delta}}$**.** We have by the definition of $\mathbf{M}$ that $\|\mathbf{M}\widehat{\boldsymbol{\Sigma}} - \mathbf{I}\|_{\max} \le \gamma_1$. Therefore, by Hölder's inequality, it follows with probability tending to 1 that,

$$\|\sqrt{n}(\mathbf{M}\widehat{\boldsymbol{\Sigma}} - \mathbf{I})\widehat{\boldsymbol{\Delta}}\|_\infty \le \sqrt{n}\|\mathbf{M}\widehat{\boldsymbol{\Sigma}} - \mathbf{I}\|_{\max}\|\widehat{\boldsymbol{\beta}} - \boldsymbol{\beta}^*\|_1 \lesssim \gamma_1 s\sqrt{\log p},$$

where the last inequality follows from the $L_2$-convergence rate and sparsity of $\widehat{\boldsymbol{\beta}}$ and Cauchy-Schwarz inequality. By condition in the theorem, $\gamma_1 s\sqrt{\log p} = o(1)$. Hence we conclude that $\|\sqrt{n}(\mathbf{M}\widehat{\boldsymbol{\Sigma}} - \mathbf{I})\widehat{\boldsymbol{\Delta}}\|_\infty = o_P(1)$.

**The term** $\sqrt{n}\theta_K^{-1}\mathbf{M}\boldsymbol{E}$**.** This term is caused by the errors in estimating quantiles $b_k^*$. By Assumption 3.1, we have $\theta_K^{-1} \le K^{-1}C_-^{-1}$. Hence for all $j = 1, \dots, p$, it holds that

$$\begin{aligned}|\sqrt{n}\theta_K^{-1}[\mathbf{M}\boldsymbol{E}]_j| &\le \theta_K^{-1}\Big|\sum_{k=1}^K f_k^*(\widehat{b}_k - b_k^*)\Big|\Big|\frac{1}{\sqrt{n}}\sum_{i=1}^n \widehat{\boldsymbol{\mu}}_j^T\boldsymbol{X}_i\Big| \\ &\le K^{-1}C_-^{-1}C_+\|\widehat{\boldsymbol{b}} - \boldsymbol{b}^*\|_1\gamma_3.\end{aligned}$$

By Cauchy-Schwarz, we have $\|\widehat{\boldsymbol{b}} - \boldsymbol{b}^*\|_1 \le \sqrt{K}\|\widehat{\boldsymbol{b}} - \boldsymbol{b}^*\|_2$. It follows from the convergence rate of $\|\widehat{\boldsymbol{b}} - \boldsymbol{b}^*\|_2$ that $\|\sqrt{n}\theta_K^{-1}\mathbf{M}\boldsymbol{E}\|_\infty \le C_-^{-1}C_+C_3\gamma_3\sqrt{s(\log p)/n}$, with probability tending to 1. By condition in the theorem, $\gamma_3\sqrt{s(\log p)/n} = o(1)$. Hence we conclude that $\|\sqrt{n}\theta_K^{-1}\mathbf{M}\boldsymbol{E}\|_\infty = o_P(1)$.

**The term** $\sqrt{n}\theta_K^{-1}\mathbf{M}\boldsymbol{D}$**.** Recall $\boldsymbol{D} = \sum_{k=1}^K \frac{1}{n}\sum_{i=1}^n f_\epsilon'(\xi_{ik})(\underset{\sim}{\boldsymbol{X}}_i^T\underset{\sim}{\widehat{\boldsymbol{\Delta}}}_k)^2\boldsymbol{X}_i$. For any $j = 1, \dots, p$, we have by Assumption 3.1 that

$$\begin{aligned}|\sqrt{n}\theta_K^{-1}[\mathbf{M}\boldsymbol{D}]_j| &\le K^{-1}C_-^{-1}\sqrt{n}\Big|\sum_{k=1}^K \frac{1}{n}\sum_{i=1}^n f_\epsilon'(\xi_{ik})(\underset{\sim}{\boldsymbol{X}}_i^T\underset{\sim}{\widehat{\boldsymbol{\Delta}}}_k)^2\widehat{\boldsymbol{\mu}}_j^T\boldsymbol{X}_i\Big| \\ &\le K^{-1}C_-^{-1}C_+'\gamma_2\sqrt{n}\left(\frac{2K}{n}\sum_{i=1}^n(\boldsymbol{X}_i^T\widehat{\boldsymbol{\Delta}})^2 + \|\widehat{\boldsymbol{b}} - \boldsymbol{b}^*\|_2^2\right).\end{aligned} \tag{6.4}$$

By conditions in the theorem, we have $n^{-1}\sum_{i=1}^n(\boldsymbol{X}_i^T\widehat{\boldsymbol{\Delta}})^2 = \widehat{\boldsymbol{\Delta}}^T\widehat{\boldsymbol{\Sigma}}\widehat{\boldsymbol{\Delta}} = \text{MSE}(\widehat{\boldsymbol{\beta}})^2 \le C_1^2 s(\log p)/n$, and $\|\widehat{\boldsymbol{b}} - \boldsymbol{b}^*\|_2^2 \le C_3^2 Ks(\log p)/n$ with probability tending to 1. Hence it follows from (6.4) that $\|\sqrt{n}\theta_K^{-1}\mathbf{M}\boldsymbol{D}\|_\infty \le C_-^{-1}C_+'(C_1^2+C_3^2)\gamma_2 s(\log p)/\sqrt{n}$ with the same probability. By condition in the theorem, we have $\gamma_2 s(\log p)/\sqrt{n} = o(1)$. Hence we conclude that $\|\sqrt{n}\theta_K^{-1}\mathbf{M}\boldsymbol{D}\|_\infty = o_P(1)$.

**The term $\sqrt{n}\theta_K^{-1}\mathbf{M}\boldsymbol{V}$.** We bound this term by applying empirical process theory. Using Lemma 6.1 and the fact that $\theta_K^{-1}$ is upper bounded by $(KC_-)^{-1}$, we conclude that when $\gamma_2(s\log p)^{3/4}/n^{1/4} = o(1)$, it holds $\sqrt{n}\|\theta_K^{-1}\mathbf{M}\boldsymbol{V}\|_\infty = o_P(1)$.

**The term $\sqrt{n}(\widehat{\theta}_K^{-1} - \theta_K^{-1})\mathbf{M}\widehat{\boldsymbol{\kappa}}$.** This term corresponds to the error in estimating $\theta_K$. Note that $\|\sqrt{n}(\widehat{\theta}_K^{-1} - \theta_K^{-1})\mathbf{M}\widehat{\boldsymbol{\kappa}}\|_\infty \le \sqrt{n}|\widehat{\theta}_K^{-1} - \theta_K^{-1}|\|\sqrt{n}\mathbf{M}\widehat{\boldsymbol{\kappa}}\|_\infty$. By (6.1) and (6.3), we have

$$\sqrt{n}\mathbf{M}\widehat{\boldsymbol{\kappa}} = -\sqrt{n}\mathbf{M}\boldsymbol{U} + \sqrt{n}\mathbf{M}\boldsymbol{V} + \sqrt{n}\mathbf{M}\boldsymbol{T} \tag{6.5}$$

where $\boldsymbol{T} := \sum_{k=1}^K n^{-1}\sum_{i=1}^n \big\{F_\epsilon\big(\underset{\sim}{\boldsymbol{X}}_i^T\underset{\sim}{\widehat{\boldsymbol{\Delta}}}_k + b_k^*\big) - F_\epsilon(b_k^*)\big\}\boldsymbol{X}_i$. We bound the three terms on the R.H.S. respectively. For the first term, we have

$$-\sqrt{n}[\mathbf{M}\boldsymbol{U}]_j = \sum_{k=1}^K \frac{1}{\sqrt{n}}\sum_{i=1}^n \big(\mathbb{1}\{\epsilon_i \le b_k^*\} - \tau_k\big)\widehat{\boldsymbol{\mu}}_j^T\boldsymbol{X}_i.$$

The summands are i.i.d. with mean zero conditioned on the design $\mathbb{X}$. Moreover, each of them is bounded by $|\widehat{\boldsymbol{\mu}}_j^T\boldsymbol{X}_i| \le \gamma_2$. Hence by Hoeffding's inequality, it holds that

$$\mathbb{P}\left(\left|\frac{1}{\sqrt{n}}\sum_{i=1}^n (\mathbb{1}\{\epsilon_i \le b_k^*\} - \tau_k)\,\widehat{\boldsymbol{\mu}}_j^T\boldsymbol{X}_i\right| > t\right) \le e^{-t^2/(2\gamma_2^2)}.$$

Applying the union bound, we have

$$\mathbb{P}\Big(\Big\|\sum_{k=1}^K \frac{1}{\sqrt{n}}\sum_{i=1}^n \big(\mathbb{1}\{\epsilon_i \le b_k^*\} - \tau_k\big)\mathbf{M}\boldsymbol{X}_i\Big\|_\infty > Kt\Big) \le pKe^{-t^2/(2\gamma_2^2)}.$$

Taking $t = 4\gamma_2\sqrt{\log p}$, we have with probability at least $1 - Kp^{-7}$ that $\|\sqrt{n}\mathbf{M}\boldsymbol{V}\|_\infty \le 4K\gamma_2\sqrt{\log p}$. The second term in the R.H.S. of (6.5) can be directly controlled using Lemma 6.1, which shows that $\|\sqrt{n}\mathbf{M}\boldsymbol{V}\|_\infty \lesssim \gamma_2 K(s\log p)^{3/4}/n^{1/4}$, with probability tending to 1. We are left to bound $\sqrt{n}\mathbf{M}\boldsymbol{T}$. By definition of $\boldsymbol{T}$, we have

$$\|\sqrt{n}\mathbf{M}\boldsymbol{T}\|_\infty \le \sum_{k=1}^K \frac{1}{\sqrt{n}}\sum_{i=1}^n \big|F_\epsilon\big(\underset{\sim}{\boldsymbol{X}}_i^T\underset{\sim}{\widehat{\boldsymbol{\Delta}}}_k + b_k^*\big) - F_\epsilon(b_k^*)\big|\|\mathbf{M}\boldsymbol{X}_i\|_\infty \le \sum_{k=1}^K \frac{1}{\sqrt{n}}\sum_{i=1}^n C_+\gamma_2|\underset{\sim}{\boldsymbol{X}}_i^T\underset{\sim}{\widehat{\boldsymbol{\Delta}}}_k|.$$

By Cauchy-Schwarz, $\frac{1}{\sqrt{n}}\sum_{i=1}^n|\underset{\sim}{\boldsymbol{X}}_i^T\underset{\sim}{\widehat{\boldsymbol{\Delta}}}_k| \le \sqrt{n}\Big(\frac{1}{n}\sum_{i=1}^n(\underset{\sim}{\boldsymbol{X}}_i^T\underset{\sim}{\widehat{\boldsymbol{\Delta}}}_k)^2\Big)^{1/2} \le \sqrt{n}\left(\text{MSE}(\widehat{\boldsymbol{\beta}})^{1/2} + |b_k|\right)$ for each $k = 1, \ldots, K$. Therefore, we have

$$\begin{aligned}
\|\sqrt{n}\mathbf{M}\boldsymbol{T}\|_\infty &\le C_+\gamma_2\sqrt{n}\sum_{k=1}^K\big(\text{MSE}(\widehat{\boldsymbol{\beta}})^{1/2} + |b_k|\big)\\
&\le C_+\gamma_2\sqrt{n}\big(K\text{MSE}(\widehat{\boldsymbol{\beta}})^{1/2} + \sqrt{K}\|\widehat{\boldsymbol{b}} - \boldsymbol{b}^*\|_2\big)\\
&\lesssim \gamma_2 K\sqrt{s\log p},
\end{aligned}$$

with probability tending to 1. Hence by (6.5) and the above bounds, we conclude that

$$\|\sqrt{n}\mathbf{M}\widehat{\boldsymbol{\kappa}}\|_\infty \lesssim \gamma_2 K(\sqrt{\log p} + (s\log p)^{3/4}/n^{1/4} + \sqrt{s\log p}), \tag{6.6}$$

with probability tending to 1. Lastly, we have $|\widehat{\theta}_K^{-1} - \theta_K^{-1}| = \widehat{\theta}_K^{-1}|\widehat{\theta}_K/\theta_K - 1| \le \widehat{\theta}_K^{-1} r_t$. As $\widehat{\theta}_K \ge \theta_K - |\widehat{\theta}_K - \theta_K| \ge KC_- - \theta_K|\widehat{\theta}_K/\theta_K - 1| \ge KC_- - KC_+ r_t$ and $r_t = o(1)$, we have $\widehat{\theta}_K \ge KC_-/2$ for large enough $n$. Therefore, it holds that $|\widehat{\theta}_K^{-1} - \theta_K^{-1}| \lesssim K^{-1} r_t$. Hence by (6.6),

$$\left\|\sqrt{n}(\widehat{\theta}_K^{-1} - \theta_K^{-1})\mathbf{M}\widehat{\boldsymbol{\kappa}}\right\|_\infty \lesssim r_t\gamma_2\left(\sqrt{\log p} + (s\log p)^{3/4}/n^{1/4} + \sqrt{s\log p}\right)$$

with probability tending to 1. So when the scaling condition $r_t\gamma_2\sqrt{s\log p} = o(1)$ is satisfied, we have $\|\sqrt{n}(\widehat{\theta}_K^{-1} - \theta_K^{-1})\mathbf{M}\widehat{\boldsymbol{\kappa}}\|_\infty = o_P(1)$. This concludes the proof and we are only left to prove Lemma 6.1.

*Proof of Lemma 6.1.* We sketch the proof here and defer details to Appendix A. For any $\boldsymbol{\Delta} \in \mathbb{R}^p$ and $\delta_1, \ldots, \delta_k \in \mathbb{R}$, let $\underset{\sim}{\boldsymbol{\Delta}} = (\boldsymbol{\Delta}^T, \delta_1, \ldots, \delta_k)^T \in \mathbb{R}^{p+K}$ and $\underset{\sim}{\boldsymbol{\Delta}}_k = (\boldsymbol{\Delta}^T, \delta_k) \in \mathbb{R}^{p+1}$ and $\boldsymbol{\delta} = (\delta_1, \ldots, \delta_K)^T$. Define

$$\psi_j(\epsilon_i, \boldsymbol{X}_i; \underset{\sim}{\boldsymbol{\Delta}}_k) := \widehat{\boldsymbol{\mu}}_j^T \boldsymbol{X}_i\big(\mathbb{1}\{\epsilon_i \le \underset{\sim}{\boldsymbol{X}}_i^T \underset{\sim}{\boldsymbol{\Delta}}_k + b_k^*\} - \mathbb{1}\{\epsilon_i \le b_k^*\}\big), \tag{6.7}$$

and let

$$\mathbb{G}_j(\underset{\sim}{\boldsymbol{\Delta}}) := \frac{1}{\sqrt{n}}\sum_{i=1}^n\sum_{k=1}^K \big\{\psi_j(\epsilon_i, \boldsymbol{X}_i; \underset{\sim}{\boldsymbol{\Delta}}_k) - \mathbb{E}[\psi_j(\epsilon_i, \boldsymbol{X}_i; \underset{\sim}{\boldsymbol{\Delta}}_k)\,|\,\mathbb{X}]\big\}. \tag{6.8}$$

Moreover, for some positive integer $q$ and real numbers $\xi_1, \xi_2 > 0$, let

$$R(q, \xi_1, \xi_2) := \big\{\underset{\sim}{\boldsymbol{\Delta}} = (\boldsymbol{\Delta}, \boldsymbol{\delta}) \in \mathbb{R}^p \times \mathbb{R}^K : \|\boldsymbol{\Delta}\|_0 \le q, \|\boldsymbol{\Delta}\|_{\widehat{\boldsymbol{\Sigma}}} \le \xi_1, \|\boldsymbol{\delta}\|_2 \le \xi_2\big\}. \tag{6.9}$$

With probability tending to 1, we have $\widehat{\underset{\sim}{\boldsymbol{\Delta}}} \in R(c^*s, C_1\xi_n, C_3\sqrt{K}\xi_n)$, where $\xi_n = \sqrt{s(\log p)/n}$. Therefore, with the same probability, we have $\sqrt{n}\|\mathbf{M}\boldsymbol{V}\|_\infty \le \max_{1\le j\le p}\sup_{\underset{\sim}{\boldsymbol{\Delta}}\in R(c^*s, C_1\xi_n, C_3\sqrt{K}\xi_n)}|\mathbb{G}_j(\underset{\sim}{\boldsymbol{\Delta}})|$. The R.H.S. is a supremum of an empirical process. To obtain a sharp convergence rate, we need to utilize the information that $\boldsymbol{\Delta}$ (and $\boldsymbol{\delta}$) lies in a ball that shrinks at the rate $\sqrt{s(\log p)/n}$ (and $\sqrt{Ks(\log p)/n}$). The main difficulty is that $\mathbb{G}_j(\underset{\sim}{\boldsymbol{\Delta}})$ is neither a Lipschitz nor sub-Gaussian process, as the indicator function is non-differentiable. To solve this problem, a key observation is that the variance of the summands can be controlled by the convergence bounds of $\boldsymbol{\Delta}$ and $\boldsymbol{\delta}$. In Appendix A, we show that $\mathrm{Var}\left[\psi_j(\epsilon_i, \boldsymbol{X}_i; \underset{\sim}{\boldsymbol{\Delta}}_k)\,|\,\mathbb{X}\right] \le 6\gamma_2^2 C_+|\underset{\sim}{\boldsymbol{X}}_i^T\underset{\sim}{\boldsymbol{\Delta}}_k|$, and that

$$\sigma_n^2 := \frac{1}{n}\sum_{i=1}^n \mathrm{Var}\left[\sum_{k=1}^K \psi_j(\epsilon_i, \boldsymbol{X}_i; \underset{\sim}{\boldsymbol{\Delta}}_k)\,|\,\mathbb{X}\right] \lesssim K^2\gamma_2^2\xi_n. \tag{6.10}$$

The technical tools we use to control $\sup_{\underset{\sim}{\boldsymbol{\Delta}}\in R(c^*s, C_1\xi_n, C_3\sqrt{K}\xi_n)}|\mathbb{G}_j(\underset{\sim}{\boldsymbol{\Delta}})|$ are Theorem 3.11 in Koltchinskii (2011) for bounding the expectation of the supremum of an empirical process, and Bousquet's concentration inequality (Bousquet, 2002) for obtaining a tail probability, both of which are able to utilize the variance information in (6.10). We present the details in the supplementary Appendix A. Roughly speaking, these theories suggest that the convergence rate of $\sup_{\underset{\sim}{\boldsymbol{\Delta}}\in R(c^*s, C_1\xi_n, C_3\sqrt{K}\xi_n)}|\mathbb{G}_j(\underset{\sim}{\boldsymbol{\Delta}})|$ is $\sigma_n$ times the complexity of the set of functions

$$\mathcal{F} := \left\{\sum_{k=1}^K \psi_j(\epsilon, \boldsymbol{x}; \underset{\sim}{\boldsymbol{\Delta}}_k)\,|\,\underset{\sim}{\boldsymbol{\Delta}} \in R(q, C_1\xi_n, C_3\sqrt{K}\xi_n)\right\}.$$

Here the complexity is measured by square root of uniform covering entropy, and we show that it is roughly of the order $\sqrt{s\log p}$. Therefore, the convergence rate of $\sup_{\underset{\sim}{\boldsymbol{\Delta}}\in R(c^*s,C_1\xi_n,C_3\sqrt{K}\xi_n)}|\mathbb{G}_j(\underset{\sim}{\boldsymbol{\Delta}})|$ is $\sigma_n\sqrt{s\log p}=K\gamma_2\sqrt{s(\log p)\xi_n}=K\gamma_2(s\log p)^{3/4}/n^{1/4}$. The conclusion of the lemma then follows by applying the union bound. □

## 6.2 Proof of Theorem 3.3

We apply the Lindeberg central limit theorem to prove the asymptotic normality of $\boldsymbol{Z}$. We check the Lindeberg condition. Define $\xi_{ij}=\theta_K^{-1}\Psi_{i,K}\widehat{\boldsymbol{\mu}}_j^T\boldsymbol{X}_i/(\widehat{\boldsymbol{\mu}}_j^T\widehat{\boldsymbol{\Sigma}}\widehat{\boldsymbol{\mu}}_j)^{1/2}$. The summands have mean zero conditional on $\mathbb{X}$. Indeed, we have

$$\mathbb{E}[\xi_{ij}\,|\,\mathbb{X}]=\mathbb{E}\Big[\theta_K^{-1}\Psi_{i,K}\widehat{\boldsymbol{\mu}}_j^T\boldsymbol{X}_i/(\widehat{\boldsymbol{\mu}}_j^T\widehat{\boldsymbol{\Sigma}}\widehat{\boldsymbol{\mu}}_j)^{1/2}\,|\,\mathbb{X}\Big]=\theta_K^{-1}\widehat{\boldsymbol{\mu}}_j^T\boldsymbol{X}_i/(\widehat{\boldsymbol{\mu}}_j^T\widehat{\boldsymbol{\Sigma}}\widehat{\boldsymbol{\mu}}_j)^{1/2}\mathbb{E}\big[\Psi_{i,K}\big]$$

by the fact that $\boldsymbol{\epsilon}$ is independent of $\mathbb{X}$. By definition, $b_k^*$ is the $\tau_k$-quantile of $\epsilon$. Hence

$$\mathbb{E}\big[\Psi_{i,K}\big]=\sum_{k=1}^K\mathbb{E}\big[\,\mathbb{1}\{\epsilon_i\le b_k^*\}-\tau_k\big]=\sum_{k=1}^K\big\{\mathbb{P}(\epsilon_i\le b_k^*)-\tau_k\big\}=0.$$

Therefore $\mathbb{E}[\xi_{ij}\,|\,\mathbb{X}]=0$. We next calculate the variance of the summands:

$$s_n^2:=\sum_{i=1}^n\mathrm{Var}\Big(\xi_{ij}\,|\,\mathbb{X}\Big)=\theta_K^{-2}\sum_{i=1}^n\widehat{\boldsymbol{\mu}}_j^T\boldsymbol{X}_i\boldsymbol{X}_i^T\widehat{\boldsymbol{\mu}}_j/(\widehat{\boldsymbol{\mu}}_j^T\widehat{\boldsymbol{\Sigma}}\widehat{\boldsymbol{\mu}}_j)\,\mathrm{Var}\big(\Psi_{i,K}\,|\,\mathbb{X}\big)=n\theta_K^{-2}\,\mathrm{Var}\big(\Psi_{i,K}\,|\,\mathbb{X}\big),$$

and $\mathrm{Var}\big(\Psi_{i,K}\,|\,\mathbb{X}\big)=\mathrm{Var}\Big(\sum_{k=1}^K\mathbb{1}\{\epsilon\le b_k^*\}-\tau_k\Big)=\sum_{k,k'=1}^K\min\{\tau_k,\tau_{k'}\}(1-\max\{\tau_k,\tau_{k'}\})=\sigma_K^2$. Hence $s_n^2=n\theta_K^{-2}\sigma_K^2$. By conditions in the theorem, we have $(\widehat{\boldsymbol{\mu}}_j^T\widehat{\boldsymbol{\Sigma}}\widehat{\boldsymbol{\mu}}_j)^{1/2}\ge c_n$ where $c_n$ satisfies $\liminf_{n\to\infty}c_n=c_\infty>0$. Therefore it follows that $|\xi_{ij}|\le c_n^{-1}\theta_K^{-1}|\Psi_{i,K}|\|\widehat{\boldsymbol{\mu}}_j^TX_i\|_\infty\le c_n^{-1}\theta_K^{-1}\gamma_2|\Psi_{i,K}|$. Hence we have for any $\varepsilon>0$,

$$\lim_{n\to\infty}\frac{1}{s_n^2}\sum_{i=1}^n\mathbb{E}\big[\xi_{ij}^2\,\mathbb{1}\{|\xi_{ij}|>\varepsilon s_n\}\,|\,\mathbb{X}\big]\le\theta_K^2\sigma_K^{-2}\lim_{n\to\infty}\frac{1}{n}\sum_{i=1}^n\mathbb{E}\big[\xi_{ij}^2\,\mathbb{1}\{|\Psi_{i,K}|>\varepsilon c_n\sigma_K\gamma_2^{-1}\sqrt{n}\}\,|\,\mathbb{X}\big].$$

We have $\mathbb{E}[\xi_{ij}^2\,|\,\mathbb{X}]=\theta_K^{-2}\widehat{\boldsymbol{\mu}}_j^T\boldsymbol{X}_i\boldsymbol{X}_i^T\widehat{\boldsymbol{\mu}}_j/(\widehat{\boldsymbol{\mu}}_j^T\boldsymbol{\Sigma}\widehat{\boldsymbol{\mu}}_j)\mathbb{E}\big[(\Psi_{1,K})^2\big]$. By the identity $\widehat{\boldsymbol{\Sigma}}=n^{-1}\sum_{i=1}^n\boldsymbol{X}_i\boldsymbol{X}_i^T$, we further derive

$$\begin{aligned}\lim_{n\to\infty}\frac{1}{s_n^2}\sum_{i=1}^n\mathbb{E}\big[\xi_{ij}^2\,\mathbb{1}\{|\xi_{ij}|>\varepsilon s_n\}\,|\,\mathbb{X}\big]&\le\sigma_K^{-2}\lim_{n\to\infty}\mathbb{E}\big[(\Psi_{1,K})^2\,\mathbb{1}\{|\Psi_{1,K}|>\varepsilon c_n\sigma_K\gamma_2^{-1}\sqrt{n}\}\big]\\&\le\lim_{n\to\infty}\sigma_K^{-4}\varepsilon^{-2}c_n^{-2}\gamma_2^2n^{-1}\mathbb{E}\big[(\Psi_{1,K})^4\big]=0,\end{aligned}$$

where the last equality is by the uniform boundedness of $\Psi_{1,K}$ and the fact that $\gamma_2n^{-1/2}=o(1)$. Hence the Lindeberg condition is satisfied. As $s_n^2/n=\theta_K^{-2}\sigma_K^2$, we have by Lindeberg central limit theorem that

$$Z_j/(\widehat{\boldsymbol{\mu}}_j^T\widehat{\boldsymbol{\Sigma}}\widehat{\boldsymbol{\mu}}_j)^{1/2}=\frac{1}{\sqrt{n}}\sum_{i=1}^n\xi_{ij}\rightsquigarrow N(0,\theta_K^{-2}\sigma_K^2),$$

which completes the proof.

## 7 Numerical Results

In this section, we empirically examine the finite sample performance of our de-biased composite quantile estimator (2.4). Our simulation will highlight the two distinctive attributes of our de-biased estimator: robustness and efficiency. Furthermore, we will empirically verify the claim that the first-stage estimators can be chosen flexibly under our de-biasing framework.

Specifically, we consider the high-dimensional linear model (1.1). In the simulation setup, we let each row of the design matrix be an independent realization of a multivariate Gaussian random vector with mean $\mathbf{0}$ and covariance matrix $\mathbf{\Sigma}$, whose entries $\mathbf{\Sigma}_{jk}$ are given as follows:

$$\mathbf{\Sigma}_{jk} = \begin{cases} 1, & \text{for } k = j; \\ 0.1, & \text{for } 1 \le |j-k| \le 5 \text{ or } |j-k| \ge p-5; \\ 0, & \text{otherwise.} \end{cases}$$

In our simulation, we take $n = 200$, $p = 250$. The regression parameter $\boldsymbol{\beta}^*$ is set to be sparse, such that $\beta_j^* = 1$ for $j \in \{1, \ldots, 5\}$ and 0 otherwise.

We evaluate the performance of our de-biased composite quantile estimator under different noise models, that is, light-tailed noise, heavy-tailed noise with second moment, and heavy-tailed noise without second moment. We use Gaussian with variance 1, t-distribution with degree of freedom 3 and Cauchy distribution with scale 1 as representatives of the above three distribution families.

For clarity of comparison, we denote our de-biased composite quantile estimator (2.4) as $\widehat{\boldsymbol{\beta}}^{\text{DCQ}}$. We choose $K = 9$ and the sequence $\tau_1 = 0.1, \tau_2 = 0.2, \ldots, \tau_9 = 0.9$. For the first-stage estimator $\widehat{\boldsymbol{\beta}}$, we adopt the penalized-1/2-quantile regression (that is, least absolute deviation (LAD)) estimator, defined as follows:

$$(\widehat{\boldsymbol{\beta}}^{\text{PLAD}}, \widehat{b}) \in \underset{\boldsymbol{\beta} \in \mathbb{R}^p, b \in \mathbb{R}}{\operatorname{argmin}} \frac{1}{2n} \sum_{i=1}^{n} |Y_i - \boldsymbol{X}_i^T \boldsymbol{\beta} - b| + \lambda \|\boldsymbol{\beta}\|_1, \tag{7.1}$$

which is (4.3) when $\tau = 1/2$. The penalized-LAD estimator is robust and converges to $\boldsymbol{\beta}^*$ at the $L_2$-rate of $\sqrt{s(\log p)/n}$ (Wang, 2013; Belloni and Chernozhukov, 2011). In our simulation we choose the tuning parameter $\lambda = 10\sqrt{(\log p)/n}$. It is observed in the simulations that our inference results are not sensitive to the choice of the constant in $\lambda$. To compute $\widehat{\boldsymbol{\kappa}}$, we need to estimate $b_k^*$, the $\tau_k$-quantiles of the noise for $k = 1, \ldots, K$. We obtain estimates $\widehat{b}_k$ by

$$\widehat{b}_k = \tau_k\text{-th quantile of } \widehat{\epsilon}_i^{\text{PLAD}} := Y_i - \boldsymbol{X}_i^T \widehat{\boldsymbol{\beta}}^{\text{PLAD}}.$$

The estimator $\mathbf{M}$ is obtained by minimizing the objective in (2.5) for $j = 1, \ldots, p$. We choose $\gamma_1 = 0.5\sqrt{(\log p)/n}$, $\gamma_2 = 5\sqrt{\log p}$ and $\gamma_3 = 5\sqrt{\log p}$. In the simulations we observe that the second and third constraints are often inactive at the solution when $\gamma_2, \gamma_3 \ge \sqrt{\log p}$, therefore the inference results are insensitive to the choices of $\gamma_2$ and $\gamma_3$. Lastly, we consider the case that $f_\epsilon$ is known for simplicity, so $\widehat{\theta}_K = \sum_{k=1}^K f_\epsilon(\widehat{b}_k)$.

We also consider the de-biased Lasso estimator proposed by Javanmard and Montanari (2014) for inference in the high-dimensional linear model. The estimator is defined as follows:

$$\widehat{\boldsymbol{\beta}}^{\text{DLasso}} = \widehat{\boldsymbol{\beta}}^{\text{Lasso}} + \mathbf{M}\mathbb{X}^T(\boldsymbol{Y} - \mathbb{X}\widehat{\boldsymbol{\beta}}^{\text{Lasso}})/n, \tag{7.2}$$

where

$$\widehat{\boldsymbol{\beta}}^{\text{Lasso}} \in \underset{\boldsymbol{\beta} \in \mathbb{R}^p}{\operatorname{argmin}} \|\boldsymbol{Y} - \mathbb{X}\boldsymbol{\beta}\|_2^2 + \lambda \|\boldsymbol{\beta}\|_1 \tag{7.3}$$

is the Lasso estimator. We follow the practice of Javanmard and Montanari (2014) by choosing $\lambda = 4\widehat{\sigma}\sqrt{(2\log p)/n}$, where $\widehat{\sigma}$ is given by the scaled Lasso

$$\{\widehat{\boldsymbol{\beta}}(\widetilde{\lambda}), \widehat{\sigma}(\widetilde{\lambda})\} \in \underset{\boldsymbol{\beta}\in\mathbb{R}^p, \sigma>0}{\operatorname{argmin}} \left\{\frac{1}{2\sigma n}\|\boldsymbol{Y} - \mathbb{X}\boldsymbol{\beta}\|_2^2 + \frac{\sigma}{2} + \widetilde{\lambda}\|\boldsymbol{\beta}\|_1\right\},$$

with $\widetilde{\lambda} = 10\sqrt{(2\log p)/n}$.

We first empirically verify the asymptotic normality of the coordinates of $\widehat{\boldsymbol{\beta}}^{\text{DCQ}}$ under the three types of noise distributions. In all three cases, we compute $\widehat{\beta}_j^{\text{DCQ}}$ for $j = 1, \ldots, p$ based on 200 realizations, and then plot their histograms and empirical densities. Figure 2 shows the results for $j = 3$ and $j = 6$, the former representing the coordinates on the support of $\boldsymbol{\beta}^*$ and the latter for those that are not. As comparisons, we also compute $\widehat{\beta}_j^{\text{DLasso}}$ based on the same realizations and plot the histograms for $j = 3$ and $j = 6$.

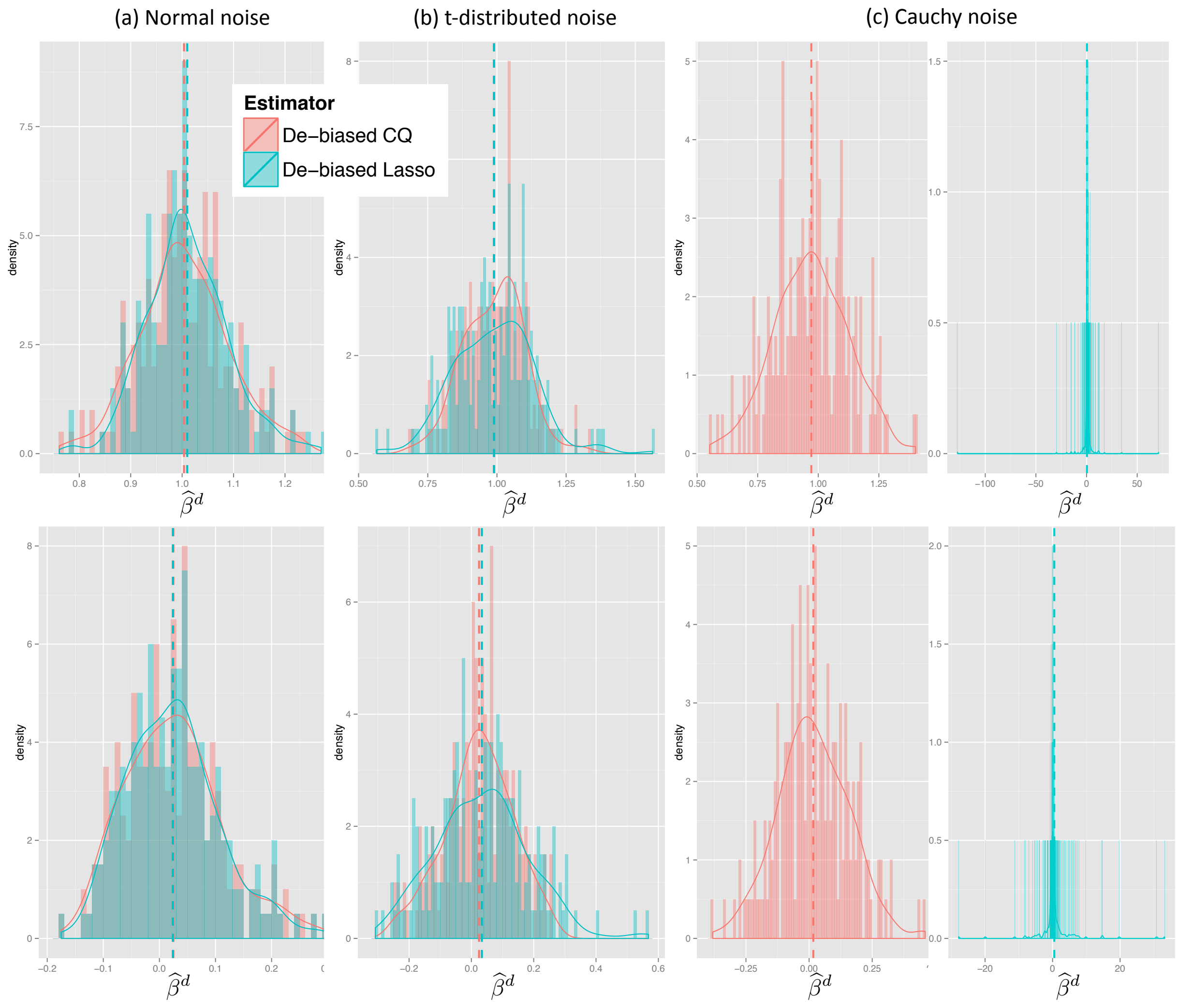

Figure 2: Histograms and empirical densities of $\widehat{\beta}_j^{\text{DCQ}}$ (red) and $\widehat{\beta}_j^{\text{DLasso}}$ (blue) for $j = 3, 6$ based on 200 realizations under (a) Gaussian noise (b) t-distributed noise and (c) Cauchy noise.

Figure 2 confirms the asymptotic normality of our de-biased CQ estimator, for all three types of noise distributions. Specifically, the distribution of $\widehat{\beta}_3^{\mathrm{DCQ}}$ is approximately Gaussian centered at 1, while $\widehat{\beta}_6^{\mathrm{DCQ}}$ is centered at 0. By contrast, the de-biased Lasso estimator fails to converge when the noise is Cauchy. Furthermore, when the noise is Gaussian, the distributions of the two estimators look almost the same, with the variance of $\widehat{\beta}_j^{\mathrm{DCQ}}$ slightly larger than that of $\widehat{\beta}_j^{\mathrm{DLasso}}$. This conforms with our theory that the relative efficiency of the de-biased CQ is approximately 95% of that of the de-biased Lasso under Gaussian noise. For $t$-distributed noise, the variance of $\widehat{\beta}_j^{\mathrm{DCQ}}$ is much smaller than that of $\widehat{\beta}_j^{\mathrm{DLasso}}$ for both $j = 3$ and $j = 6$, which shows that the de-biased CQ estimator is robust and efficient under heavy-tailed noises.

We next compare the confidence intervals constructed by the de-biased CQ procedure with other methods. According to Theorem 3.10, the 95%-confidence interval of $\beta_j^*$ based on the de-biased CQ estimator is constructed as follows:

$$\mathrm{CI}_j^{\mathrm{DCQ}} = \widehat{\beta}_j^{\mathrm{DCQ}} \pm 1.96\sigma_K \widehat{\theta}_K^{-1} (\mathbf{M}\widehat{\mathbf{\Sigma}}\mathbf{M}^T)_{jj}^{1/2}/\sqrt{n},$$

where, recall, $\sigma_K^2 = \sum_{k,k'=1}^{K} \min\{\tau_k, \tau_{k'}\}(1 - \max\{\tau_k, \tau_{k'}\})$. The confidence interval based on de-biased Lasso estimator is constructed as follows:

$$\mathrm{CI}_j^{\mathrm{DLasso}} = \widehat{\beta}_j^{\mathrm{DLasso}} \pm 1.96\widehat{\sigma}(\mathbf{M}\widehat{\mathbf{\Sigma}}\mathbf{M}^T)_{jj}^{1/2}/\sqrt{n}.$$

In addition to comparison with the de-biased Lasso, we evaluate the performance of our procedure when $K = 1$, which corresponds to the de-biased single quantile method. We adopt $\tau_1 = 1/2$, so the estimator is defined as follows:

$$\widehat{\boldsymbol{\beta}}^{\mathrm{DQuant}} = \widehat{\boldsymbol{\beta}}^{\mathrm{PLAD}} + \frac{1}{2n}\sum_{i=1}^{n} f_\epsilon(\widehat{b})^{-1}\,\mathrm{sign}(Y_i - \boldsymbol{X}_i\widehat{\boldsymbol{\beta}}^{\mathrm{PLAD}} - \widehat{b})\mathbf{M}\boldsymbol{X}_i,$$

where $\widehat{b}$ is an estimate of the median of the noise and we can directly use the one in (7.1). The confidence interval is constructed as

$$\mathrm{CI}_j^{\mathrm{DQuant}} = \widehat{\beta}_j^{\mathrm{DQuant}} \pm 1.96(4f_\epsilon(\widehat{b}))^{-1}(\mathbf{M}\widehat{\mathbf{\Sigma}}\mathbf{M}^T)_{jj}^{1/2}/\sqrt{n},$$

We compute the empirical coverage probability $\mathrm{CP}_j^{\mathrm{DCQ}} = \mathbb{P}_L\big(\beta_j^* \in \mathrm{CI}_j^{\mathrm{DCQ}}\big)$, where $\mathbb{P}_L$ with $L = 200$ denotes the empirical probability based on 200 realizations. $\mathrm{CP}_j^{\mathrm{DLasso}}$ and $\mathrm{CP}_j^{\mathrm{DQuant}}$ are defined analogously. Moreover, we compute the average lengths $\mathrm{AL}_j^{\mathrm{DCQ}}, \mathrm{AL}_j^{\mathrm{DLasso}}, \mathrm{AL}_j^{\mathrm{DQuant}}$ of confidence intervals $\mathrm{CI}_j^{\mathrm{DCQ}}, \mathrm{CI}_j^{\mathrm{DLasso}}, \mathrm{CI}_j^{\mathrm{DQuant}}$, respectively, for $j = 1, \ldots, p$. In Figure 3 we plot the coverage probabilities (denoted by dots) and average interval lengths (denoted by lines) under the three noise distributions.

The coverage probabilities in Figure 3 demonstrate that the de-baised CQ is robust: $\mathrm{CP}^{\mathrm{DCQ}}$ is approximately around 95% under all three noise conditions, while $\mathrm{CP}^{\mathrm{DLasso}}$ falls below 90% and 50% with t-distributed and Cauchy noises, respectively. The average interval lengths demonstrate the the de-biased CQ is efficient: $\mathrm{AL}^{\mathrm{DCQ}}$ is only slightly larger than $\mathrm{AL}^{\mathrm{DLasso}}$ and much smaller than $\mathrm{AL}^{\mathrm{DQuant}}$ under Gaussian noise, which conforms with our theory that the de-biased CQ estimator preserves efficiency under the Gaussian noise while the de-biased single quantile estimator does not. Moreover, the de-biased CQ is more efficient than the two competitors under the t-distributed noise, and has similar performance as de-biased quantile method under Cauchy noise.

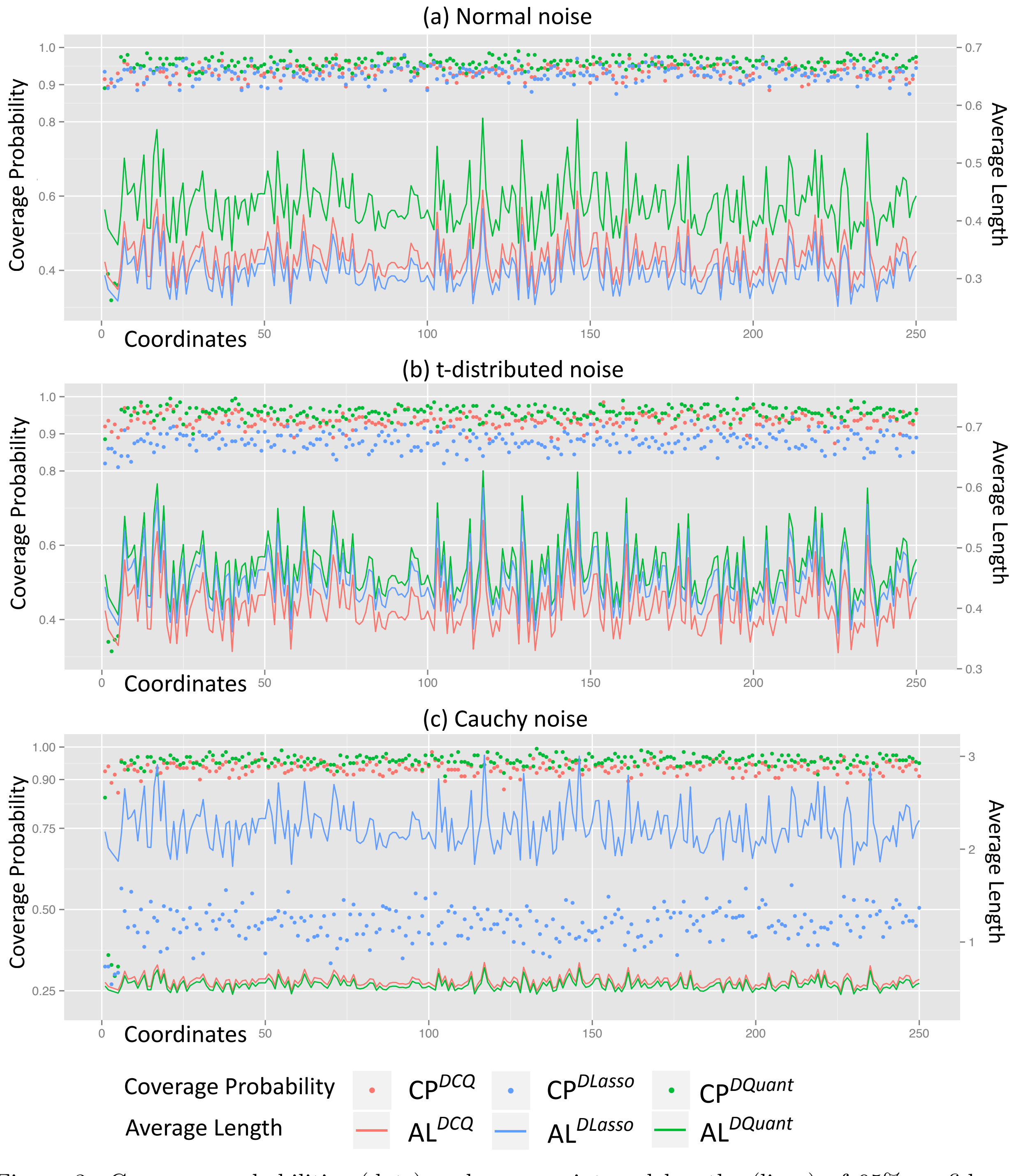

Figure 3: Coverage probabilities (dots) and average interval lengths (lines) of 95%-confidence intervals constructed by three methods: de-biased composite quantile (red), de-biased Lasso (blue) and de-biased single quantile (green), under different noise distributions.

Lastly, we empirically verify the claim that the first-stage estimator can be chosen flexibly for our de-biasing method. We compute confidence intervals constructed by various combinations of first-stage estimators and second-stage de-biasing methods. For the first-stage estimator, we take the penalized CQR $\widehat{\boldsymbol{\beta}}^{\mathrm{PCQR}}$, penalized LAD $\widehat{\beta}^{\mathrm{PLAD}}$, and Lasso $\widehat{\beta}^{\mathrm{Lasso}}$, defined as per (2.3), (7.1) and (7.3), respectively. For the second-stage de-biasing methods, we consider the ones associated with composite quantile loss and square loss functions, as shown in (2.4) and (7.2), respectively. We summarize the results of covering probabilities and interval lengths by taking average across all coordinates, coordinates on true support $\mathcal{T}$ and those on $\mathcal{T}^c$, that is, we define

$$\overline{\mathrm{CP}} = \frac{1}{p}\sum_{j=1}^{p} \mathrm{CP}_j, \quad \overline{\mathrm{CP}}_{\mathcal{T}} = \frac{1}{s}\sum_{j\in\mathcal{T}} \mathrm{CP}_j, \quad \overline{\mathrm{CP}}_{\mathcal{T}} = \frac{1}{p-s}\sum_{j\in\mathcal{T}^c} \mathrm{CP}_j,$$

and

$$\overline{\mathrm{AL}} = \frac{1}{p}\sum_{j=1}^{p} \mathrm{AL}_j, \quad \overline{\mathrm{AL}}_{\mathcal{T}} = \frac{1}{s}\sum_{j\in\mathcal{T}} \mathrm{AL}_j, \quad \overline{\mathrm{AL}}_{\mathcal{T}} = \frac{1}{p-s}\sum_{j\in\mathcal{T}^c} \mathrm{AL}_j.$$

The comparison results are shown in Tables 1, 2 and 3, under different noise distributions.

Table 1: Coverage probabilities and average interval lengths under Gaussian noise.

| | $\overline{\mathrm{CP}}$ | $\overline{\mathrm{CP}}_{\mathcal{T}}$ | $\overline{\mathrm{CP}}_{\mathcal{T}^c}$ | $\overline{\mathrm{AL}}$ | $\overline{\mathrm{AL}}_{\mathcal{T}}$ | $\overline{\mathrm{AL}}_{\mathcal{T}^c}$ |
|---|---|---|---|---|---|---|
| PCQR + CQ | 0.9317 | 0.9320 | 0.9317 | 0.3357 | 0.2987 | 0.3364 |
| PLAD + CQ | 0.9382 | 0.9200 | 0.9386 | 0.3365 | 0.2995 | 0.3372 |
| Lasso + CQ | 0.9362 | 0.9360 | 0.9362 | 0.3405 | 0.3031 | 0.3413 |
| PCQR + Square | 0.9279 | 0.9040 | 0.9284 | 0.3080 | 0.2742 | 0.3087 |
| PLAD + Square | 0.9370 | 0.9240 | 0.9372 | 0.3218 | 0.2865 | 0.3226 |
| Lasso + Square | 0.9314 | 0.9160 | 0.9318 | 0.3115 | 0.2772 | 0.3122 |

Table 2: Coverage probabilities and average interval lengths under t-distributed noise.

| | $\overline{\mathrm{CP}}$ | $\overline{\mathrm{CP}}_{\mathcal{T}}$ | $\overline{\mathrm{CP}}_{\mathcal{T}^c}$ | $\overline{\mathrm{AL}}$ | $\overline{\mathrm{AL}}_{\mathcal{T}}$ | $\overline{\mathrm{AL}}_{\mathcal{T}^c}$ |
|---|---|---|---|---|---|---|
| PCQR + CQ | 0.9292 | 0.9200 | 0.9294 | 0.4079 | 0.3630 | 0.4088 |
| PLAD + CQ | 0.9334 | 0.8960 | 0.9342 | 0.4063 | 0.3617 | 0.4073 |
| Lasso + CQ | 0.8886 | 0.8440 | 0.8895 | 0.4051 | 0.3605 | 0.4060 |
| PCQR + Square | 0.9343 | 0.9440 | 0.9341 | 0.5738 | 0.5107 | 0.5751 |
| PLAD + Square | 0.9419 | 0.9520 | 0.9417 | 0.5892 | 0.5244 | 0.5905 |
| Lasso + Square | 0.8788 | 0.8480 | 0.8794 | 0.4511 | 0.4015 | 0.4521 |

Table 3: Coverage probabilities and average interval lengths under Cauchy noise.

| | $\overline{\mathrm{CP}}$ | $\overline{\mathrm{CP}}_{\mathcal{T}}$ | $\overline{\mathrm{CP}}_{\mathcal{T}^c}$ | $\overline{\mathrm{AL}}$ | $\overline{\mathrm{AL}}_{\mathcal{T}}$ | $\overline{\mathrm{AL}}_{\mathcal{T}^c}$ |
|---|---|---|---|---|---|---|
| PCQR + CQ | 0.9289 | 0.8720 | 0.9300 | 0.6101 | 0.5430 | 0.6115 |
| PLAD + CQ | 0.9358 | 0.8760 | 0.9371 | 0.5814 | 0.5175 | 0.5827 |
| Lasso + CQ | 0.5192 | 0.3160 | 0.5233 | 171.54 | 152.67 | 171.92 |
| PCQR + Square | 0.9494 | 0.9440 | 0.9496 | 13.794 | 12.277 | 13.825 |
| PLAD + Square | 0.9506 | 0.9440 | 0.9508 | 13.813 | 12.294 | 13.844 |
| Lasso + Square | 0.4586 | 0.3040 | 0.4618 | 2.0069 | 1.7862 | 2.0114 |

From the tables we can see that the PCQR + CQ and PLAD + CQ combinations produce almost identical results under all three distributions. Under the Gaussian noise, the Lasso + CQ combination also performs similarly as the previous two. This empirically illustrates our claim that the first-stage estimator can be chosen differently for the de-biasing procedure. Under heavy-tailed noise, the methods associated with the Lasso or square de-biasing are outperformed by our robust de-biasing methods. It is interesting to notice that PCQR + Square and PLAD + Square combinations obtain the correct coverage probabilities even under heavy-tailed noises, but at the cost of having very wide confidence intervals. In contrast, the Lasso + Square combination has poor coverage probabilities under heavy-tailed noises.

# 8 Conclusion and Discussion

In this paper we propose a robust and efficient procedure for constructing confidence intervals and testing hypotheses in high-dimensional linear models when the noise distribution is unknown. The estimator adopts the de-biasing framework and exploits the composite quantile loss function. Our procedure is robust when the noise is heavy-tailed (e.g., without the first or second moments). It also preserves efficiency in the sense that the maximum efficiency loss to the least-square approach is at most 70%, and can be more efficient than the latter in many examples. We construct uniformly valid confidence intervals for low-dimensional coordinates of the high-dimensional regression parameter, where the unknown noise is allowed to follow a wide range of distributions. Moreover, in our procedure we do not require the sparsity condition on the precision matrix of the design variables, which is imposed by most work on high-dimensional inference. We also consider the high-dimensional simultaneous test of the entire vector $\boldsymbol{\beta}^* \in \mathbb{R}^p$ using multiplier bootstrap. As a by-product, we establish the rate of convergence of the $L_1$-penalized CQR that is optimal for estimation in $s$-sparse linear regression (Raskutti et al., 2012), and a bound on the number of selected components $\widehat{s}$. Lastly, we extend this approach to the "big $N$, big $p$" regime and exploit the divide-and-conquer estimator that achieves the "oracle" property: it reduces computational complexity while preserving the asymptotic power of the oracle test based on the entire sample.

Our result in Theorem 3.2 provides an alternative method to asymptotic inference in Section 3. It is motivated by the observation that the dominating term in decomposition (3.1) has the same distribution despite different noises $\epsilon$. Recall from Theorem 3.2 that $\sqrt{n}(\widehat{\boldsymbol{\beta}}^d - \boldsymbol{\beta}^*) = \boldsymbol{Z} + \boldsymbol{W}$, where

$$\boldsymbol{Z} = n^{-1/2} \sum_{i=1}^{n} \theta_K^{-1} \mathbf{M} \boldsymbol{X}_i \Psi_{i,K},$$

where $\Psi_{i,K} = \sum_{k=1}^{K} \big(\tau_k - \mathbb{1}\{\epsilon_i \le b_k^*\}\big)$. By the definitions of $b_k^*$ and $\tau_k$, we have $\mathbb{1}\{\epsilon_i \le b_k^*\} \overset{d}{=} \mathbb{1}\{u_i \le \tau_k\}$, where $u_i$ are i.i.d. Uniform(0,1) random variables. Therefore, regardless of the specific distribution of the noise $\epsilon$, we have

$$\boldsymbol{Z} \,|\, \mathbb{X} \overset{d}{=} \frac{1}{\sqrt{n}} \sum_{i=1}^{n} \theta_K^{-1} \mathbf{M} \boldsymbol{X}_i \widetilde{\Psi}_{i,K} \,|\, \mathbb{X},$$

where $\widetilde{\Psi}_{i,K} = \sum_{k=1}^{K} \big(\tau_k - \mathbb{1}\{u_i \le \tau_k\}\big)$. Note that $\widetilde{\Psi}_{i,K}$ is essentially a sum of $K$ zero-mean Bernoulli random variables with success probability $\tau_k$. Hence, the conditional distribution of $\frac{1}{\sqrt{n}} \sum_{i=1}^{n} \theta_K^{-1} \mathbf{M} \boldsymbol{X}_i \widetilde{\Psi}_{i,K}$ given $\mathbb{X}$ is the same no matter what distribution $\epsilon$ follows. The exact distribution can be approximated using simulation. As $\boldsymbol{W}$ is asymptotically negligible, we can therefore apply inferential analysis based on distribution of $\boldsymbol{Z} \,|\, \mathbb{X}$ for any fixed set of coordinates of $\boldsymbol{\beta}^*$ as well as the high-dimensional simultaneous test (3.16). For example, for the high-dimensional simultaneous test, define

$$V_{\mathcal{G}} = \max_{j \in \mathcal{G}} \frac{1}{\sqrt{n}} \sum_{i=1}^{n} \widehat{\boldsymbol{\Theta}}_{j,\cdot} \boldsymbol{X}_i \widetilde{\Psi}_{i,K},$$

and let $c'_\alpha = \inf\{t \in \mathbb{R} : \mathbb{P}(V_{\mathcal{G}} \geq t \,|\, \mathbb{X}) \leq \alpha\}$. The test is constructed as $\Psi'_h = \mathbb{1}\{T_{\mathcal{G}} > c'_\alpha\}$. By Theorem 3.2 and discussions above, it can be easily shown that

$$\sup_{\alpha \in (0,1)} \big| \mathbb{P}(T_{\mathcal{G}} > c'_\alpha \,|\, \mathbb{X}) - \alpha \big| = o_P(1),$$

Therefore the test based on $\Psi'_h$ is valid.

### Acknowledgement

The authors are grateful for the support of NSF CAREER Award DMS1454377, NSF IIS1408910, NSF IIS1332109, NIH R01MH102339, NIH R01GM083084, and NIH R01HG06841. This work is also supported by an IBM Corporation Faculty Research Fund at the University of Chicago Booth School of Business. This work was completed in part with resources provided by the University of Chicago Research Computing Center.

## A Proof of Lemma 6.1

In this section we detail the proof of Lemma 6.1 in order to the complete the proof of Theorem 3.2. Our main strategy is to first bound $\mathbf{M}\boldsymbol{V}$ by a supremum over all $\boldsymbol{\Delta}$ (and $\boldsymbol{\delta}$) that lies within the statistical precision of $\sqrt{s(\log p)/n}$ (and $\sqrt{Ks(\log p)/n}$), and then control the latter using uniform entropy method. The main technical tools are Theorem 3.11 in Koltchinskii (2011) for bounding the expectation of a supremum, and Bousquet's concentration inequality for obtaining a tail probability for the supremum based on the bound on expectation. Both theories allow us to utilize the variance information in the summands, which works to our advantage.

We first provide the lemma for bounding supremum of a general empirical process. Let $\mathcal{F}$ be a class of functions $f$ such that $f : \mathcal{X} \to \mathbb{R}$. For any probability measure $P$ defined on $\mathcal{X}$, let

$$\|f\|_{L_2(P)} = \sqrt{\int_{x \in \mathcal{X}} f(x)^2 dP(x)}.$$

For a sequence of independent random variables $X_1, \ldots, X_n \sim P$, let $P_n$ be the empirical measure, so that $P_n(f) = \frac{1}{n}\sum_{i=1}^{n} f(X_i)$. Let $\log N(\mathcal{F}, L_2(P), \varepsilon)$ be the $\varepsilon$-covering number of $\mathcal{F}$ under the norm $\|\cdot\|_{L_2(P)}$. Denote by $E$ an envelop function of $\mathcal{F}$, that is, a measurable function such that $\sup_{f \in \mathcal{F}} |f(x)| \leq |E(x)|$ for all $x \in \mathcal{X}$. Define

$$\sigma_n^2 := \sup_{f \in \mathcal{F}} \frac{1}{n} \sum_{i=1}^{n} \mathrm{Var}[f(X_i)],$$

and

$$U_n := \sup_{f \in \mathcal{F}} \sup_{x \in \mathcal{X}} |f(x)|.$$

**Lemma A.1** (Theorem 3.11 in Koltchinskii (2011)). There exists a generic constant $C > 0$ such that

$$\mathbb{E}\left[\sup_{f\in\mathcal{F}}\left|\frac{1}{n}\sum_{i=1}^{n} f(X_i) - \mathbb{E}[f(X_i)]\right|\right] \le \frac{C}{\sqrt{n}}\mathbb{E}\left[\int_0^{2\sigma_n} \sqrt{\log N(\mathcal{F}, L_2(P_n), \varepsilon)}d\varepsilon\right].$$

Based on the above lemma, to further bound the supremum itself, we need to control the deviation from its expectation. The following Bousquet's inequality serves at our purpose:

**Lemma A.2** (Bousquet (2002)). Define

$$\mathbb{Z} := \sup_{f\in\mathcal{F}}\left|\frac{1}{n}\sum_{i=1}^{n} f(X_i) - \mathbb{E}[f(X_i)]\right|.$$

Then we have for any $t > 0$,

$$\mathbb{P}\left(\mathbb{Z} \ge \mathbb{E}[\mathbb{Z}] + t\sqrt{2(\sigma_n^2 + 2U_n\mathbb{E}[\mathbb{Z}])} + \frac{2t^2U_n}{3}\right) \le \exp(-nt^2).$$

With these technical tools, we are ready to prove Lemma 6.1. We use $C$, $c$, etc. to denote constants, whose value may change from line to line.

Recall definitions of $\psi_j(\epsilon_i, \boldsymbol{X}_i; \underset{\sim}{\boldsymbol{\Delta}}_k)$, $\mathbb{G}_j(\underset{\sim}{\boldsymbol{\Delta}})$ and $R(q, \xi_1, \xi_2)$ in (6.7) - (6.9). Moreover, define the event $\mathcal{E} = \{\widehat{\underset{\sim}{\boldsymbol{\Delta}}} \in R(c^*s, C_1\xi_n, C_3\sqrt{K}\xi_n)\}$, where $\xi_n = \sqrt{s(\log p)/n}$ and constants $c^*$, $C_1$ and $C_3$ are as defined in the statement of the lemma. By conditions in the lemma, we have $\mathbb{P}(\mathcal{E}^c) \to 0$. For any $t > 0$, it holds that

$$\begin{aligned}\mathbb{P}(\sqrt{n}\|\mathbf{M}\boldsymbol{V}\|_\infty > t) &\le \mathbb{P}(\sqrt{n}\|\mathbf{M}\boldsymbol{V}\|_\infty > t, \mathcal{E}) + \mathbb{P}(\mathcal{E}^c)\\ &\le \mathbb{P}\left(\max_{1\le j\le p}\sup_{\underset{\sim}{\boldsymbol{\Delta}}\in R(c^*s, C_1\xi_n, C_3\sqrt{K}\xi_n)}\left|\mathbb{G}_j(\underset{\sim}{\boldsymbol{\Delta}})\right| > t\right) + \mathbb{P}(\mathcal{E}^c).\end{aligned} \tag{A.1}$$

In the following, we aim to derive a tail probability for $\sup_{\underset{\sim}{\boldsymbol{\Delta}}\in R(q, C_1\xi_n, C_3\sqrt{K}\xi_n)}\left|\mathbb{G}_j(\underset{\sim}{\boldsymbol{\Delta}})\right|$, where $q = c^*s$. We first control its expectation by applying Lemma A.1. First, we provide a bound for $\frac{1}{n}\sum_{i=1}^n \mathrm{Var}\left[\sum_{k=1}^K \psi_j(\epsilon_i, \boldsymbol{X}_i; \underset{\sim}{\boldsymbol{\Delta}}_k) \,|\, \mathbb{X}\right]$. We have

$$\begin{aligned}&\mathrm{Var}\left[\sum_{k=1}^{K}\psi_j(\epsilon_i, \boldsymbol{X}_i; \underset{\sim}{\boldsymbol{\Delta}}_k) \,|\, \mathbb{X}\right]\\ &\le (\widehat{\boldsymbol{\mu}}_j^T\boldsymbol{X}_i)^2 K\sum_{k=1}^{K}\mathrm{Var}\left[\mathbb{I}(\varepsilon_i \le \underset{\sim}{\boldsymbol{X}}_i^T\underset{\sim}{\boldsymbol{\Delta}}_k + b_k^*) - \mathbb{I}(\varepsilon_i \le b_k^*) \,|\, \mathbb{X}\right]\\ &= (\widehat{\boldsymbol{\mu}}_j^T\boldsymbol{X}_i)^2 K\sum_{k=1}^{K}\Big\{F_\epsilon(\underset{\sim}{\boldsymbol{X}}_i^T\underset{\sim}{\boldsymbol{\Delta}}_k + b_k^*)(1 - F_\epsilon(\underset{\sim}{\boldsymbol{X}}_i^T\underset{\sim}{\boldsymbol{\Delta}}_k + b_k^*)) + F_\epsilon(b_k^*)(1 - F_\epsilon(b_k^*))\\ &\qquad - 2F_\epsilon\left(\min\{\underset{\sim}{\boldsymbol{X}}_i^T\underset{\sim}{\boldsymbol{\Delta}}_k + b_k^*, b_k^*\}\right) + 2F_\epsilon(\underset{\sim}{\boldsymbol{X}}_i^T\underset{\sim}{\boldsymbol{\Delta}}_k + b_k^*)F_\epsilon(b_k^*)\Big\}.\end{aligned} \tag{A.2}$$

Note that

$$
\begin{aligned}
&F_\epsilon(\utilde{\boldsymbol{X}}_i^T\utilde{\boldsymbol{\Delta}}_k + b_k^*)(1 - F_\epsilon(\utilde{\boldsymbol{X}}_i^T\utilde{\boldsymbol{\Delta}}_k + b_k^*)) + F_\epsilon(b_k^*)(1 - F_\epsilon(b_k^*)) \\
&\qquad\qquad\qquad - 2F\big(\min(\utilde{\boldsymbol{X}}_i^T\utilde{\boldsymbol{\Delta}}_k + b_k^*, b_k^*)\big) + 2F_\epsilon(\utilde{\boldsymbol{X}}_i^T\utilde{\boldsymbol{\Delta}}_k + b_k^*)F_\epsilon(b_k^*) \\
&= F_\epsilon(\utilde{\boldsymbol{X}}_i^T\utilde{\boldsymbol{\Delta}}_k + b_k^*)(1 - F_\epsilon(\utilde{\boldsymbol{X}}_i^T\utilde{\boldsymbol{\Delta}}_k + b_k^*)) - F_\epsilon(b_k^*)(1 - F_\epsilon(b_k^*)) \\
&\qquad + 2\Big(F_\epsilon(\utilde{\boldsymbol{X}}_i^T\utilde{\boldsymbol{\Delta}}_k + b_k^*)F_\epsilon(b_k^*) - F_\epsilon(b_k^*)^2\Big) + 2\Big(F_\epsilon(b_k^*) - F_\epsilon(\min\{\utilde{\boldsymbol{X}}_i^T\utilde{\boldsymbol{\Delta}}_k + b_k^*, b_k^*\})\Big) \\
&= (I) + (II) + (III).
\end{aligned} \tag{A.3}
$$

We control (I), (II) and (III) separately. For (I), we have

$$
\begin{aligned}
|(I)| &= \Big|F_\epsilon(\utilde{\boldsymbol{X}}_i^T\utilde{\boldsymbol{\Delta}}_k + b_k^*)(1 - F_\epsilon(\utilde{\boldsymbol{X}}_i^T\utilde{\boldsymbol{\Delta}}_k + b_k^*)) - F_\epsilon(b_k^*)(1 - F_\epsilon(b_k^*))\Big| \\
&\le \Big|F_\epsilon(\utilde{\boldsymbol{X}}_i^T\utilde{\boldsymbol{\Delta}}_k + b_k^*)(1 - F_\epsilon(\utilde{\boldsymbol{X}}_i^T\utilde{\boldsymbol{\Delta}}_k + b_k^*)) - F_\epsilon(\utilde{\boldsymbol{X}}_i^T\utilde{\boldsymbol{\Delta}}_k + b_k^*)(1 - F_\epsilon(b_k^*))\Big| \\
&\qquad\qquad + \Big|F_\epsilon(\utilde{\boldsymbol{X}}_i^T\utilde{\boldsymbol{\Delta}}_k + b_k^*)(1 - F_\epsilon(b_k^*)) - F_\epsilon(b_k^*)(1 - F_\epsilon(b_k^*))\Big| \\
&= F_\epsilon(\utilde{\boldsymbol{X}}_i^T\utilde{\boldsymbol{\Delta}}_k + b_k^*)\Big|F_\epsilon(b_k^*) - F_\epsilon(\utilde{\boldsymbol{X}}_i^T\utilde{\boldsymbol{\Delta}}_k + b_k^*)\Big| + (1 - F_\epsilon(b_k^*))\Big|F_\epsilon(\utilde{\boldsymbol{X}}_i^T\utilde{\boldsymbol{\Delta}}_k + b_k^*) - F_\epsilon(b_k^*)\Big| \\
&\le 2C_+\Big|\utilde{\boldsymbol{X}}_i^T\utilde{\boldsymbol{\Delta}}_k\Big|,
\end{aligned}
$$

where the last inequality is by mean value theorem and boundedness of the density function as well as the cumulative distribution function. We apply the similar deduction to (II) and (III) using mean value theorem, and it follows that $|(II)| \le 2C_+|\utilde{\boldsymbol{X}}_i^T\utilde{\boldsymbol{\Delta}}_k|$, and $|(III)| \le 2C_+|\utilde{\boldsymbol{X}}_i^T\utilde{\boldsymbol{\Delta}}_k|$. Therefore, by (A.2) and (A.3), it holds

$$
\operatorname{Var}\left[\sum_{k=1}^K \psi_j(\epsilon_i, \boldsymbol{X}_i; \utilde{\boldsymbol{\Delta}}_k)\,|\,\mathbb{X}\right] \le 6C_+\gamma_2^2 K\sum_{k=1}^K |\utilde{\boldsymbol{X}}_i^T\utilde{\boldsymbol{\Delta}}_k| \le 6C_+\gamma_2^2\left(K^2|\boldsymbol{X}_i^T\boldsymbol{\Delta}| + K\|\boldsymbol{\delta}\|_1\right).
$$

We further derive that

$$
\begin{aligned}
\frac{1}{n}\sum_{i=1}^n \operatorname{Var}\left[\sum_{k=1}^K \psi_j(\epsilon_i, \boldsymbol{X}_i; \utilde{\boldsymbol{\Delta}}_k)\,|\,\mathbb{X}\right] &\le 6C_+\gamma_2^2\left(\frac{1}{n}\sum_{i=1}^n K^2\left|\boldsymbol{X}_i^T\boldsymbol{\Delta}\right| + K\|\boldsymbol{\delta}\|_1\right) \\
&\le 6C_+\gamma_2^2\left(K^2\left(\frac{1}{n}\sum_{i=1}^n(\boldsymbol{X}_i^T\boldsymbol{\Delta})^2\right)^{1/2} + K\|\boldsymbol{\delta}\|_1\right) \\
&= 6C_+\gamma_2^2\left(K^2\left(\boldsymbol{\Delta}^T\widehat{\boldsymbol{\Sigma}}\boldsymbol{\Delta}\right)^{1/2} + K\|\boldsymbol{\delta}\|_1\right) \le CK^2\gamma_2^2\xi_n,
\end{aligned}
$$

where the second inequality is by Cauchy-Schwarz. Let $\sigma_n = CK^2\gamma_2^2\xi_n$. We proved

$$
\frac{1}{n}\sum_{i=1}^n \operatorname{Var}\left[\sum_{k=1}^K \psi_j(\epsilon_i, \boldsymbol{X}_i; \utilde{\boldsymbol{\Delta}}_k)\,|\,\mathbb{X}\right] \le \sigma_n.
$$

Next, we compute the uniform covering entropy of the following set of functions:

$$
\mathcal{F} := \left\{\sum_{k=1}^K \psi_j(\epsilon, \boldsymbol{x}; \utilde{\boldsymbol{\Delta}}_k)\,|\,\utilde{\boldsymbol{\Delta}} \in R\left(q, C_1\xi_n, C_3\sqrt{K}\xi_n\right)\right\}.
$$

For some $\mathcal{T} \subset \{1, 2, \ldots, p\}$ with $|\mathcal{T}| = q$, define

$$\mathcal{F}_{\mathcal{T}} := \left\{ \sum_{k=1}^{K} \psi_j(\epsilon, \boldsymbol{x}; \underset{\sim}{\boldsymbol{\Delta}}_k) \,|\, \text{support}(\boldsymbol{\Delta}) \subset \mathcal{T}, \|\boldsymbol{\Delta}\|_{\widehat{\boldsymbol{\Sigma}}} \le C_1 \xi_n, \|\boldsymbol{\delta}\|_2 \le C_3 \sqrt{K} \xi_n \right\}.$$

By Lemma 2.6.15 in van der Vaart and Wellner (1996), the VC dimension of $\mathcal{F}_{\mathcal{T}}$ is bounded by $q + K + 2$. As $\left| \sum_{k=1}^{K} \psi_j(\epsilon, \boldsymbol{x}; \underset{\sim}{\boldsymbol{\Delta}}_k) \right| \le K\gamma_2$, we take $E = K\gamma_2$ as the envelop function. By Lemma Theorem 2.6.7 in van der Vaart and Wellner (1996), for any probability measure $Q$ we have

$$N(\mathcal{F}_{\mathcal{T}}, L_2(Q), \varepsilon K\gamma_2) \le C(q + K + 2)(16e)^{q+K+2} \left( \frac{1}{\varepsilon} \right)^{2(q+K+1)},$$

for some generic constant $C$. Therefore the $L_2(Q)$-covering number of $\mathcal{F}$ is bounded as

$$N(\mathcal{F}, L_2(Q), \varepsilon K\gamma_2) \le C(q + K + 2)(16e)^{q+K+2} \binom{p}{q} \left( \frac{1}{\varepsilon} \right)^{2(q+K+1)} \le C \left( \frac{16e}{\varepsilon} \right)^{cq} \left( \frac{ep}{q} \right)^{q}, \tag{A.4}$$

where we used the inequalities $\binom{p}{q} \le (pe/q)^q$ and $q + K + 2 \le (16e)^{q+K+2}$. Here we treat $K$ as a large constant that does not grow with $n$. Applying Lemma A.1, we have

$$\mathbb{E} \left[ \sup_{\underset{\sim}{\boldsymbol{\Delta}} \in R(c^* s, C_1 \xi_n, C_3 \sqrt{K} \xi_n)} \left| \mathbb{G}_j(\underset{\sim}{\boldsymbol{\Delta}}) \right| \,|\, \mathbb{X} \right] \le C \mathbb{E} \left[ \int_0^{2\sigma_n} \sqrt{\log N(\mathcal{F}, L_2(P_n \,|\, \mathbb{X}), \varepsilon)} d\varepsilon \right]$$

By change of variable, it holds that

$$\begin{aligned} \int_0^{2\sigma_n} \sqrt{\log N(\mathcal{F}, L_2(P_n \,|\, \mathbb{X}), \varepsilon)} d\varepsilon &= \int_0^{2\sigma_n/(K\gamma_2)} \sqrt{\log N(\mathcal{F}, L_2(P_n \,|\, \mathbb{X}), \varepsilon K\gamma_2)} K\gamma_2 d\varepsilon \\ &\le C \int_0^{2\sigma_n/(K\gamma_2)} \sqrt{\sup_Q \log N(\mathcal{F}, L_2(Q), \varepsilon K\gamma_2)} K\gamma_2 d\varepsilon. \end{aligned}$$

By (A.4), we have $\sup_Q \log N(\mathcal{F}, L_2(Q), \varepsilon K\gamma_2) \le cq \log(p/\varepsilon)$ for some large enough constant $c$. Thereofore

$$\mathbb{E} \left[ \sup_{\underset{\sim}{\boldsymbol{\Delta}} \in R(c^* s, C_1 \xi_n, C_3 \sqrt{K} \xi_n)} \left| \mathbb{G}_j(\underset{\sim}{\boldsymbol{\Delta}}) \right| \,|\, \mathbb{X} \right] \le C' \int_0^{2\sigma_n/(K\gamma_2)} \sqrt{q \log(p/\varepsilon)} K\gamma_2 d\varepsilon$$

To evaluate the integral, note that $\int_0^t \sqrt{\log(1/\varepsilon)} d\varepsilon \le ct\sqrt{\log(1/t)}$ for some generic constant $c$. By change of variable we have $\int_0^t \sqrt{\log(p/\varepsilon)} d\varepsilon \le ct\sqrt{\log(p/t)}$. Hence

$$C' \int_0^{2\sigma_n/(K\gamma_2)} \sqrt{q \log(p/\varepsilon)} K\gamma_2 d\varepsilon \le C'' \sigma_n \sqrt{q \log(pK\gamma_2/\sigma_n)}.$$

By the definitions of $\sigma_n$ and $\xi_n$, we have $\log(pK\gamma_2/\sigma_n) = \log(p/C\sqrt{\xi_n}) = \log(p\sqrt{n}/C\sqrt{s \log p}) \lesssim \log(p \vee n)$. Therefore, we obtain

$$\mathbb{E} \left[ \sup_{\underset{\sim}{\boldsymbol{\Delta}} \in R(c^* s, C_1 \xi_n, C_3 \sqrt{K} \xi_n)} \left| \mathbb{G}_j(\underset{\sim}{\boldsymbol{\Delta}}) \right| \,|\, \mathbb{X} \right] \le C''' K\gamma_2 \sqrt{q \log(p \vee n)} \sqrt{\xi_n} =: r_n. \tag{A.5}$$

We next apply Lemma A.2 to obtain a tail probability of the above supremum. By the fact that $\sum_{k=1}^K \psi_j(\epsilon_i, \boldsymbol{X}_i; \underset{\sim}{\boldsymbol{\Delta}}_k)$ is uniformly bounded by $K\gamma_2$, we have

$$\mathbb{P}\left(\sup_{\underset{\sim}{\boldsymbol{\Delta}} \in R(c^* s, C_1\xi_n, C_3\sqrt{K}\xi_n)} \big|\mathbb{G}_j(\underset{\sim}{\boldsymbol{\Delta}})/\sqrt{n}\big| \geq \frac{r_n}{\sqrt{n}} + t\sqrt{2(\sigma_n^2 + 2K\gamma_2 r_n/\sqrt{n})} + \frac{2t^2 K\gamma_2}{3} \,|\, \mathbb{X}\right) \leq \exp(-nt^2).$$

We change the variable by replacing the $t$ with $t/\sqrt{n}$ and apply law of iterated expectation. Hence

$$\mathbb{P}\left(\sup_{\underset{\sim}{\boldsymbol{\Delta}} \in R(c^* s, C_1\xi_n, C_3\sqrt{K}\xi_n)} \big|\mathbb{G}_j(\underset{\sim}{\boldsymbol{\Delta}})\big| \geq r_n + t\sqrt{2(\sigma_n^2 + 2K\gamma_2 r_n/\sqrt{n})} + \frac{2t^2 K\gamma_2}{3\sqrt{n}}\right) \leq \exp(-t^2).$$

Plugging in $r_n = C'''K\gamma_2\sqrt{q\log(p\vee n)}\sqrt{\xi_n}$ and $\sigma_n^2 = CK^2\gamma_2^2\xi_n$ and letting $t = 2\sqrt{\log(p\vee n)}$, we obtain

$$\mathbb{P}\left(\sup_{\underset{\sim}{\boldsymbol{\Delta}} \in R(c^* s, C_1\xi_n, C_3\sqrt{K}\xi_n)} \big|\mathbb{G}_j(\underset{\sim}{\boldsymbol{\Delta}})\big| \geq CK\gamma_2(s\log(p\vee n))^{3/4}/n^{1/4}\right) \leq (p\vee n)^{-4}, \tag{A.6}$$

where we used the definition $q = c^* s$ and $\xi_n = \sqrt{s(\log p)/n}$ and the condition that $\xi_n = o(1)$. By (A.1), (A.6) and applying union bound, we obtain

$$\mathbb{P}\left(\sqrt{n}\|\mathbf{M}\boldsymbol{V}\|_\infty > CK\gamma_2(s\log(p\vee n))^{3/4}/n^{1/4}\right) \leq (p\vee n)^{-3} + \mathbb{P}(\mathcal{E}^c) \to 0, \tag{A.7}$$

as $n \to \infty$. This concludes the proof.

# B Proof of Remaining Results in Section 3

In this section, we provide detailed proofs for theorems and propositions in Section 3. We first prove the feasibility results of the CLIME estimator $\widehat{\mathbf{M}}$ in Propositions 3.7 and 3.8. We then prove Proposition 3.9 that shows the convergence rate of $\widehat{f}_k$ and Theorems 3.10, 3.11 that establish asymptotic normality for the low-dimensional coordinates of the de-biased estimator. Lastly, we prove the results of simultaneous test and multiplier bootstrap in Theorem 3.13.

## B.1 Proof of Feasibility of the Local Curvature Estimate

*Proof of Proposition 3.7.* (i) The proof is almost the same as the proof of Lemma 6.2 in Javanmard and Montanari (2014). Note that our sub-Gaussian design is equivalent to theirs under Assumption 3.6.

(ii) By definition of sub-Gaussian random variables and Assumption 3.5, $[\boldsymbol{\Sigma}^{-1}]_{\cdot,j}^T\boldsymbol{X}_i$ is sub-Gaussian with variance proxy $\|[\boldsymbol{\Sigma}^{-1}]_{\cdot,j}\|_2^2\sigma_x^2$. Moreover, $\|[\boldsymbol{\Sigma}^{-1}]_{\cdot,j}\|_2^2 \leq \|\boldsymbol{\Sigma}^{-1}\|_2^2 = \rho_{\min}^{-2}$, and so the variance proxy of $[\boldsymbol{\Sigma}^{-1}]_{\cdot,j}^T\boldsymbol{X}_i$ is upper bounded by $\rho_{\min}^{-2}\sigma_x^2$. By sub-Gaussianity, we have

$$\mathbb{P}\big(|[\boldsymbol{\Sigma}^{-1}]_{\cdot,j}^T\boldsymbol{X}_i| > t\big) \leq 2\exp(-t^2\rho_{\min}^2/\sigma_x^2).$$

Taking $t = a_2\sqrt{\log(p\vee n)}$, it follows that

$$\mathbb{P}\big(|[\boldsymbol{\Sigma}^{-1}]_{\cdot,j}^T\boldsymbol{X}_i| > a_2\sqrt{\log(p\vee n)}\big) \leq 2\exp\big(-a_2^2\rho_{\min}^2\log(p\vee n)/\sigma_x^2\big).$$

Applying union bound for $i = 1, \ldots, n$ and $j = 1, \ldots, p$ and taking $a_2 = \sqrt{3}\sigma_x/\rho_{\min}$, we have

$$\mathbb{P}\big(\|\mathbb{X}\boldsymbol{\Sigma}^{-1}\|_{\max} > a_2\sqrt{\log(p \vee n)}\big) \leq 2pn \exp\big(-a_2^2\rho_{\min}^2 \log(p \vee n)/\sigma_x^2\big) \leq 2(n \vee p)^{-1}.$$

(iii) From (ii) we have $[\boldsymbol{\Sigma}^{-1}]_{\cdot,j}^T \boldsymbol{X}_i$ for $i = 1, \ldots, n$ are independent zero-mean sub-Gaussian random variables with variance proxy $\rho_{\min}^{-2}\sigma_x^2$. Therefore, we have

$$\mathbb{P}\Big(\Big|\frac{1}{\sqrt{n}}\sum_{i=1}^{n}[\boldsymbol{\Sigma}^{-1}]_{\cdot,j}^T\boldsymbol{X}_i\Big| > t\Big) \leq 2\exp(-t^2\rho_{\min}^2/\sigma_x^2).$$

Applying union bound over $j = 1, \ldots, p$, and taking $t = a_3\sqrt{\log(p \vee n)}$, where $a_3 = \sqrt{2}\sigma_x/\rho_{\min}$, we obtain

$$\mathbb{P}\Big(\Big\|\frac{1}{\sqrt{n}}\sum_{i=1}^{n}[\boldsymbol{\Sigma}^{-1}]_{\cdot,j}^T\boldsymbol{X}_i\Big\|_\infty > a_3\sqrt{\log(p \vee n)}\Big) \leq 2p\exp(-a_3^2\rho_{\min}^2\log(p \vee n)/\sigma_x^2) \leq 2(p \vee n)^{-1}.$$

This concludes the proof. □

*Proof of Proposition 3.8.* The bound $\|\mathbb{X}\boldsymbol{\Sigma}^{-1}\|_{\max} \leq C_X$ follows directly from the Assumption 3.5a. For the other two bounds, the proof follows similarly as that of Proposition 3.7, by noting that $\boldsymbol{X}$ is sub-Gaussian using Hoeffding's Lemma. Indeed, let $\boldsymbol{X}_i' = \boldsymbol{\Sigma}^{-1}\boldsymbol{X}_i$, then $\boldsymbol{X}_i'$ is sub-Guassian with variance proxy $C_X^2$, which implies that $\boldsymbol{X} = \boldsymbol{\Sigma}\boldsymbol{X}_i'$ is sub-Gaussian with variance proxy $\rho_{\max}^2 C_X^2$. This concludes the proof. □

## B.2 Proof of Proposition 3.9

*Proof.* The proof follows similarly to that of Lemma 22 in Belloni et al. (2013a). By Taylor expansion, we have for any $\tau \in (h, 1-h)$,

$$Q(\tau + h) - Q(\tau) = hQ'(\tau) + \frac{1}{2}h^2Q''(\tau) + \frac{1}{6}h^3Q'''(\widetilde{\tau}_1)$$

and

$$Q(\tau - h) - Q(\tau) = -hQ'(\tau) + \frac{1}{2}h^2Q''(\tau) - \frac{1}{6}h^3Q'''(\widetilde{\tau}_2),$$

for some $\widetilde{\tau}_1, \widetilde{\tau}_2 \in (\tau - h, \tau + h)$. Subtracting the two equations yields

$$Q(\tau + h) - Q(\tau - h) = 2hQ'(\tau) + \frac{1}{6}h^3(Q'''(\widetilde{\tau}_1) + Q'''(\widetilde{\tau}_2)).$$

Therefore, for any $k = 1, \ldots, K$, we have

$$Q'(\tau_k) = \frac{Q(\tau_k + h) - Q(\tau_k - h)}{2h} + R_s h^2,$$

where $|R_s| \leq C$ for some constant $C$. Let

$$\widehat{Q}'(\tau_k) := \widehat{f}_k^{-1} = \frac{\widehat{Q}(\tau_k + h) - \widehat{Q}(\tau_k - h)}{2h}.$$

Therefore, on the event $\mathcal{E} := \left\{|\widehat{Q}(\tau_k+u) - Q(\tau_k+u)| \le C\sqrt{s(\log p)/n} \text{ for } u = \pm h\right\}$, we have

$$|\widehat{Q}'(\tau_k) - Q'(\tau_k)| \le \max_{u=\pm h} \frac{|\widehat{Q}(\tau_k+u) - Q(\tau_k+u)|}{h} + Ch^2 \le C_1\sqrt{\frac{s\log p}{h^2 n}} + Ch^2,$$

with probability tending to 1. Since $Q'_\epsilon(\tau_k) = f_k^{*-1}$, we have on the event $\mathcal{E}$,

$$|\widehat{f}_k - f_k^*| = \frac{|\widehat{Q}'(\tau_k) - Q'(\tau_k)|}{\widehat{Q}'(\tau_k)Q'(\tau_k)} \le |\widehat{f}_k f_k^*| \left(C_1\sqrt{\frac{s\log p}{h^2 n}} + Ch^2\right).$$

Therefore, by Assumption 3.1 and the scaling condition that $\sqrt{s(\log p)/h^2 n} = o(1)$ and $h = o(1)$, we have $|\widehat{f}_k| \le |\widehat{f}_k| C_+ o(1) + C_+$. Therefore $|\widehat{f}_k|$ is uniformly bounded, and by the above equation we obtain

$$\mathbb{P}\left(|\widehat{f}_k - f_k^*| \ge C\left(\sqrt{\frac{s\log p}{h^2 n}} + h^2\right)\right) \le \mathbb{P}(\mathcal{E}^c) \to 0,$$

which completes the proof. □

## B.3 Proof of Theorem 3.10

Before proving Theorem 3.10, we need the following preliminary lemma:

**Lemma B.1.** For all $j = 1, \dots, p$, let $\widehat{\boldsymbol{\mu}}_j$ be a solution to the minimization problem (2.5). Then we have with probability at least $1 - p^{-1}$ that

$$\widehat{\boldsymbol{\mu}}_j^T \widehat{\boldsymbol{\Sigma}} \widehat{\boldsymbol{\mu}}_j \ge \frac{(1-\gamma_1)^2}{\boldsymbol{\Sigma}_{jj} + c\sigma_x^2\sqrt{(\log p)/n}},$$

for some universal constant $c$.

The proof of Lemma B.1 is deferred to Appendix E. The conclusion of Theorem 3.10 then follows by combining Theorems 3.2, 3.3 and Lemma B.1 as shown below.

*Proof of Theorem 3.10.* By Theorem 3.2, we have

$$\frac{\sqrt{n}(\widehat{\beta}_j^d - \beta_j^*)}{(\widehat{\boldsymbol{\mu}}_j^T \widehat{\boldsymbol{\Sigma}} \widehat{\boldsymbol{\mu}}_j)^{1/2}} = \frac{Z_j}{(\widehat{\boldsymbol{\mu}}_j^T \widehat{\boldsymbol{\Sigma}} \widehat{\boldsymbol{\mu}}_j)^{1/2}} + \frac{W_j}{(\widehat{\boldsymbol{\mu}}_j^T \widehat{\boldsymbol{\Sigma}} \widehat{\boldsymbol{\mu}}_j)^{1/2}}. \tag{B.1}$$

By Lemma B.1, and the fact that $\boldsymbol{\Sigma}_{jj} \le \rho_{\max}$ for all $j$, we have with probability tending to 1 that

$$\frac{|W_j|}{(\widehat{\boldsymbol{\mu}}_j^T \widehat{\boldsymbol{\Sigma}} \widehat{\boldsymbol{\mu}}_j)^{1/2}} \le \frac{|W_j|(\rho_{\max} + c\sigma_x^2\sqrt{(\log p)/n})^{1/2}}{1-\gamma_1}. \tag{B.2}$$

We have $|W_j| = o_P(1)$ by Theorem 3.2 and $\gamma_1 = a_1\sqrt{(\log p)/n} \to 0$. Hence $|W_j|/(\widehat{\boldsymbol{\mu}}_j^T \widehat{\boldsymbol{\Sigma}} \widehat{\boldsymbol{\mu}}_j)^{1/2} = o_P(1)$. Moreover, $\theta_K Z_j/(\widehat{\boldsymbol{\mu}}_j^T \widehat{\boldsymbol{\Sigma}} \widehat{\boldsymbol{\mu}}_j)^{1/2} \rightsquigarrow N(0, \sigma_K^2)$ by Theorem 3.3, and $\widehat{\theta}_K/\theta_K \to 1$ in probability by (2.7). Therefore, applying Slusky's theorem, we have

$$\frac{\sqrt{n}(\widehat{\beta}_j^d - \beta_j^*)}{\widehat{\theta}_K^{-1}(\widehat{\boldsymbol{\mu}}_j^T \widehat{\boldsymbol{\Sigma}} \widehat{\boldsymbol{\mu}}_j)^{1/2}} \rightsquigarrow N(0, \sigma_K^2),$$

which concludes the proof. □

## B.4 Proof of Theorem 3.11

*Proof.* By Theorem 3.2, we have

$$\sqrt{n}\sigma_K^{-1}\widehat{\theta}_K\mathbf{A}^{-1/2}(\mathbf{Q}\widehat{\boldsymbol{\beta}}^d-\mathbf{Q}\boldsymbol{\beta}^*)=\sigma_K^{-1}\widehat{\theta}_K\mathbf{A}^{-1/2}\mathbf{Q}\boldsymbol{Z}+\sigma_K^{-1}\widehat{\theta}_K\mathbf{A}^{-1/2}\mathbf{Q}\boldsymbol{W}. \tag{B.3}$$

We first prove that the first term on the R.H.S weakly converges to a multivariate normal distribution. The proof is similar to that of Theorem 3.3 coupled with the application of Cramer-Wold device. For any $\boldsymbol{u}\in\mathbb{R}^q$, it holds

$$\sigma_K^{-1}\widehat{\theta}_K\boldsymbol{u}^T\mathbf{A}^{-1/2}\mathbf{Q}\boldsymbol{Z}=\frac{1}{\sqrt{n}}\sum_{i=1}^n\sigma_K^{-1}\theta_K^{-1}\widehat{\theta}_K\boldsymbol{u}^T\mathbf{A}^{-1/2}\mathbf{Q}\mathbf{M}\boldsymbol{X}_i\Psi_{i,K}.$$

Let $\xi_i=\sigma_K^{-1}\boldsymbol{u}^T\mathbf{A}^{-1/2}\mathbf{Q}\mathbf{M}\boldsymbol{X}_i\Psi_{i,K}$, then we have $\sigma_K^{-1}\widehat{\theta}_K\boldsymbol{u}^T\mathbf{A}^{-1/2}\mathbf{Q}\boldsymbol{Z}=n^{-1/2}\sum_{i=1}^n\widehat{\theta}_K/\theta_K\xi_i$. Similar to the arguments in the proof of Theorem 3.3, we have $\mathbb{E}[\xi_i\,|\,\mathbb{X}]=0$, and

$$\begin{aligned}s_n^2&:=\sum_{i=1}^n\mathrm{Var}\left(\xi_i\,|\,\mathbb{X}\right)\\&=\sigma_K^{-2}\sum_{i=1}^n\boldsymbol{u}^T\mathbf{A}^{-1/2}\mathbf{Q}\mathbf{M}\boldsymbol{X}_i\boldsymbol{X}_i^T\mathbf{M}^T\mathbf{Q}^T\mathbf{A}^{-1/2}\boldsymbol{u}\,\mathrm{Var}\left(\Psi_{i,K}\,|\,\mathbb{X}\right)\\&=n\boldsymbol{u}^T\mathbf{A}^{-1/2}\mathbf{Q}\mathbf{M}\widehat{\boldsymbol{\Sigma}}\mathbf{M}^T\mathbf{Q}^T\mathbf{A}^{-1/2}\boldsymbol{u}\\&=n\|\boldsymbol{u}\|_2^2,\end{aligned}$$

where we used the definition that $\mathbf{A}=\mathbf{Q}\mathbf{M}\widehat{\boldsymbol{\Sigma}}\mathbf{M}^T\mathbf{Q}^T$. Moreover, each $\xi_i$ can be controlled by

$$|\xi_i|=\sigma_K^{-1}|\Psi_{i,K}|\left|\boldsymbol{u}^T\mathbf{A}^{-1/2}\mathbf{Q}\mathbf{M}\boldsymbol{X}_i\right|\le\sigma_K^{-1}|\Psi_{i,K}|\left\|\boldsymbol{u}^T\mathbf{A}^{-1/2}\right\|_2\left\|\mathbf{Q}\mathbf{M}\boldsymbol{X}_i\right\|_2. \tag{B.4}$$

By Lemma F.1, we have $\left\|\mathbf{Q}\mathbf{M}\boldsymbol{X}_i\right\|_2\le C_q\|\mathbf{M}\boldsymbol{X}_i\|_\infty\le C_q\gamma_2$. Also, by Lemma F.2 and Lemma B.1 we have

$$\begin{aligned}\|\mathbf{A}\|_2=\left\|\mathbf{Q}\mathbf{M}\widehat{\boldsymbol{\Sigma}}\mathbf{M}^T\mathbf{Q}^T\right\|_2&\ge S_{\min}^2(\mathbf{Q})\min_j\widehat{\boldsymbol{\mu}}_j^T\widehat{\boldsymbol{\Sigma}}\widehat{\boldsymbol{\mu}}_j\\&\ge S_{\min}^2(\mathbf{Q})\frac{(1-\gamma_1)^2}{\rho_{\max}+c\sigma_x^2\sqrt{(\log p)/n}},\end{aligned} \tag{B.5}$$

where $S_{\min}(\mathbf{Q})$ is the singular value of $\mathbf{Q}$. Note that $S_{\min}(\mathbf{Q})$ is positive as $\mathbf{Q}$ is full rank. Therefore,

$$\left\|\boldsymbol{u}^T\mathbf{A}^{-1/2}\right\|_2\le\|\boldsymbol{u}\|_2\frac{\left(\rho_{\max}+c\sigma_x^2\sqrt{(\log p)/n}\right)^{1/2}}{S_{\min}(\mathbf{Q})|1-\gamma_1|}=:c_n\|\boldsymbol{u}\|_2.$$

We have $c_n\to c_\infty$ for some constant $c_\infty>0$ when $n\to\infty$. Hence by (B.4), we have

$$|\xi_i|\le\sigma_k^{-1}C_qc_n\|\boldsymbol{u}\|_2\gamma_2|\Psi_{i,K}|.$$

Therefore, for any $\varepsilon > 0$, it holds that

$$
\begin{aligned}
&\lim_{n\to\infty} \frac{1}{s_n^2} \sum_{i=1}^n \mathbb{E}\left[\xi_i^2 \,\mathbb{1}\{|\xi_i| > \varepsilon s_n\} \,|\, \mathbb{X}\right] \\
&= \|\boldsymbol{u}\|_2^{-2} \lim_{n\to\infty} \frac{1}{n} \sum_{i=1}^n \mathbb{E}\left[\xi_i^2 \,\mathbb{1}\{|\Psi_{i,K}| > \varepsilon C_q^{-1} c_n^{-1} \sigma_K \gamma_2^{-1} \sqrt{n}\} \,|\, \mathbb{X}\right] \\
&= \sigma_K^{-2} \|\boldsymbol{u}\|_2^{-2} \lim_{n\to\infty} \boldsymbol{u}^T \mathbf{A}^{-1/2} \mathbf{Q} \mathbf{M} \widehat{\boldsymbol{\Sigma}} \mathbf{M}^T \mathbf{Q}^T \mathbf{A}^{-1/2} \boldsymbol{u}^T \mathbb{E}\left[(\Psi_{1,K})^2 \,\mathbb{1}\{|\Psi_{1,K}| > \varepsilon C_q^{-1} c_n^{-1} \sigma_K \gamma_2^{-1} \sqrt{n}\}\right] \\
&= \lim_{n\to\infty} \sigma_K^{-2} \mathbb{E}\left[(\Psi_{1,K})^2 \,\mathbb{1}\{|\Psi_{1,K}| > \varepsilon C_q^{-1} c_n^{-1} \gamma_2^{-1} \sigma_K \sqrt{n}\}\right] \\
&\leq \lim_{n\to\infty} \varepsilon^{-2} C_q^2 c_n^2 \sigma_K^{-4} \gamma_2^2 n^{-1} \mathbb{E}\left[(\Psi_{1,K})^4\right] \\
&= 0,
\end{aligned}
$$

where the last equality is by the boundedness of $\Psi_{1,K}$ and the fact that $\gamma_2 n^{-1/2} = o(1)$ under the choice $\gamma_2 = a_2\sqrt{\log(p \vee n)}$ and the scaling conditions stated in the theorem. Therefore, by Lindeberg CLT and the fact that $\boldsymbol{Z}$ does not contain $\boldsymbol{\beta}^*$, we have

$$
\lim_{n\to\infty} \sup_{\boldsymbol{\beta}^* \in \mathbb{R}^p, \|\boldsymbol{\beta}^*\|_0 \leq s} \left| \mathbb{P}_* \left( \sigma_K^{-1} \theta_K \boldsymbol{u}^T \mathbf{A}^{-1/2} \mathbf{Q} \boldsymbol{Z} \leq x \right) - \Phi(x) \right| = 0,
$$

for any $x \in \mathbb{R}$. By the arbitrariness of $\boldsymbol{u}$ and Cramer-Wold device, we have

$$
\lim_{n\to\infty} \sup_{\boldsymbol{\beta}^* \in \mathbb{R}^p, \|\boldsymbol{\beta}^*\|_0 \leq s} \left| \mathbb{P}_* \left( \sigma_K^{-1} \theta_K \mathbf{A}^{-1/2} \mathbf{Q} \boldsymbol{Z} \leq \boldsymbol{x} \right) - \boldsymbol{\Phi}(\boldsymbol{x}) \right| = 0, \tag{B.6}
$$

for any $\boldsymbol{x} \in \mathbb{R}^q$. Now by (B.3), we have

$$
\begin{aligned}
&\mathbb{P}_* \left( \sqrt{n} \sigma_K^{-1} \widehat{\theta}_K \mathbf{A}^{-1/2} (\mathbf{Q} \widehat{\boldsymbol{\beta}}^d - \mathbf{Q} \boldsymbol{\beta}^*) \leq \boldsymbol{x} \right) \\
&\quad = \mathbb{P}_* \left( \sigma_K^{-1} \theta_K \mathbf{A}^{-1/2} \mathbf{Q} \boldsymbol{Z} + \sigma_K^{-1} \theta_K \mathbf{A}^{-1/2} \mathbf{Q} \boldsymbol{W} \leq (1 + r_t) \boldsymbol{x} \right) + \mathbb{P}_* \left( \left| \frac{\widehat{\theta}_K}{\theta_K} - 1 \right| \geq r_t \right).
\end{aligned} \tag{B.7}
$$

For the first probability on the R.H.S. of (B.7), we have for any $\varepsilon \in \mathbb{R}$, it holds

$$
\begin{aligned}
&\mathbb{P}_* \left( \sigma_K^{-1} \theta_K \mathbf{A}^{-1/2} \mathbf{Q} \boldsymbol{Z} + \sigma_K^{-1} \theta_K \mathbf{A}^{-1/2} \mathbf{Q} \boldsymbol{W} \leq (1 + r_t) \boldsymbol{x} \right) \\
&\leq \mathbb{P}_* \left( \sigma_K^{-1} \theta_K \mathbf{A}^{-1/2} \mathbf{Q} \boldsymbol{Z} \leq (1 + r_t) \boldsymbol{x} + \varepsilon \mathbf{1}_q \right) + \mathbb{P}_* \left( \exists \text{ some } \ell \in [q] \text{ that } \sigma_K^{-1} \theta_K [\mathbf{A}^{-1/2} \mathbf{Q} \boldsymbol{W}]_\ell \geq \varepsilon \right) \\
&\leq \mathbb{P}_* \left( \sigma_K^{-1} \theta_K \mathbf{A}^{-1/2} \mathbf{Q} \boldsymbol{Z} \leq (1 + r_t) \boldsymbol{x} + \varepsilon \mathbf{1}_q \right) + q \mathbb{P}_* \left( \sigma_K^{-1} \theta_K \|\mathbf{A}^{-1/2} \mathbf{Q} \boldsymbol{W}\|_\infty \geq \varepsilon \right).
\end{aligned} \tag{B.8}
$$

For the first probability in (B.8), by (B.6) and continuity of the function $\boldsymbol{\Phi}(\cdot)$, we have

$$
\limsup_{n\to\infty} \sup_{\boldsymbol{\beta}^* \in \mathbb{R}^p, \|\boldsymbol{\beta}^*\|_0 \leq s} \left\{ \mathbb{P}_* \left( \sigma_K^{-1} \theta_K \mathbf{A}^{-1/2} \mathbf{Q} \boldsymbol{Z} \leq (1 + r_t) \boldsymbol{x} + \varepsilon \mathbf{1}_q \right) - \boldsymbol{\Phi}(\boldsymbol{x} + \varepsilon \mathbf{1}_q) \right\} \leq 0. \tag{B.9}
$$

For the second probability in (B.8), we have

$$
\begin{aligned}
\|\sigma_K^{-1} \theta_K \mathbf{A}^{-1/2} \mathbf{Q} \boldsymbol{W}\|_\infty &\leq \sigma_K^{-1} \theta_K \|\mathbf{A}^{-1/2} \mathbf{Q} \boldsymbol{W}\|_2 \\
&\leq \sigma_K^{-1} \theta_K C_q \|\mathbf{A}^{-1/2}\|_2 \|\boldsymbol{W}\|_\infty.
\end{aligned} \tag{B.10}
$$

where the first inequality is by the trivial bound that $\|\cdot\|_\infty \le \|\cdot\|_2$, and the last inequality is by Lemma F.1. Recall $\|\mathbf{A}^{-1/2}\|_2 \le c_n$. Moreover, following the same argument as the proof of Theorem 3.2 for obtaining $\|\boldsymbol{W}\|_\infty = o_P(1)$, we can prove the uniform version that $\lim_{n\to\infty}\sup_{\boldsymbol{\beta}^*\in\mathbb{R}^p,\|\boldsymbol{\beta}^*\|_0\le s}\mathbb{P}(\|\boldsymbol{W}\|_\infty \ge \epsilon) = 0$ under the condition (3.11). Hence by (B.10), we conclude

$$\limsup_{n\to\infty}\sup_{\boldsymbol{\beta}^*\in\mathbb{R}^p,\|\boldsymbol{\beta}^*\|_0\le s} q\mathbb{P}_*\left(\sigma_K^{-1}\theta_K\|\mathbf{A}^{-1/2}\mathbf{Q}\boldsymbol{W}\|_\infty \ge \varepsilon\right) = 0. \tag{B.11}$$

Combining (B.7), (B.8), (B.9) and (B.11) and by the property of $\widehat{\theta}_K^{-1}$ defined as per (3.12), we have

$$\begin{aligned}\limsup_{n\to\infty}\sup_{\boldsymbol{\beta}^*\in\mathbb{R}^p,\|\boldsymbol{\beta}^*\|_0\le s}&\left\{\mathbb{P}_*\left(\sqrt{n}\sigma_K^{-1}\widehat{\theta}_K\mathbf{A}^{-1/2}(\mathbf{Q}\widehat{\boldsymbol{\beta}}^d-\mathbf{Q}\boldsymbol{\beta}^*)\le\boldsymbol{x}\right)-\boldsymbol{\Phi}(\boldsymbol{x})\right\}\\&\le\left|\boldsymbol{\Phi}(\boldsymbol{x}+\varepsilon\mathbf{1}_q)-\boldsymbol{\Phi}(\boldsymbol{x})\right|.\end{aligned}\tag{B.12}$$

Taking $\epsilon\to 0$, the R.H.S. of (B.12) tends to 0. Hence

$$\limsup_{n\to\infty}\sup_{\boldsymbol{\beta}^*\in\mathbb{R}^p,\|\boldsymbol{\beta}^*\|_0\le s}\left\{\mathbb{P}\left(\sqrt{n}\sigma_K^{-1}\widehat{\theta}_K\mathbf{A}^{-1/2}(\mathbf{Q}\widehat{\boldsymbol{\beta}}^d-\mathbf{Q}\boldsymbol{\beta}^*)\le\boldsymbol{x}\right)-\boldsymbol{\Phi}(\boldsymbol{x})\right\}\le 0.$$

Applying symmetric argument to above, we obtain

$$\liminf_{n\to\infty}\inf_{\boldsymbol{\beta}^*\in\mathbb{R}^p,\|\boldsymbol{\beta}^*\|_0\le s}\left\{\mathbb{P}_*\left(\sqrt{n}\sigma_K^{-1}\widehat{\theta}_K\mathbf{A}^{-1/2}(\mathbf{Q}\widehat{\boldsymbol{\beta}}^d-\mathbf{Q}\boldsymbol{\beta}^*)\le\boldsymbol{x}\right)-\boldsymbol{\Phi}(\boldsymbol{x})\right\}\ge 0.$$

Then (3.13) follows by combining the above two inequalities. □

## B.5 Proof of Theorem 3.13

Before presenting the proof, we define the following preliminaries: for any $\mathcal{G}\subset\{1,2,\ldots,p\}$ with $|\mathcal{G}| = d$, let

$$T_{0,\mathcal{G}} := \max_{j\in\mathcal{G}}\frac{1}{\sqrt{n}}\sum_{i=1}^n\theta_K^{-1}\Psi_{i,K}[\boldsymbol{\Sigma}^{-1}\boldsymbol{X}_i]_j.$$

Furthermore, let

$$U_{0,\mathcal{G}} := \max_{j\in\mathcal{G}} n^{-1/2}\sum_{i=1}^n\Gamma_{i,j},$$

where $\{\boldsymbol{\Gamma}_i = (\Gamma_{i,1},\ldots\Gamma_{i,p})\}$ for $i = 1,\ldots,n$ is a sequence of mean zero independent Gaussian vectors with $\mathbb{E}[\boldsymbol{\Gamma}_i\boldsymbol{\Gamma}_i^T] = \theta_K^{-2}\sigma_K^2\boldsymbol{\Sigma}^{-1}$. Lastly, it is useful to recall that

$$U_{\mathcal{G}} = \max_{j\in\mathcal{G}}\frac{1}{\sqrt{n}}\sum_{i=1}^n\sigma_K\widehat{\theta}_K^{-1}\widehat{\boldsymbol{\mu}}_j'^T\boldsymbol{X}_ig_i,$$

where $\{g_i\}$ is a sequence of i.i.d. $N(0,1)$ random variables, and $c_\alpha = \inf\{t\in\mathbb{R}:\mathbb{P}(U_{\mathcal{G}} > t\,|\,\mathbb{X})\le\alpha\}$.

We approximate $T_{\mathcal{G}}$ by $T_{0,\mathcal{G}}$, and next apply Gaussian approximation to $T_{0,\mathcal{G}}$ and $W_{0,\mathcal{G}}$. Then, we argue that $W_{0,\mathcal{G}}$ and $W_{\mathcal{G}}$ are close. Hence we can approximate the quantiles of $T_{\mathcal{G}}$ by those of $W_{\mathcal{G}}$. We first provide the following two technical lemmas, whose proofs are deferred to Appendix E.

**Lemma B.2.** Suppose Assumptions 3.5 and 3.6 hold. For any $\mathcal{G} \subset \{1, 2, \ldots, p\}$ with $|\mathcal{G}| = d$, if $(\log(dn))^7/n \le C_1 n^{-c_1}$ for some constants $c_1, C_1 > 0$, then we have

$$\sup_{x \in \mathbb{R}} \Big| \mathbb{P}\big(T_{0,\mathcal{G}} \le x\big) - \mathbb{P}\big(U_{0,\mathcal{G}} \le x\big) \Big| \le n^{-c},$$

for some constant $c > 0$.

**Lemma B.3.** Suppose Assumptions 3.5 and 3.6 hold, and $\|\mathbf{M}' - \mathbf{\Sigma}^{-1}\|_{1,\max} \lesssim s_1\sqrt{(\log p)/n}$ with probability tending to 1. Moreover, suppose $s_1$ satisfy $s_1 \log p \sqrt{\log(d \vee n)}/\sqrt{n} = o(1)$. Then there exist $\zeta_1$ and $\zeta_2$ such that

$$\mathbb{P}\left( \max_{j \in \mathcal{G}} \frac{1}{\sqrt{n}} \left| \sum_{i=1}^n \theta_K^{-1} \Psi_{i,K} \widehat{\boldsymbol{\mu}}_j'^T \boldsymbol{X}_i - \sum_{i=1}^n \theta_K^{-1} \Psi_{i,K} [\mathbf{\Sigma}^{-1} \boldsymbol{X}_i]_j \right| > \zeta_1 \right) \le \zeta_2,$$

where $\zeta_1 \sqrt{1 \vee \log(d/\zeta_1)} = o(1)$ and $\zeta_2 = o(1)$.

We now present the detailed proof of Theorem 3.13.

*Proof of Theorem 3.13.* By Theorem 3.2, we have

$$\sqrt{n}(\widehat{\boldsymbol{\beta}}^d - \boldsymbol{\beta}^*) = \boldsymbol{Z} + \boldsymbol{W}, \tag{B.13}$$

where $\boldsymbol{Z} = n^{-1/2} \sum_{i=1}^n \theta_K^{-1} \mathbf{M}' \boldsymbol{X}_i \Psi_{i,K}$ and $\|\boldsymbol{W}\|_\infty = o_P(1)$. By (B.13) and the definitions of $T_{\mathcal{G}}$ and $T_{0,\mathcal{G}}$, we have

$$|T_{\mathcal{G}} - T_{0,\mathcal{G}}| \le \max_{j \in \mathcal{G}} \frac{1}{\sqrt{n}} \left| \sum_{i=1}^n \theta_K^{-1} \Psi_{i,K} \widehat{\boldsymbol{\mu}}_j'^T \boldsymbol{X}_i - \sum_{i=1}^n \theta_K^{-1} \Psi_{i,K} [\mathbf{\Sigma}^{-1} \boldsymbol{X}_i]_j \right| + \|\boldsymbol{W}\|_\infty,$$

where we used the fact that $\max_j a_j - \max_j b_j \le \max_j |a_j - b_j|$ for any two finite sequences $\{a_j\}, \{b_j\}$. By the above inequality and Lemma B.3, there exist $\zeta_1$ and $\zeta_2$ such that

$$\mathbb{P}\left(|T_{\mathcal{G}} - T_{0,\mathcal{G}}| \ge \zeta_1\right) \le \zeta_2, \tag{B.14}$$

where $\zeta_1 \sqrt{1 \vee \log(d/\zeta_1)} = o(1)$ and $\zeta_2 = o(1)$.

We next turn to bound the distance between quantiles of $U_{\mathcal{G}}$ and $U_{0,\mathcal{G}}$. Let $c_{0,\mathcal{G}}(\alpha) := \inf\{t \in \mathbb{R} : \mathbb{P}(U_{0,\mathcal{G}} \le t) \ge 1 - \alpha\}$, and

$$\Lambda := \max_{1 \le j,\ell \le p} \theta_K^{-2} \sigma_K^2 \left| \widehat{\boldsymbol{\mu}}_j'^T \widehat{\mathbf{\Sigma}} \widehat{\boldsymbol{\mu}}_\ell' - [\mathbf{\Sigma}^{-1}]_{j\ell} \right|. \tag{B.15}$$

Define $\boldsymbol{V} = n^{-1/2} \sum_{i=1}^n \sigma_K \widehat{\theta}_K^{-1} \mathbf{M}' \boldsymbol{X}_i g_i$ so $U_{\mathcal{G}} = \max_{j \in \mathcal{G}} \boldsymbol{V}_j$. Conditional on $\mathbb{X}$, $\boldsymbol{V}$ is a $p$-dimensional Gaussian random vector with mean $\mathbf{0}$ and covariance $\theta_K^{-2} \sigma_K^2 \mathbf{M}' \widehat{\mathbf{\Sigma}} \mathbf{M}'^T$. Hence $\Lambda$ is essentially the max norm of the difference between covariances of $\boldsymbol{V} \,|\, \mathbb{X}$ and $n^{-1/2} \sum_{i=1}^n \mathbf{\Gamma}_i$, which are both Gaussian. Therefore, using Gaussian comparison (Lemma 3.1 in Chernozhukov et al. (2013)) and applying the same argument as in the proof of Lemma 3.2 in Chernozhukov et al. (2013), we obtain for any $\nu > 0$, that

$$\mathbb{P}\left(c_{0,\mathcal{G}}(\alpha) \le c_{\mathcal{G}}(\alpha + \pi(\nu))\right) \ge 1 - \mathbb{P}(\Lambda > \nu), \tag{B.16}$$
$$\mathbb{P}\left(c_{\mathcal{G}}(\alpha) \le c_{0,\mathcal{G}}(\alpha + \pi(\nu))\right) \ge 1 - \mathbb{P}(\Lambda > \nu), \tag{B.17}$$

where $\pi(\nu) := c\nu^{1/3}(1 \vee \log(p/\nu))^{2/3}$ with some generic constant $c > 0$. By Lemma B.2, we have

$$\sup_{\alpha \in (0,1)} |\mathbb{P}(T_{0,\mathcal{G}} > c_{\mathcal{G}}(\alpha)) - \alpha| \leq \sup_{\alpha \in (0,1)} |\mathbb{P}(U_{0,\mathcal{G}} > c_{\mathcal{G}}(\alpha)) - \alpha| + n^{-c}. \tag{B.18}$$

To further control $\mathbb{P}(U_{0,\mathcal{G}} > c_{\mathcal{G}}(\alpha))$, we define

$$\mathcal{E}_1 = \{c_{0,\mathcal{G}}(\alpha - \pi(\nu)) \leq c_{\mathcal{G}}(\alpha)\} \quad \text{and} \quad \mathcal{E}_2 = \{c_{\mathcal{G}}(\alpha) \leq c_{0,\mathcal{G}}(\alpha + \pi(\nu))\}.$$

We have

$$\begin{aligned}
\mathbb{P}\big(U_{0,\mathcal{G}} > c_{\mathcal{G}}(\alpha)\big) &= \mathbb{P}\big(U_{0,\mathcal{G}} > c_{\mathcal{G}}(\alpha), \mathcal{E}_1\big) + \mathbb{P}\big(U_{0,\mathcal{G}} > c_{\mathcal{G}}(\alpha), \mathcal{E}_1^c\big) \\
&\leq \mathbb{P}\big(U_{0,\mathcal{G}} > c_{0,\mathcal{G}}(\alpha - \pi(\nu))\big) + \mathbb{P}(\mathcal{E}_1^c) \\
&\leq \alpha - \pi(\nu) + \mathbb{P}(\Lambda > \nu),
\end{aligned}$$

where the last inequality is by the definition of $c_{0,\mathcal{G}}(\alpha)$ and (B.16). Similarly, we have

$$\begin{aligned}
\mathbb{P}\big(U_{0,\mathcal{G}} > c_{\mathcal{G}}(\alpha)\big) &= 1 - \mathbb{P}\big(U_{0,\mathcal{G}} \leq c_{\mathcal{G}}(\alpha)\big) \\
&= 1 - \mathbb{P}\big(U_{0,\mathcal{G}} \leq c_{\mathcal{G}}(\alpha), \mathcal{E}_2\big) - \mathbb{P}\big(U_{0,\mathcal{G}} \leq c_{\mathcal{G}}(\alpha), \mathcal{E}_2^c\big) \\
&\geq 1 - \mathbb{P}\big(U_{0,\mathcal{G}} \leq c_{0,\mathcal{G}}(\alpha + \pi(\nu))\big) - \mathbb{P}(\mathcal{E}_2^c) \\
&\geq \alpha + \pi(\nu) - \mathbb{P}(\Lambda > \nu),
\end{aligned}$$

where the last inequality is by the definition of $c_{0,\mathcal{G}}(\alpha)$ and (B.17). Therefore we conclude

$$\big|\mathbb{P}\big(U_{0,\mathcal{G}} > c_{\mathcal{G}}(\alpha)\big) - \alpha\big| \leq |\pi(\nu) - \mathbb{P}(\Lambda > \nu)|$$

and it follows from (B.18) that

$$\sup_{\alpha \in (0,1)} \big|\mathbb{P}\big(T_{0,\mathcal{G}} > c_{\mathcal{G}}(\alpha)\big) - \alpha\big| \leq \pi(\nu) + \mathbb{P}(\Lambda > \nu) + n^{-c}. \tag{B.19}$$

Define the event $\mathcal{E}_3 = \{|T_{0,\mathcal{G}} - T_{\mathcal{G}}| \leq \zeta_1\}$. By (B.14), we have $\mathbb{P}(\mathcal{E}_3^c) \leq \zeta_2$. Hence, we deduce that for any $\alpha$,

$$\begin{aligned}
\mathbb{P}(T_{\mathcal{G}} \geq c_G(\alpha)) - \alpha &\leq \mathbb{P}(T_{\mathcal{G}} \geq c_G(\alpha), \mathcal{E}_3) + \mathbb{P}(\mathcal{E}_3^c) - \alpha \\
&\leq \mathbb{P}(T_{0,\mathcal{G}} \geq c_G(\alpha) - \zeta_1) + \zeta_2 - \alpha, \\
&\leq \mathbb{P}(T_{0,\mathcal{G}} \geq c_G(\alpha)) + C\zeta_1\sqrt{1 \vee \log(d/\zeta_1)} + \zeta_2 - \alpha \\
&\leq \pi(\nu) + \mathbb{P}(\Lambda > \nu) + n^{-c} + C\zeta_1\sqrt{1 \vee \log(d/\zeta_1)} + \zeta_2,
\end{aligned}$$

where the second last inequality is by Corollary 16 of Wasserman (2014) (Gaussian anti-concentration). By similar arguments, we get the same bound for $\alpha - \mathbb{P}(T_{\mathcal{G}} \geq c_G(\alpha))$, so we have

$$\sup_{\alpha} \big|\mathbb{P}(T_{\mathcal{G}} \geq c_G(\alpha)) - \alpha\big| \leq \pi(\nu) + \mathbb{P}(\Lambda > \nu) + n^{-c} + C\zeta_1\sqrt{1 \vee \log(d/\zeta_1)} + \zeta_2.$$

To complete the proof, we bound $\Lambda$. We first obtain a bound for $\|\mathbf{M}' - \boldsymbol{\Sigma}^{-1}\|_{\max}$. Note that for all $j = 1, \ldots, p$, $X_j$ are zero-mean sub-Gaussian random variables with variance proxy $\sigma_x^2$, hence it holds that $\mathbb{E}[\exp(X_j^2/6\sigma_x^2)] \leq 2$ (See Page 47 of van Handel (2014)). Therefore, by Theorem 4 of Cai et al.

(2011), we have $\|\mathbf{M}' - \boldsymbol{\Sigma}^{-1}\|_{\max} \le 4a_1 R\sqrt{(\log p)/n}$, and so $\|\mathbf{M}' - \boldsymbol{\Sigma}^{-1}\|_{1,\max} \le 4a_1 R s_1\sqrt{(\log p)/n}$ with probability at least $1-4p^{-3}$. It follows that

$$
\begin{aligned}
\|\mathbf{M}'\widehat{\boldsymbol{\Sigma}}\mathbf{M}'^T - \boldsymbol{\Sigma}^{-1}\|_{\max} &\le \|\mathbf{M}'\widehat{\boldsymbol{\Sigma}}\mathbf{M}'^T - \mathbf{M}'\widehat{\boldsymbol{\Sigma}}\boldsymbol{\Sigma}^{-1}\|_{\max} + \|\mathbf{M}'\widehat{\boldsymbol{\Sigma}}\boldsymbol{\Sigma}^{-1} - \boldsymbol{\Sigma}^{-1}\|_{\max} \\
&\le \|\mathbf{M}'\widehat{\boldsymbol{\Sigma}}\|_{\max}\|\mathbf{M}' - \boldsymbol{\Sigma}^{-1}\|_{1,\max} + \|\mathbf{M}'\widehat{\boldsymbol{\Sigma}} - \mathbf{I}\|_{\max}\|\boldsymbol{\Sigma}^{-1}\|_{1,\max} \\
&\le (1+\gamma_1)4C_0 R s_1\sqrt{\frac{\log p}{n}} + R\gamma_1 \\
&= O\left(s_1\sqrt{\frac{\log p}{n}}\right),
\end{aligned}
$$

with probability at least $1-4p^{-3}$. By the definition of $\Lambda$ in (B.15), we obtain $\Lambda \lesssim s_1\sqrt{(\log p)/n}$, with probability tending to 1. Hence, choosing $\nu = Cs_1\sqrt{(\log p)/n}$ and by the condition $s_1 = o(n^{1/2}/\log^{5/2}(p\vee n))$ in the theorem, we get

$$
\sup_{\alpha}\left|\mathbb{P}(T_{\mathcal{G}} \ge c_G(\alpha)) - \alpha\right| = o(1),
$$

which concludes the proof. $\square$

# C Proof of Results in Section 4

In this section, we provide detailed proofs for results in Section 4, including Theorems 4.3 and 4.5 , and Proposition 4.7.

## C.1 Proof of Theorem 4.3

We need the following preliminary lemmas in order to prove Theorem 4.3. Define the composite quantile loss function

$$
\widehat{\mathcal{L}}_{n,K}(\underset{\sim}{\boldsymbol{\beta}}) := \sum_{k=1}^{K}\frac{1}{n}\sum_{i=1}^{n}\phi_{\tau_k}\left(Y_i - \underset{\sim}{\boldsymbol{X}}_i^T\underset{\sim}{\boldsymbol{\beta}}_k\right),
$$

and its population version

$$
\mathcal{L}_K(\underset{\sim}{\boldsymbol{\beta}}) := \mathbb{E}\left[\widehat{\mathcal{L}}_{n,K}(\underset{\sim}{\boldsymbol{\beta}})\right] = \sum_{k=1}^{K}\mathbb{E}\left[\phi_{\tau_k}\left(Y - \underset{\sim}{\boldsymbol{X}}^T\underset{\sim}{\boldsymbol{\beta}}_k\right)\right].
$$

The first lemma shows that $\mathcal{L}_K(\cdot)$ has a local quadratic curvature in a neighborhood around $\underset{\sim}{\boldsymbol{\beta}}^*$ in terms of the norm $\|\cdot\|_{\mathbf{S}}$.

**Lemma C.1.** Suppose Assumptions 3.1 and 4.1 hold. Define $\mathcal{H}(\underset{\sim}{\boldsymbol{\Delta}}) := \mathcal{L}_K(\underset{\sim}{\boldsymbol{\beta}}^* + \underset{\sim}{\boldsymbol{\Delta}}) - \mathcal{L}_K(\underset{\sim}{\boldsymbol{\beta}}^*)$. Then for any $\underset{\sim}{\boldsymbol{\Delta}} \in \mathcal{A}$, we have

$$
\mathcal{H}(\underset{\sim}{\boldsymbol{\Delta}}) \ge \min\left\{\frac{\|\underset{\sim}{\boldsymbol{\Delta}}\|_{\mathbf{S}}^2}{4}, \frac{\eta}{4}\|\underset{\sim}{\boldsymbol{\Delta}}\|_{\mathbf{S}}\right\}.
$$

where $\eta = 3/(2C_+' m_0)$.

The second lemma shows that under a suitably chosen $\lambda$, $\widehat{\underset{\sim}{\beta}} - \underset{\sim}{\beta}^*$ lies in a restricted set $\mathcal{A}$ with probability tending to 1.

**Lemma C.2.** Suppose Assumption 4.2 holds. Then for the choice of $\lambda \geq 4K \max\{\widetilde{\sigma}_x, 1\}\sqrt{(\log p)/n}$, we have with probability at least $1 - 4Kp^{-7} - \delta_n$ that

$$\|\widehat{\beta}_{\mathcal{T}^c}\|_1 \leq 3\|(\widehat{\beta} - \beta^*)_{\mathcal{T}}\|_1 + \|\widehat{b} - b^*\|_1/K.$$

The third lemma bounds the difference between $\widehat{\mathcal{L}}_{n,K}$ and $\mathcal{L}_K$. We use empirical process theory to obtain a tail probability for the maximum over the set $\mathcal{A}$.

**Lemma C.3.** Suppose Assumptions 3.1, 3.6, 4.2 hold and $p > 3$. Then we have with probability at least $1 - 6p^{-3} - 3\delta_n$ that

$$\sup_{\underset{\sim}{\Delta} \in \mathcal{A}, \|\underset{\sim}{\Delta}\|_{\mathbf{S}} \leq \xi} \left| \widehat{\mathcal{L}}_{n,K}(\underset{\sim}{\beta}^* + \underset{\sim}{\Delta}) - \widehat{\mathcal{L}}_{n,K}(\underset{\sim}{\beta}^*) - \big(\mathcal{L}_K(\underset{\sim}{\beta}^* + \underset{\sim}{\Delta}) - \mathcal{L}_K(\underset{\sim}{\beta}^*)\big) \right| \leq C_E \sqrt{K} \xi \sqrt{\frac{s \log p}{n}},$$

where $C_E = \max\big\{\sqrt{6}, 32\widetilde{\sigma}_x(4\rho_{\min}^{-1/2} + 1)\big\} C_-^{-1/2}$.

We defer the proof of the above lemmas to Appendix E. With these preliminary lemmas, we are ready to prove Theorem 4.3.

*Proof of Theorem 4.3.* Define the following event

$$\mathcal{E}_1 = \left\{ \sup_{\underset{\sim}{\Delta} \in \mathcal{A}, \|\underset{\sim}{\Delta}\|_{\mathbf{S}} \leq \xi} \left| \widehat{\mathcal{L}}_{n,K}(\underset{\sim}{\beta}^* + \underset{\sim}{\Delta}) - \widehat{\mathcal{L}}_{n,K}(\underset{\sim}{\beta}^*) - \big(\mathcal{L}_K(\underset{\sim}{\beta}^* + \underset{\sim}{\Delta}) - \mathcal{L}_K(\underset{\sim}{\beta}^*)\big) \right| \leq C_E \sqrt{K} \xi \sqrt{s \frac{\log p}{n}} \right\},$$

and

$$\mathcal{E}_2 = \{\widehat{\underset{\sim}{\beta}} - \underset{\sim}{\beta}^* \in \mathcal{A}\}.$$

By Lemma C.2 and C.3, when $\lambda \geq 4K \max\{\widetilde{\sigma}_x, 1\}\sqrt{\frac{\log p}{n}}$, we have $\mathbb{P}(\mathcal{E}_1^c \cup \mathcal{E}_2^c) \leq 1 - 4Kp^{-7} - 6p^{-3} - 3\delta_n$.

The following derivation is based on the condition that events $\mathcal{E}_1$ and $\mathcal{E}_2$ hold. Let $\|\widehat{\underset{\sim}{\beta}} - \underset{\sim}{\beta}^*\|_{\mathbf{S}} = \xi$. By optimality condition, we have

$$\widehat{\mathcal{L}}_{n,K}(\widehat{\underset{\sim}{\beta}}) - \widehat{\mathcal{L}}_{n,K}(\underset{\sim}{\beta}^*) + \lambda(\|\widehat{\beta}\|_1 - \|\widehat{\beta}^*\|_1) \leq 0. \tag{C.1}$$

On the events $\mathcal{E}_1$ and $\mathcal{E}_2$, we have

$$\left| \widehat{\mathcal{L}}_{n,K}(\widehat{\underset{\sim}{\beta}}) - \widehat{\mathcal{L}}_{n,K}(\underset{\sim}{\beta}^*) - \big(\mathcal{L}_K(\widehat{\underset{\sim}{\beta}}) - \mathcal{L}_K(\underset{\sim}{\beta}^*)\big) \right| \leq C_E \sqrt{K} \xi \sqrt{(s \log p)/n},$$

so it follows that

$$\begin{aligned} \widehat{\mathcal{L}}_{n,K}(\widehat{\underset{\sim}{\beta}}) - \widehat{\mathcal{L}}_{n,K}(\underset{\sim}{\beta}^*) &\geq \mathcal{L}_K(\widehat{\underset{\sim}{\beta}}) - \mathcal{L}_K(\underset{\sim}{\beta}^*) - C_E \sqrt{K} \xi \sqrt{\frac{s \log p}{n}} \\ &\geq \min\left\{\frac{\xi^2}{4}, \frac{\eta}{4}\xi\right\} - C_E \sqrt{K} \xi \sqrt{\frac{s \log p}{n}}, \end{aligned} \tag{C.2}$$

where the second inequality is by Lemma C.1. Furthermore, as

$$\xi^2 = \|\underset{\sim}{\widehat{\boldsymbol{\beta}}} - \underset{\sim}{\boldsymbol{\beta}}^*\|_{\mathbf{S}}^2 = \sum_{k=1}^{K} f_k^*(\|\widehat{\boldsymbol{\beta}} - \boldsymbol{\beta}^*\|_{\boldsymbol{\Sigma}}^2 + (\widehat{b}_k - b_k^*)^2)$$
$$\geq C_-(K\rho_{\min}\|\widehat{\boldsymbol{\beta}} - \boldsymbol{\beta}^*\|_2^2 + \|\widehat{\boldsymbol{b}} - \boldsymbol{b}^*\|_2^2),$$

we have $\|\widehat{\boldsymbol{\beta}} - \boldsymbol{\beta}^*\|_2 \leq C_-^{-1/2}\rho_{\min}^{-1/2}K^{-1/2}\xi$ and $\|\widehat{\boldsymbol{b}} - \boldsymbol{b}^*\|_2 \leq C_-^{-1/2}\xi$. Therefore, on the event $\mathcal{E}_2$, it holds that

$$\begin{aligned}\|\widehat{\boldsymbol{\beta}}^*\|_1 - \|\widehat{\boldsymbol{\beta}}\|_1 &\leq \|\widehat{\boldsymbol{\beta}} - \boldsymbol{\beta}^*\|_1 \\ &\leq 4\|\widehat{\boldsymbol{\beta}}_{\mathcal{T}} - \boldsymbol{\beta}^*\|_1 + \|\widehat{\boldsymbol{b}} - \boldsymbol{b}^*\|_1/K \\ &\leq 4\sqrt{s}\|\widehat{\boldsymbol{\beta}} - \boldsymbol{\beta}^*\|_2 + \|\widehat{\boldsymbol{b}} - \boldsymbol{b}^*\|_2/\sqrt{K} \\ &\leq C_-^{-1/2}(4\rho_{\min}^{-1/2} + 1)\sqrt{s/K}\xi.\end{aligned} \tag{C.3}$$

Now in view of (C.1), (C.2) and (C.3), we get

$$\min\left\{\frac{\xi^2}{4}, \frac{\eta}{4}\xi\right\} - C_E\sqrt{K}\xi\sqrt{\frac{s\log p}{n}} - C_-^{-1/2}(4\rho_{\min}^{-1/2} + 1)\sqrt{\frac{s}{K}}\xi\lambda \leq 0. \tag{C.4}$$

As $\lambda \leq 4K\max\{\widetilde{\sigma}_x, 1\}\zeta\sqrt{(\log p)/n}$, (C.4) implies that either

$$\frac{\eta}{4}\xi - C_E\sqrt{K}\xi\sqrt{\frac{s\log p}{n}} - 4C_-^{-1/2}\max\{\widetilde{\sigma}_x, 1\}(4\rho_{\min}^{-1/2} + 1)\zeta\sqrt{K}\xi\sqrt{\frac{s\log p}{n}} \leq 0, \tag{C.5}$$

or

$$\frac{\xi^2}{4} - C_E\sqrt{K}\xi\sqrt{\frac{s\log p}{n}} - 4C_-^{-1/2}\max\{\widetilde{\sigma}_x, 1\}(4\rho_{\min}^{-1/2} + 1)\zeta\sqrt{K}\xi\sqrt{\frac{s\log p}{n}} \leq 0. \tag{C.6}$$

By the definition of $\eta = 3/(2C_+'m_0)$, (C.5) cannot hold under the scaling condition $\sqrt{(s\log p)/n} < \frac{3}{8C_+'m_0C_0\sqrt{K}}$, since $C_0 \geq C_E + 4C_-^{-1/2}\max\{\widetilde{\sigma}_x, 1\}\zeta(4\rho_{\min}^{-1/2} + 1)$ by definitions of $C_E$ and $C_0$. Hence (C.6) must hold, which implies that

$$\|\underset{\sim}{\widehat{\boldsymbol{\beta}}} - \underset{\sim}{\boldsymbol{\beta}}^*\|_{\mathbf{S}} = \xi \leq 4C_0\sqrt{K}\sqrt{\frac{s\log p}{n}}.$$

Since

$$\|\underset{\sim}{\widehat{\boldsymbol{\beta}}} - \underset{\sim}{\boldsymbol{\beta}}^*\|_{\mathbf{S}}^2 = \sum_{k=1}^{K} f_\epsilon(b_k^*)\left((\widehat{\boldsymbol{\beta}} - \boldsymbol{\beta}^*)^T\boldsymbol{\Sigma}(\widehat{\boldsymbol{\beta}} - \boldsymbol{\beta}^*) + (\widehat{b}_k - b_k^*)^2\right),$$

we conclude

$$\|\widehat{\boldsymbol{\beta}} - \boldsymbol{\beta}^*\|_2 \leq \frac{4C_0}{\sqrt{C_-\rho_{\min}}}\sqrt{\frac{s\log p}{n}},$$

and

$$\|\widehat{\boldsymbol{b}} - \boldsymbol{b}^*\|_2 \leq \frac{4C_0}{\sqrt{C_-}}\sqrt{K}\sqrt{\frac{s\log p}{n}},$$

as desired. □

## C.2 Proof of Theorem 4.5

In this section, we show that the sparsity of $\widehat{\boldsymbol{\beta}}$ is of the order $s$. We first obtain a crude bound for $\widehat{s}$ in the following lemma.

**Lemma C.4.** For any choice of $\lambda > 0$, the following inequality holds almost surely:

$$\widehat{s} \le K^2 \lambda^{-2} \psi(\widehat{s}).$$

In particular, if we choose $\lambda \ge K\sqrt{2\psi(n/\log(p \vee n))\log(p\vee n)/n}$, we have

$$\widehat{s} \le n/\log(p \vee n).$$

The proof is deferred to Appendix E.

*Proof of Theorem 4.5.* We first reformulate (2.3) into the following linear programming problem,

$$\begin{aligned} \min_{\substack{\boldsymbol{\Pi}^+,\boldsymbol{\Pi}^- \in \mathbb{R}^{n\times K}\\ \boldsymbol{\beta}^+,\boldsymbol{\beta}^-\in\mathbb{R}^p, \boldsymbol{b}\in\mathbb{R}^K}} & \sum_{k=1}^K \frac{1}{n}\sum_{k=1}^K \big(\tau_k \Pi_{ik}^+ + (1-\tau_k)\Pi_{ik}^-\big) + \lambda \sum_{j=1}^p (\beta_j^+ + \beta_j^-) \\ \text{s.t.} \quad & \Pi_{ik}^+ - \Pi_{ik}^- = Y_i - \boldsymbol{X}_i^T(\boldsymbol{\beta}^+ - \boldsymbol{\beta}^-) - b_k, \text{ for } i = 1,\dots,n, k=1,\dots,K. \end{aligned} \tag{C.7}$$

This problem has the dual of the form

$$\begin{aligned} \max_{\boldsymbol{a}\in\mathbb{R}^{n\times K}} & \sum_{i=1}^n \sum_{k=1}^K a_{ik} Y_i \\ \text{s.t.} \quad & \Big|\frac{1}{n}\sum_{i=1}^n\sum_{k=1}^K a_{ik}X_{ij}\Big| \le \lambda, \text{ for } j=1,\dots,p \\ & \tau_k - 1 \le a_{ik} \le \tau_k, \text{ for } i=1,\dots,n, k=1,\dots,K. \end{aligned} \tag{C.8}$$

Denote $\widetilde{\boldsymbol{a}} \in \mathbb{R}^{n\times K}$ to be the optimal solution of (C.8). By complementary slackness, we have $\widehat{\beta}_j > 0$ if and only if $n^{-1}\sum_{i=1}^n\sum_{k=1}^K \widetilde{a}_{ik}X_{ij} = \lambda$ and $\widehat{\beta}_j < 0$ if and only if $n^{-1}\sum_{i=1}^n\sum_{k=1}^K \widetilde{a}_{ik}X_{ij} = -\lambda$. Therefore, this implies that

$$\Big[\sum_{k=1}^K \mathbb{X}^T \widetilde{\boldsymbol{a}}_k\Big]_j = \operatorname{sign}(\widehat{\beta}_j) n\lambda, \tag{C.9}$$

for $j \in \widehat{\mathcal{T}}$, where $\widetilde{\boldsymbol{a}}_k = (\widetilde{a}_{1k},\dots,\widetilde{a}_{nk})^T \in \mathbb{R}^n$, and $\widehat{\mathcal{T}}$ is the support of $\widehat{\boldsymbol{\beta}}$. Then we have

$$\sqrt{\widehat{s}} = \Big\|\widehat{\boldsymbol{\beta}}_{\widehat{\mathcal{T}}}\Big\|_2 = \Big\|\frac{\sum_{k=1}^K \mathbb{X}_{\widehat{\mathcal{T}}}^T \widetilde{\boldsymbol{a}}_k}{n\lambda}\Big\|_2. \tag{C.10}$$

Define $\widehat{a}_{ik} := \tau_k - \mathbb{1}\{Y_i \le \boldsymbol{X}_i^T\widehat{\boldsymbol{\beta}} + \widehat{b}_k\}$ and $a_{ik}^* := \tau_k - \mathbb{1}\{\epsilon_i \le b_k^*\}$. By (C.10), it holds that

$$\begin{aligned} \lambda\sqrt{\widehat{s}} &= \Big\|\frac{1}{n}\sum_{i=1}^n \boldsymbol{X}_{i,\widehat{\mathcal{T}}}\sum_{k=1}^K \widetilde{a}_{ik}\Big\|_2 \\ &\le \Big\|\frac{1}{n}\sum_{i=1}^n \boldsymbol{X}_{i,\widehat{\mathcal{T}}}\sum_{k=1}^K (\widetilde{a}_{ik}-\widehat{a}_{ik})\Big\|_2 + \Big\|\frac{1}{n}\sum_{i=1}^n \boldsymbol{X}_{i,\widehat{\mathcal{T}}}\sum_{k=1}^K (\widehat{a}_{ik}-a_{ik}^*)\Big\|_2 + \Big\|\frac{1}{n}\sum_{i=1}^n \boldsymbol{X}_{i,\widehat{\mathcal{T}}}\sum_{k=1}^K a_{ik}^*\Big\|_2. \end{aligned} \tag{C.11}$$

We control the three terms in (C.11) separately. For the first term, by complementary slackness, observe that $Y_i > \boldsymbol{X}_i^T\widehat{\boldsymbol{\beta}} + \widehat{b}_k$ implies $\Pi_{ik}^+ > 0$ and $\Pi_{ik}^- = 0$, which further implies that $\widetilde{a}_{ik} = \tau_k = \widehat{a}_{ik}$. Similarly, $Y_i < \boldsymbol{X}_i^T\widehat{\boldsymbol{\beta}} + \widehat{b}_k$ implies $\widetilde{a}_{ik} = \tau_k - 1 = \widehat{a}_{ik}$. Hence, we can conclude that

$$\widetilde{a}_{ik} \neq \widehat{a}_{ik} \quad \text{only if} \quad Y_i = \boldsymbol{X}_i^T\widehat{\boldsymbol{\beta}} + \widehat{b}_k. \tag{C.12}$$

Moreover, we claim that $\big|\{(i,k) : Y_i = \boldsymbol{X}_i^T\widehat{\boldsymbol{\beta}} + \widehat{b}_k\}\big| = \|\widehat{\boldsymbol{\beta}}\|_0 + \|\boldsymbol{b}\|_0 = \widehat{s} + K$. Indeed, the optimal solution of (C.7) is a basic solution, which contains $nk$ nonzero variables. Hence the number of nonzeros of $\Pi_+ + \Pi_-$ and the number of nonzeros of $\beta^+ + \beta^-$ and $b^+ + b^-$ sum up to $nk$, while the number of nonzeros of $\Pi_+ + \Pi_-$ and the number of zeros of $\Pi_+ + \Pi_-$ also sum up to $nk$. Therefore the number of nonzeros of $\beta^+ + \beta^-$ and $b^+ + b^-$ is equal to the number of zeros of $\Pi_+ + \Pi_-$, which further equal to $\big|\{(i,k) : Y_i = \boldsymbol{X}_i^T\widehat{\boldsymbol{\beta}} + \widehat{b}_k\}\big|$. Hence, by relationship (C.12), we have $\big|\{(i,k) : \widehat{a}_{ik} \neq \widetilde{a}_{ik}\}\big| \le \widehat{s} + K$. Therefore, it follows that

$$\begin{aligned}
\Big\|\frac{1}{n}\sum_{i=1}^n \boldsymbol{X}_{i,\widehat{\mathcal{T}}}\sum_{k=1}^K(\widetilde{a}_{ik} - \widehat{a}_{ik})\Big\|_2 &\le \sup_{\boldsymbol{\theta}_{\widehat{\mathcal{T}}^c}=\mathbf{0}, \|\boldsymbol{\theta}\|\le 1}\Big|\boldsymbol{\theta}^T\frac{1}{n}\sum_{i=1}^n \boldsymbol{X}_{i,\widehat{\mathcal{T}}}\sum_{k=1}^K(\widetilde{a}_{ik} - \widehat{a}_{ik})\Big| \\
&\le \sup_{\boldsymbol{\theta}_{\widehat{\mathcal{T}}^c}=\mathbf{0}, \|\boldsymbol{\theta}\|\le 1}\left(\frac{1}{n}\sum_{i=1}^n(\boldsymbol{\theta}^T\boldsymbol{X}_i)^2\right)^{1/2}\left(\frac{1}{n}\sum_{i=1}^n\left(\sum_{k=1}^K\widetilde{a}_{ik} - \widehat{a}_{ik}\right)^2\right)^{1/2} \\
&\le n^{-1/2}\sqrt{\psi(\widehat{\mathcal{T}})}\left(\sum_{i=1}^n\left(\sum_{k=1}^K\widetilde{a}_{ik} - \widehat{a}_{ik}\right)^2\right)^{1/2} \\
&\le \sqrt{K/n}\sqrt{\psi(\widehat{\mathcal{T}})}\left(\sum_{i=1}^n\sum_{k=1}^K(\widetilde{a}_{ik} - \widehat{a}_{ik})^2\right)^{1/2} \\
&\le \sqrt{K/n}\sqrt{\psi(\widehat{\mathcal{T}})}\sqrt{\widehat{s} + K}.
\end{aligned} \tag{C.13}$$

We next control the last term in (C.11). By (E.10) where we used Hoeffding's inequality, it holds that

$$\mathbb{P}\left(\Big\|\sum_{k=1}^K\frac{1}{n}\sum_{i=1}^n a_{ik}^*\boldsymbol{X}_i\Big\|_\infty \ge K\widetilde{\sigma}_x\sqrt{\frac{\log p}{n}}\right) \le 2Kp^{-7} + \delta_n.$$

Therefore, it follows that with probability as least $1 - 2Kp^{-7} - \delta_n$, we have

$$\Big\|\frac{1}{n}\sum_{i=1}^n \boldsymbol{X}_{i,\widehat{\mathcal{T}}}\sum_{k=1}^K a_{ik}^*\Big\|_2 \le \sqrt{\widehat{s}}\Big\|\sum_{k=1}^K\frac{1}{n}\sum_{i=1}^n a_{ik}^*\boldsymbol{X}_i\Big\|_\infty \le \sqrt{\widehat{s}}K\widetilde{\sigma}_x\sqrt{\frac{\log p}{n}}. \tag{C.14}$$

Lastly, we control the second term in (C.11). We first define the following notations. For any $\underset{\sim}{\boldsymbol{\Delta}}_k = (\boldsymbol{\Delta}^T, \delta_k) \in \mathbb{R}^{p+1}$ and $\boldsymbol{\theta} \in \mathbb{R}^p$, define

$$h(\epsilon_i, \boldsymbol{X}_i; \boldsymbol{\theta}, \underset{\sim}{\boldsymbol{\Delta}}_k) = \boldsymbol{\theta}^T\boldsymbol{X}_i\big(\mathbb{1}\{\epsilon_i \le \underset{\sim}{\boldsymbol{X}}_i^T\underset{\sim}{\boldsymbol{\Delta}}_k + b_k^*\} - \mathbb{1}\{\epsilon_i \le b_k^*\}\big)$$

and

$$\mathbb{G}(\boldsymbol{\theta}, \underset{\sim}{\boldsymbol{\Delta}}_k) := \frac{1}{\sqrt{n}}\sum_{i=1}^n\big\{h(\epsilon_i, \boldsymbol{X}_i; \boldsymbol{\theta}, \underset{\sim}{\boldsymbol{\Delta}}_k) - \mathbb{E}[h(\epsilon_i, \boldsymbol{X}_i; \boldsymbol{\theta}, \underset{\sim}{\boldsymbol{\Delta}}_k)]\big\}.$$

Moreover, define

$$\mathcal{E}_1(q,\xi) := \sup_{\substack{\boldsymbol{\theta}\in\mathbb{S}(q)\\ \underset{\sim}{\boldsymbol{\Delta}}\in R(q,\xi)}} \sum_{k=1}^{K} n^{-1/2}|\mathbb{G}(\boldsymbol{\theta}, \underset{\sim}{\boldsymbol{\Delta}}_k)|, \quad \text{and} \quad \mathcal{E}_2(q,\xi) := \sup_{\substack{\boldsymbol{\theta}\in\mathbb{S}(q)\\ \underset{\sim}{\boldsymbol{\Delta}}\in R(q,\xi)}} \sum_{k=1}^{K} \big|\mathbb{E}\big[h(\epsilon_i, \boldsymbol{X}_i; \boldsymbol{\theta}, \underset{\sim}{\boldsymbol{\Delta}}_k)\big]\big|,$$

where

$$R(q,\xi) := \big\{\underset{\sim}{\boldsymbol{\Delta}} = (\boldsymbol{\Delta}, \boldsymbol{\delta}) \in \mathbb{R}^{p+K} : \|\boldsymbol{\Delta}\|_0 \le q, \|\underset{\sim}{\boldsymbol{\Delta}}\|_{\mathbf{S}} \le \xi\big\},$$

and

$$\mathbb{S}(q) = \{\boldsymbol{\theta} \in \mathbb{R}^p : \|\boldsymbol{\theta}\|_0 \le q, \|\boldsymbol{\theta}\|_2 \le 1\}.$$

With these definitions and by Theorem 4.3, we have with probability at least $1-4Kp^{-7}-6p^{-3}-3\delta_n$ that

$$\begin{aligned}
&\Big\|\frac{1}{n}\sum_{i=1}^{n} \boldsymbol{X}_{i,\widehat{\mathcal{T}}} \sum_{k=1}^{K}(\widehat{a}_{ik} - a^*_{ik})\Big\|_2 \\
&\le \Big\|\sum_{k=1}^{K}\Big(\frac{1}{n}\sum_{i=1}^{n} \boldsymbol{X}_{i,\widehat{\mathcal{T}}}(\widehat{a}_{ik} - a^*_{ik}) - \mathbb{E}\big[\boldsymbol{X}_{i,\widehat{\mathcal{T}}}(\widehat{a}_{ik} - a^*_{ik})\big]\Big)\Big\|_2 + \Big\|\sum_{k=1}^{K}\mathbb{E}\big[\boldsymbol{X}_{i,\widehat{\mathcal{T}}}(\widehat{a}_{ik} - a^*_{ik})\big]\Big\|_2 \qquad \text{(C.15)} \\
&\le \sup_{\boldsymbol{\theta}\in\mathbb{S}(q)} \sum_{k=1}^{K}\Big|n^{-1/2}\mathbb{G}(\boldsymbol{\theta}, \underset{\sim}{\widehat{\boldsymbol{\Delta}}}_k)\Big| + \sup_{\boldsymbol{\theta}\in\mathbb{S}(q)} \sum_{k=1}^{K}\big|\mathbb{E}\big[h(\epsilon_i, \boldsymbol{X}_i; \boldsymbol{\theta}, \underset{\sim}{\boldsymbol{\Delta}}_k)\big]\big| \\
&\le \mathcal{E}_1(\widehat{s}, \xi_n) + \mathcal{E}_2(\widehat{s}, \xi_n).
\end{aligned}$$

where $\xi_n = 4C_0\sqrt{K}\sqrt{(s\log p)/n}$. By Lemma C.5, we have with probability at least $1-(p\vee n)^{-3}$ that

$$\mathcal{E}_1(\widehat{s}, \xi_n) \le c_0 K\sqrt{\widehat{s}\log(p\vee n)/n}\sqrt{\psi(\widehat{s})},$$

for some universal constant $c_0$. By Lemma C.6, we have $\mathcal{E}_2(\widehat{s},\xi_n) \le \sqrt{3K\rho_{\max}C_+/C_-}\xi_n$. Therefore, combining (C.11), (C.13), (C.14) and (C.15), we get with probability at least $1-6Kp^{-7}-8p^{-3}-4\delta_n$ that

$$\lambda\sqrt{\widehat{s}} \le \sqrt{K/n}\sqrt{\psi(\widehat{s})}\sqrt{\widehat{s}+K} + K\widetilde{\sigma}_x\sqrt{\frac{\widehat{s}\log p}{n}} + c_0K\sqrt{\psi(\widehat{s})\frac{\widehat{s}\log(p\vee n)}{n}} + \sqrt{3K\rho_{\max}C_+/C_-}\xi_n.$$

By Lemma C.4 and Assumption 4.4 that $\psi(n/\log(n\vee p)) \le \psi_0$ with probability at least $1-d_n$, we have

$$\begin{aligned}
\lambda\sqrt{\widehat{s}}/K &\le \psi_0^{1/2}\sqrt{\frac{\widehat{s}}{n}} + \widetilde{\sigma}_x\sqrt{\frac{\widehat{s}\log p}{n}} + c_0\psi_0^{1/2}\sqrt{\frac{\widehat{s}\log(p\vee n)}{n}} + \sqrt{3\rho_{\max}C_+/(KC_-)}\xi_n \\
&\le C_S\sqrt{\frac{\widehat{s}\log(p\vee n)}{n}} + \sqrt{3\rho_{\max}C_+/(KC_-)}\xi_n, \qquad \text{(C.16)}
\end{aligned}$$

with probability at least $1-6Kp^{-7}-8p^{-3}-4\delta_n-d_n$, where $C_S := \widetilde{\sigma}_x + \psi_0^{1/2} + c_0\psi_0^{1/2}$. When $\lambda \ge 2KC_S\sqrt{(\log p)/n}$, it holds

$$C_S\sqrt{\frac{\widehat{s}\log(p\vee n)}{n}} \le \sqrt{3\rho_{\max}C_+/(KC_-)}\xi_n = 4\sqrt{3\rho_{\max}C_+/C_-}C_0\sqrt{\frac{s\log p}{n}},$$

therefore,

$$\widehat{s} \leq \frac{48\rho_{\max} C_+ C_0^2}{C_- C_S^2} s,$$

which completes the proof. $\square$

Next, we state Lemma C.5 and Lemma C.6 that were used in the proof.

**Lemma C.5.** Suppose conditions in Theorem 4.5 hold. We have with probability at least $1-(p \vee n)^{-3}$,

$$\mathcal{E}_1(q,\xi) \leq c_0 K \sqrt{q \log(p \vee n)/n} \sqrt{\psi(q)}.$$

for any $q \leq p$ and $\xi > 0$, where $c_0$ is some universal constant.

The proof of this lemma follows similarly as that of Lemma 6.1.

**Lemma C.6.** Suppose conditions in Theorem 4.5 hold. We have

$$\mathcal{E}_2(q,\xi) \leq \sqrt{3K\rho_{\max} C_+ / C_-}\,\xi$$

for any $q \leq p$ and $\xi > 0$.

The proof of the above lemma is deferred to Appendix E.

### C.3 Proof of Proposition 4.7

We decompose the proof of Proposition 4.7 so that Part (i), (ii) and (iii) of the proposition are proved by Lemma C.7, C.8 and C.9 respectively.

**Lemma C.7.** Suppose Assumptions 3.1, 3.5, 3.6 hold. Then for any $\underset{\sim}{\boldsymbol{\beta}} = (\boldsymbol{\beta}^T, b_1, \ldots, b_K)^T \in \mathbb{R}^{p+K}$,

$$\sum_{k=1}^{K} \mathbb{E}\big[|\underset{\sim}{\boldsymbol{X}}^T \underset{\sim}{\boldsymbol{\beta}}_k|^3\big] \leq m_0 \|\underset{\sim}{\boldsymbol{\beta}}\|_{\mathbf{S}}^3,$$

where $\underset{\sim}{\boldsymbol{\beta}}_k = (\boldsymbol{\beta}^T, b_k)^T$ and $m_0 = C_-^{3/2} \min\{\rho_{\min}^{3/2}, 1\}/(3\max\{8^{5/4}\sigma_x^{3/2}, 1\})$.

*Proof.* By definition, we have

$$\begin{aligned}
\|\underset{\sim}{\boldsymbol{\beta}}\|_{\mathbf{S}}^2 &= \sum_{k=1}^{K} f_k^*(\|\boldsymbol{\beta}\|_{\boldsymbol{\Sigma}}^2 + b_k^2) \\
&\geq C_-\Big(K\rho_{\min}\|\boldsymbol{\beta}\|_2^2 + \sum_{k=1}^{K} b_k^2\Big) \\
&\geq C_- \min\{\rho_{\min}, 1\}(K\|\boldsymbol{\beta}\|_2^2 + \|\boldsymbol{b}\|^2),
\end{aligned}$$

where $\boldsymbol{b} = (b_1, \ldots, b_K)^T \in \mathbb{R}^K$. On the other hand, by Assumption 3.5, $\underset{\sim}{\boldsymbol{X}}^T \boldsymbol{\beta}_k$ is a sub-Gaussian random variable with mean $b_k$ and variance proxy $\|\boldsymbol{\beta}\|_2^2 \sigma_x^2$. Therefore, by the moment bound for sub-Gaussian random variables (e.g., see Page 47 of van Handel (2014)),

$$\begin{aligned}
\mathbb{E}\big[|\underset{\sim}{\boldsymbol{X}}^T \underset{\sim}{\boldsymbol{\beta}}_k|^3\big] &\leq 3\mathbb{E}\big[|\underset{\sim}{\boldsymbol{X}}^T \underset{\sim}{\boldsymbol{\beta}}_k - b_k|^3 + |b_k|^3\big] \\
&\leq 3(8^{5/4}\sigma_x^{3/2}\|\boldsymbol{\beta}\|_2^3 + |b_k|^3) \\
&\leq 3\max\{8^{5/4}\sigma_x^{3/2}, 1\}(\|\boldsymbol{\beta}\|_2^3 + |b_k|^3).
\end{aligned}$$

And so

$$\sum_{k=1}^{K} \mathbb{E}\big[|\underset{\sim}{\boldsymbol{X}}^T \underset{\sim}{\boldsymbol{\beta}}_k|^3\big] \leq 3\max\{8^{5/4}\sigma_x^{3/2}, 1\}\Big(K\|\boldsymbol{\beta}\|_2^3 + \sum_{k=1}^{K} |b_k|^3\Big). \tag{C.17}$$

As $K\|\boldsymbol{\beta}\|_2^3 + \sum_{k=1}^{K} |b_k|^3 \leq (K\|\boldsymbol{\beta}\|_2^2 + \|\boldsymbol{b}\|^2)^{3/2}$, by (C.17) and (C.17), we have

$$\sum_{k=1}^{K} \mathbb{E}\big[|\underset{\sim}{\boldsymbol{X}}^T \underset{\sim}{\boldsymbol{\beta}}_k|^3\big] \leq C_-^{3/2} \min\{\rho_{\min}^{3/2}, 1\}/(3\max\{8^{5/4}\sigma_x^{3/2}, 1\})\|\underset{\sim}{\boldsymbol{\Delta}}\|_{\mathbf{S}}^3,$$

which finishes the proof. □

**Lemma C.8.** Suppose Assumption 3.5 holds. When $\log p \leq n/2$, we have

$$\mathbb{P}\left(\max_{j=[p]} \left\{\frac{1}{n}\sum_{i=1}^{n} X_{ij}^2\right\} \geq c\sigma_x^2\right) \leq 2\exp(-n/2).$$

*Proof.* By definition, $X_{ij}$ is $\sigma_x$-sub-Gaussian. By Lemma 5.14 in Vershynin (2012), we have $X_{ij}^2$ is $c\sigma_x^2$ sub-exponential random variables, for some universal constant $c$, that is,

$$\mathbb{P}(X_{ij}^2 - \mathbb{E}[X_{ij}^2] > t) \leq \exp(1 - t/(c\sigma_x^2)).$$

Using Bernstein's inequality for sub-exponential random variables (Proposition 5.16 of Vershynin (2012)), we get

$$\mathbb{P}\left(\frac{1}{n}\sum_{i=1}^{n} X_{ij}^2 - \mathbb{E}[X_{ij}^2] \geq t\right) \leq 2\exp\left(-c_1 \min\left\{\frac{nt^2}{c^2\sigma_x^4}, \frac{nt}{c\sigma_x^2}\right\}\right),$$

where $c_1$ is a universal constant. Applying union bound, we have

$$\mathbb{P}\left(\max_{1\leq j\leq p} \left\{\frac{1}{n}\sum_{i=1}^{n} X_{ij}^2 - \mathbb{E}[X_{ij}^2]\right\} \geq t\right) \leq 2p\exp\left(-c_1 \min\left\{\frac{nt^2}{c^2\sigma_x^4}, \frac{nt}{c\sigma_x^2}\right\}\right).$$

Let $t = c\sigma_x^2$, and by the moment bound for sub-Gaussian random variables $\mathbb{E}[X_{ij}^2] \leq 4\sigma_x^2$ (See Page 47 of van Handel (2014)), we have

$$\mathbb{P}\left(\max_{1\leq j\leq p} \left\{\frac{1}{n}\sum_{i=1}^{n} X_{ij}^2\right\} \geq c\sigma_x^2\right) \leq 2p\exp(-n),$$

for some universal constant $c$. When $\log p \leq n/2$, we have

$$\mathbb{P}\left(\max_{1\leq j\leq p} \left\{\frac{1}{n}\sum_{i=1}^{n} X_{ij}^2\right\} \geq c\sigma_x^2\right) \leq 2\exp(-n/2),$$

as desired. □

**Lemma C.9** (Theorem 5.65 in Vershynin (2012))**.** Define

$$\psi(q, \mathbf{A}) = \sup_{\|\boldsymbol{x}\|_0 \le q} \frac{\|\mathbf{A}\boldsymbol{x}\|_2^2}{\|\boldsymbol{x}\|_2^2}.$$

Let $\mathbf{A}$ be an $n \times p$ matrix with independent $\sigma^2$-sub-Gaussian rows. Then for every real number $\delta \in (0,1)$ and integer $q$ satisfying $1 \le q \le p$ and $n \ge C\delta^{-2} q \log(ep/q)$, we have with probability at least $1 - 2\exp(-c\delta^2 n)$ that

$$\psi(q, \bar{\mathbf{A}}) \le \delta + 1,$$

where $\bar{\mathbf{A}}$ is the normalized matrix $\bar{\mathbf{A}} = \frac{1}{\sqrt{n}}\mathbf{A}$ and $c, C$ depend only on the sub-Gaussian variance proxy $\sigma^2$.

# D Proof of Results in Section 5

In this section we provide the detailed proofs of results in Section 5. We first prove the main result for asymptotic normality of the divide-and-conquer estimator $\bar{\boldsymbol{\beta}}^d$ in Theorem 5.2. Then we show the oracle property of the divide-and-conquer hypothesis test in Theorem 5.4.

## D.1 Proof of Theorem 5.2

*Proof.* From the proof of Theorem 3.2 in Section 6, on data $\mathcal{D}_\ell$ for each $\ell$, we have

$$\begin{aligned}\sqrt{n}(\widehat{\boldsymbol{\beta}}^d(\ell) - \boldsymbol{\beta}^*) &= \sqrt{n}\theta_K^{-1}\mathbf{M}^{(\ell)}\boldsymbol{U}^{(\ell)} - \sqrt{n}\theta_K^{-1}\mathbf{M}^{(\ell)}\boldsymbol{V}^{(\ell)} - \sqrt{n}\theta_K^{-1}\mathbf{M}^{(\ell)}\boldsymbol{E}^{(\ell)} - \sqrt{n}\theta_K^{-1}\mathbf{M}^{(\ell)}\boldsymbol{D}^{(\ell)} \\ &\quad + \sqrt{n}((\theta_K^{(\ell)})^{-1} - \theta_K^{-1})\mathbf{M}^{(\ell)}\widehat{\boldsymbol{\kappa}}^{(\ell)} - \sqrt{n}(\mathbf{M}^{(\ell)}\widehat{\boldsymbol{\Sigma}}^{(\ell)} - \mathbf{I})\widehat{\boldsymbol{\delta}}^{(\ell)}, \end{aligned} \tag{D.1}$$

where the superscript $(\ell)$ denotes the quantity computed on data split $\mathcal{D}_\ell$. By definition, we have

$$\begin{aligned}\sqrt{N}(\bar{\boldsymbol{\beta}}^d - \boldsymbol{\beta}^*) = \frac{1}{\sqrt{m}}\sum_{\ell=1}^m \Big\{&\sqrt{n}\theta_K^{-1}\mathbf{M}^{(\ell)}\boldsymbol{U}^{(\ell)} - \sqrt{n}\theta_K^{-1}\mathbf{M}^{(\ell)}\boldsymbol{V}^{(\ell)} - \sqrt{n}\theta_K^{-1}\mathbf{M}^{(\ell)}\boldsymbol{E}^{(\ell)} \\ &- \sqrt{n}\theta_K^{-1}\mathbf{M}^{(\ell)}\boldsymbol{D}^{(\ell)} + \sqrt{n}((\theta_K^{(\ell)})^{-1} - \theta_K^{-1})\mathbf{M}^{(\ell)}\widehat{\boldsymbol{\kappa}}^{(\ell)} - \sqrt{n}(\mathbf{M}^{(\ell)}\widehat{\boldsymbol{\Sigma}}^{(\ell)} - \mathbf{I})\widehat{\boldsymbol{\delta}}^{(\ell)}\Big\}. \end{aligned} \tag{D.2}$$

We show that the first term on the R.H.S weakly converges to a normal distribution, and the rest of the terms are asymptotically ignorable under the growth condition of $m$ specified in the theorem. We apply Lindeberg CLT to prove the former, and utilize results derived under single machine combined with union bound to prove the latter.

**The asymptotically normal term** $\frac{1}{\sqrt{m}}\sum_{\ell=1}^m \sqrt{n}\theta_K^{-1}\mathbf{M}^{(\ell)}\boldsymbol{U}^{(\ell)}$. In the following, we show that

$$\frac{1}{\sqrt{m}}\sum_{\ell=1}^m \sqrt{n}\theta_K^{-1}[\mathbf{M}^{(\ell)}\boldsymbol{U}^{(\ell)}]_j \left(m^{-1}\sum_{\ell=1}^m (\widehat{\boldsymbol{\mu}}_j^{(\ell)})^T \widehat{\boldsymbol{\Sigma}}^{(\ell)}\widehat{\boldsymbol{\mu}}_j^{(\ell)}\right)^{-1/2} \rightsquigarrow N(0, \theta_K^{-2}\sigma_K^2). \tag{D.3}$$

By the definition of $\boldsymbol{U}^{(\ell)}$, we have

$$\frac{1}{\sqrt{m}}\sum_{\ell=1}^m \sqrt{n}\theta_K^{-1}\mathbf{M}^{(\ell)}\boldsymbol{U}^{(\ell)} = \frac{1}{\sqrt{N}}\sum_{\ell=1}^m\sum_{i\in\mathcal{D}_\ell} \theta_K^{-1}\mathbf{M}^{(\ell)}\boldsymbol{X}_i\Psi_{i,K}.$$

The R.H.S. is a sum of $N$ independence terms, as samples among each data split are independent. This allows us to apply Lindeberg CLT. Let

$$\xi_{ij}^{(\ell)} = \theta_K^{-1}(\widehat{\boldsymbol{\mu}}_j^{(\ell)})^T \boldsymbol{X}_i \Psi_{i,K} \left( m^{-1} \sum_{\ell=1}^m (\widehat{\boldsymbol{\mu}}_j^{(\ell)})^T \widehat{\boldsymbol{\Sigma}}^{(\ell)} \widehat{\boldsymbol{\mu}}_j^{(\ell)} \right)^{-1/2}.$$

Then we have $\mathbb{E}[\xi_{ij}^{(\ell)}] = \mathbf{0}$ by similar computation as in the proof of Theorem 3.3. By independence of data between sub-samples, we have

$$\begin{aligned}
s_N^2 &:= \operatorname{Var}\left[ \sum_{\ell=1}^m \sum_{i \in \mathcal{D}_\ell} \xi_{ij}^{(\ell)} \,|\, \mathbb{X}^{(1)}, \ldots, \mathbb{X}^{(m)} \right] \\
&= \sum_{\ell=1}^m \sum_{i \in \mathcal{D}_\ell} \operatorname{Var}[\xi_{ij}^{(\ell)} \,|\, \mathbb{X}^{(1)}, \ldots, \mathbb{X}^{(m)}] \\
&= \theta_K^{-2} \sigma_K^2 \sum_{\ell=1}^m \sum_{i \in \mathcal{D}_\ell} (\widehat{\boldsymbol{\mu}}_j^{(\ell)})^T \boldsymbol{X}_i \boldsymbol{X}_i^T \widehat{\boldsymbol{\mu}}_j^{(\ell)} \big( m^{-1} \sum_{\ell=1}^m (\widehat{\boldsymbol{\mu}}_j^{(\ell)})^T \widehat{\boldsymbol{\Sigma}}^{(\ell)} \widehat{\boldsymbol{\mu}}_j^{(\ell)} \big)^{-1} \\
&= N \theta_K^{-2} \sigma_K^2,
\end{aligned} \tag{D.4}$$

where $\mathbb{X}^{(\ell)}$ denotes the design on $\mathcal{D}(\ell)$. Similar to the argument in the proof of Theorem 3.3, we have $|\xi_{ij}^{(\ell)}| \le c_n^{-1} \theta_K^{-1} \gamma_2 |\Psi_{i,K}|$ for all $i \in \mathcal{D}_\ell$, where $c_n \to c_\infty > 0$. Therefore, for any $\varepsilon > 0$,

$$\begin{aligned}
&\lim_{N \to \infty} \frac{1}{s_N^2} \sum_{\ell=1}^m \sum_{i \in \mathcal{D}_\ell} \mathbb{E}\left[ (\xi_{ij}^{(\ell)})^2 \mathbb{1}\{|\xi_{ij}^{(\ell)}| > \varepsilon s_N\} \,|\, \mathbb{X} \right] \\
&= \theta_K^2 \sigma_K^{-2} \lim_{N \to \infty} \frac{1}{N} \sum_{\ell=1}^m \sum_{i \in \mathcal{D}_\ell} \mathbb{E}\Big[ (\xi_{ij}^{(\ell)})^2 \mathbb{1}\left\{ \big|\Psi_{i,K}\big| > \varepsilon c_n \sigma_K \gamma_2^{-1} \sqrt{N} \right\} \,|\, \mathbb{X} \Big] \\
&= \sigma_K^{-2} \lim_{N \to \infty} \mathbb{E}\Big[ (\Psi_{1,K})^2 \mathbb{1}\left\{ \big|\Psi_{1,K}\big| > \varepsilon c_n \sigma_K \gamma_2^{-1} \sqrt{N} \right\} \,|\, \mathbb{X} \Big] \\
&= \sigma_K^{-4} \varepsilon^{-2} c_\infty^{-2} \gamma_2^2 N^{-1} \lim_{N \to \infty} \mathbb{E}\big[ (\Psi_{1,K})^4 \big] = 0,
\end{aligned}$$

where the second equality follows from the same computation as that of (D.4) and the fact that $\Psi_{i,K}$ are i.i.d. for $i \in \mathcal{D}_\ell$, $\ell = 1, \ldots, m$. The last equality is by the fact that $\Psi_{1,K}$ is uniformly bounded and that $\gamma_2 N^{-1/2} = o(1)$, which is implied by condition (5.3). Lastly, we have $s_N^2 / N = \theta_K^{-2} \sigma_K^2$, and we proved the conclusion of (D.3).

**The term $m^{-1/2} \sum_{\ell=1}^m \sqrt{n} \theta_K^{-1} \mathbf{M}^{(\ell)} \boldsymbol{E}^{(\ell)}$.** From the proof of Theorem 3.2, we have for each $\ell = 1, \ldots, m$,

$$\| \sqrt{n} \theta_K^{-1} \mathbf{M}^{(\ell)} \boldsymbol{E}^{(\ell)} \|_\infty \le \theta_K^{-1} C_+ \gamma_3 \sqrt{K} \| \widehat{\boldsymbol{b}}^{(\ell)} - \boldsymbol{b}^* \|_2.$$

By Corollary 4.9, we have

$$\mathbb{P}\left( \| \sqrt{n} \theta_K^{-1} \mathbf{M}^{(\ell)} \boldsymbol{E}^{(\ell)} \|_\infty \ge C_1 \gamma_3 \sqrt{\frac{s \log p}{n}} \right) \le 4Kp^{-7} + 6p^{-3} + 3\exp(-cn),$$

for some constant $C_1$. Applying union bound, we have

$$\mathbb{P}\left(\left\|m^{-1/2}\sum_{\ell=1}^{m}\sqrt{n}\theta_K^{-1}\mathbf{M}^{(\ell)}\boldsymbol{E}^{(\ell)}\right\|_\infty \geq C_1\sqrt{m}\gamma_3\sqrt{\frac{s\log p}{n}}\right) \leq m(4Kp^{-7}+6p^{-3}+3\exp(-cn)).$$

From (5.3) and the choice of $\gamma_3$, $\sqrt{m}\gamma_3\sqrt{s(\log p)/n} = o(1)$, $mp^{-3} = o(1)$ and $m\exp(-cn) = o(1)$. Hence we conclude that

$$\left\|m^{-1/2}\sum_{\ell=1}^{m}\sqrt{n}\theta_K^{-1}\mathbf{M}^{(\ell)}\boldsymbol{E}^{(\ell)}\right\|_\infty = o_P(1).$$

**The term $m^{-1/2}\sum_{\ell=1}^{m}\sqrt{n}\theta_K^{-1}\mathbf{M}^{(\ell)}\boldsymbol{D}^{(\ell)}$.** By the proof of Theorem 3.2, for any $\ell = 1,\dots,m$ we have

$$\|\sqrt{n}\theta_K^{-1}\mathbf{M}^{(\ell)}\boldsymbol{D}^{(\ell)}\|_\infty \leq C'_+\gamma_2\sqrt{n}(2K\mathrm{MSE}(\widehat{\boldsymbol{\beta}}^{(\ell)})^2 + 2\|\widehat{\boldsymbol{b}}^{(\ell)} - \boldsymbol{b}^*\|_2^2).$$

By Corollary 4.9, and the fact that $\theta_K^{-1} \leq K^{-1}C_-^{-1}$, we have

$$\mathbb{P}\left(\|\sqrt{n}\theta_K^{-1}\mathbf{M}^{(\ell)}\boldsymbol{D}^{(\ell)}\|_\infty \geq C'_2\gamma_2\frac{s\log p}{\sqrt{n}}\right) \leq 4Kp^{-7}+6p^{-3}+3\exp(-cn),$$

for some constant $C'_2$. Applying union bound, we have

$$\mathbb{P}\left(\left\|m^{-1/2}\sum_{\ell=1}^{m}\sqrt{n}\theta_K^{-1}\mathbf{M}^{(\ell)}\boldsymbol{D}^{(\ell)}\right\|_\infty \geq C'_2\sqrt{m}\gamma_2\frac{s\log p}{\sqrt{n}}\right) \leq m(4Kp^{-7}+6p^{-3}+3\exp(-cn)).$$

By condition (5.3) and the choice of $\gamma_2$, we have $\sqrt{m}\gamma_2\frac{s\log p}{\sqrt{n}} = o(1)$, $mp^{-3} = o(1)$ and $m\exp(-cn) = o(1)$, which implies

$$\left\|m^{-1/2}\sum_{\ell=1}^{m}\sqrt{n}\theta_K^{-1}\mathbf{M}^{(\ell)}\boldsymbol{D}^{(\ell)}\right\|_\infty = o_P(1).$$

**The Term $m^{-1/2}\sum_{\ell=1}^{m}\sqrt{n}(\mathbf{M}^{(\ell)}\widehat{\boldsymbol{\Sigma}}^{(\ell)} - \mathbf{I})\widehat{\boldsymbol{\delta}}^{(\ell)}$.** As $\|\mathbf{M}^{(\ell)}\widehat{\boldsymbol{\Sigma}}^{(\ell)} - \mathbf{I}\|_{\max} \leq \gamma_2$, we have

$$\|\sqrt{n}(\mathbf{M}^{(\ell)}\widehat{\boldsymbol{\Sigma}}^{(\ell)} - \mathbf{I})\widehat{\boldsymbol{\delta}}^{(\ell)}\|_\infty \leq \sqrt{n}\gamma_1\|\widehat{\boldsymbol{\beta}}^{(\ell)} - \boldsymbol{\beta}^*\|_1.$$

Therefore, by the convergence of $\|\widehat{\boldsymbol{\beta}}^{(\ell)} - \boldsymbol{\beta}^*\|_2$, the empirical sparsity of $\widehat{\boldsymbol{\beta}}$ and Cauchy-Schwarz, it holds that

$$\mathbb{P}\left(\|\sqrt{n}(\mathbf{M}^{(\ell)}\widehat{\boldsymbol{\Sigma}}^{(\ell)} - \mathbf{I})\widehat{\boldsymbol{\delta}}^{(\ell)}\|_\infty > C'_3\gamma_1 s\sqrt{\log p}\right) \leq 4Kp^{-7}+6p^{-3}+3\exp(-cn).$$

By union bound, we have

$$\mathbb{P}\left(\left\|m^{-1/2}\sum_{\ell=1}^{m}\sqrt{n}(\mathbf{M}^{(\ell)}\widehat{\boldsymbol{\Sigma}}^{(\ell)} - \mathbf{I})\widehat{\boldsymbol{\delta}}^{(\ell)}\right\|_\infty > C'_3\gamma_1\sqrt{m}s\sqrt{\log p}\right) \leq m(4Kp^{-7}+6p^{-3}+3\exp(-cn)).$$

By condition (5.3) and the choice of $\gamma_1$, we have $\gamma_1\sqrt{m}s\sqrt{\log p} = o(1)$, $mp^{-3} = o(1)$ and $m\exp(-cn) = o(1)$, which implies

$$\left\|m^{-1/2}\sum_{\ell=1}^{m}\sqrt{n}(\mathbf{M}^{(\ell)}\widehat{\boldsymbol{\Sigma}}^{(\ell)} - \mathbf{I})\widehat{\boldsymbol{\delta}}^{(\ell)}\right\|_\infty = o_P(1).$$

**The term $m^{-1/2}\sum_{\ell=1}^m \sqrt{n}\theta_K^{-1}\mathbf{M}^{(\ell)}\boldsymbol{V}^{(\ell)}$.** By (A.7) and the property of $\widehat{\boldsymbol{\beta}}^{(\ell)}$ and $\widehat{b}_k^{(\ell)}$, we have

$$\mathbb{P}\Big(\big\|\sqrt{n}\mathbf{M}^{(\ell)}\boldsymbol{V}^{(\ell)}\big\|_\infty > CK\gamma_2(s\log(p\vee n))^{3/4}/n^{1/4}\Big) \le 4Kp^{-7} + 7p^{-3} + 3\exp(-cn),$$

Therefore, applying union bound, we have

$$\mathbb{P}\Big(\Big\|m^{-1/2}\sum_{\ell=1}^m \sqrt{n}\theta_K^{-1}\mathbf{M}^{(\ell)}\boldsymbol{V}^{(\ell)}\Big\|_\infty > C\sqrt{m}\gamma_2(s\log(p\vee n))^{3/4}/n^{1/4}\Big)$$
$$\le m(4Kp^{-7} + 7p^{-3} + 3\exp(-cn)).$$

Therefore, by condition (5.3) and the choice of $\gamma_2$, we have $\sqrt{m}\gamma_2(s\log(p\vee n))^{3/4}/n^{1/4} = o(1)$, $mp^{-3} = o(1)$ and $m\exp(-cn) = o(1)$. Hence

$$\Big\|m^{-1/2}\sum_{\ell=1}^m \sqrt{n}\mathbf{M}^{(\ell)}\boldsymbol{V}^{(\ell)}\Big\|_\infty = o_P(1).$$

**The term $m^{-1/2}\sum_{\ell=1}^m \sqrt{n}((\theta_K^{(\ell)})^{-1} - \theta_K^{-1})\mathbf{M}^{(\ell)}\widehat{\boldsymbol{\kappa}}^{(\ell)}$.** For simplicity, we only prove the case when $f_\epsilon$ is known. The case when it is not known can be similarly proved. We have

$$\begin{aligned}|(\theta_K^{(\ell)})^{-1} - \theta_K^{-1}| &= |(\theta_K^{(\ell)})^{-1}\theta_K^{-1}(\theta_K^{(\ell)} - \theta_K)| \le (C_-K)^{-2}|\theta_K^{(\ell)} - \theta_K| \\ &\le C'_+(C_-K)^{-2}\|\widehat{\boldsymbol{b}}^{(\ell)} - \boldsymbol{b}^*\|_1 \le C'_+C_-^{-2}K^{-3/2}\|\widehat{\boldsymbol{b}}^{(\ell)} - \boldsymbol{b}^*\|_2.\end{aligned}$$

Therefore, we have

$$\mathbb{P}\big(|(\theta_K^{(\ell)})^{-1} - \theta_K^{-1}| \ge C'_2K^{-1}\sqrt{s(\log p)/n}\big) \le 4Kp^{-7} + 6p^{-3} + 3\exp(-cn). \tag{D.5}$$

Following the same argument as the proof of Theorem 3.2 and by the property of $\widehat{\boldsymbol{\beta}}$ and $\widehat{b}_k$, we have

$$\mathbb{P}\Big(\|\sqrt{n}\mathbf{M}^{(\ell)}\widehat{\boldsymbol{\kappa}}^{(\ell)}\|_\infty \ge C\gamma_2K(\sqrt{\log p}+s^{3/4}(\log p)^{3/4}/n^{1/4}+\sqrt{s\log p})\Big) \le 4Kp^{-7}+6p^{-3}+3\exp(-cn). \tag{D.6}$$

Combining (D.5) and (D.6) and applying union bound, we have

$$\mathbb{P}\Big(\|m^{-1/2}\sum_{\ell=1}^m \sqrt{n}((\theta_K^{(\ell)})^{-1} - \theta_K^{-1})\mathbf{M}^{(\ell)}\widehat{\boldsymbol{\kappa}}^{(\ell)}\|_\infty \ge C''_2\sqrt{m}\gamma_2 s(\log p)/\sqrt{n}\Big)$$
$$\le m(4Kp^{-7} + 6p^{-3} + 3\exp(-cn)).$$

By condition (5.3) and the choice of $\gamma_2$, we have $\sqrt{m}\gamma_2 s(\log p)/\sqrt{n} = o(1)$, $mp^{-3} = o(1)$ and $m\exp(-cn) = o(1)$. Hence

$$\|m^{-1/2}\sum_{\ell=1}^m \sqrt{n}((\theta_K^{(\ell)})^{-1} - \theta_K^{-1})\mathbf{M}^{(\ell)}\widehat{\boldsymbol{\kappa}}^{(\ell)}\|_\infty = o_P(1).$$

Combing all the above results, we complete the proof of this theorem. □

## D.2 Proof of Theorem 5.4

*Proof.* The proof utilizes the results in Theorem 5.2 and the arguments in the proof of Theorem 3.5 in Javanmard and Montanari (2014). Following that proof, we have for any $\ell = 1, \dots, m$,

$$\mathbb{P}(\widehat{\boldsymbol{\mu}}_j^{(\ell)T}\widehat{\boldsymbol{\Sigma}}^{(\ell)}\widehat{\boldsymbol{\mu}}_j \geq [\boldsymbol{\Sigma}^{-1}]_{j,j} + \varepsilon) \leq 2e^{-cn} + 2p^{-5},$$

for any constant $\varepsilon$ small enough and some constant $c$ that only depends on $\sigma_x, \rho_{\min}$ and $\varepsilon$. Applying union bound, we have

$$\mathbb{P}\left(\frac{1}{m}\sum_{\ell=1}^{m}\widehat{\boldsymbol{\mu}}_j^{(\ell)T}\widehat{\boldsymbol{\Sigma}}^{(\ell)}\widehat{\boldsymbol{\mu}}_j^{(\ell)} \geq [\boldsymbol{\Sigma}^{-1}]_{j,j} + \varepsilon\right) \leq 2e^{\log m - cn} + 2mp^{-5}.$$

When $m$ satisfies the growth bound (5.3), we have by Borel-Cantelli lemma that

$$\limsup_{n\to\infty}\left(\frac{1}{m}\sum_{\ell=1}^{m}\widehat{\boldsymbol{\mu}}_j^{(\ell)T}\widehat{\boldsymbol{\Sigma}}^{(\ell)}\widehat{\boldsymbol{\mu}}_j^{(\ell)} - [\boldsymbol{\Sigma}^{-1}]_{j,j}\right) \leq 0.$$

The rest of the proof then follows similarly as that of Theorem 3.5 in Javanmard and Montanari (2014). □

# E Proof of Technical Lemmas

In this section, we present the detailed proofs of technical lemmas given in Appendices B and C.

## E.1 Proof of Lemma B.1

*Proof.* By Lemma 3.1 in Javanmard and Montanari (2014), we have

$$\widehat{\boldsymbol{\mu}}_j^T\widehat{\boldsymbol{\Sigma}}\widehat{\boldsymbol{\mu}}_j \geq \frac{(1-\gamma_1)^2}{\widehat{\boldsymbol{\Sigma}}_{jj}}.$$

Note that $\widehat{\boldsymbol{\Sigma}}_{jj} = \frac{1}{n}\sum_{i=1}^{n} X_{ij}^2$. By Lemma 5.14 in Vershynin (2012), each $X_{ij}^2$ is $c\sigma_x^2$-sub-exponential, where $c$ is some universal constant. According to Bernstein's inequality, we have

$$\mathbb{P}\left(\frac{1}{n}\sum_{i=1}^{n}X_{ij}^2 - \boldsymbol{\Sigma}_{jj} \geq t\right) \leq \exp\left(-c\min\{nt^2/\sigma_x^4, nt/\sigma_x^2\}\right).$$

Applying union bound, we have

$$\mathbb{P}\left(\max_j\left\{\frac{1}{n}\sum_{i=1}^{n}X_{ij}^2 - \boldsymbol{\Sigma}_{jj}\right\} \geq t\right) \leq p\exp\left(-c\min\{nt^2/(c^2\sigma_x^4), nt/(c\sigma_x^2)\}\right).$$

Letting $t = 2c\sigma_x^2\sqrt{(\log p)/n}$, we have with probability at leat $1-p^{-3}$ that $\widehat{\boldsymbol{\Sigma}}_{jj} < \boldsymbol{\Sigma}_{jj} + 2c\sigma_x^2\sqrt{(\log p)/n}$ for all $j = 1, \dots, p$. Therefore the conclusion follows. □

## E.2 Proof of Lemma B.2

*Proof.* Recall

$$T_{0,\mathcal{G}} = \max_{j\in\mathcal{G}} n^{-1/2}\sum_{i=1}^{n}\Xi_{ij} \quad \text{and} \quad U_{0,\mathcal{G}} = \max_{j\in\mathcal{G}} n^{-1/2}\sum_{i=1}^{n}\Gamma_{i,j},$$

where $\boldsymbol{\Xi}_i := \theta_K^{-1}\Psi_{i,K}\boldsymbol{\Sigma}^{-1}\boldsymbol{X}_i$. Note that $\boldsymbol{\Gamma}_i$ are i.i.d. Gaussian vector, and $\boldsymbol{\Gamma}_i$ and $\boldsymbol{\Xi}_i$ both have zero mean and the same covariance matrix $\mathbb{E}[\boldsymbol{\Gamma}_i\boldsymbol{\Gamma}_i^T] = \theta_K^{-2}\sigma_K^2\boldsymbol{\Sigma}^{-1}$. This allows us to directly apply Corollary 2.1 in Chernozhukov et al. (2013).

It suffices to verify Condition (E1) therein. For any $j$,

$$|\Xi_{ij}| \le \theta_K^{-1}|\Psi_{i,K}|\big|[\boldsymbol{\Sigma}^{-1}\boldsymbol{X}_i]_j\big| \le \theta_K^{-1}K\big|[\boldsymbol{\Sigma}^{-1}\boldsymbol{X}_i]_j\big| \le C_-^{-1}\big|[\boldsymbol{\Sigma}^{-1}\boldsymbol{X}_i]_j\big|,$$

and $[\boldsymbol{\Sigma}^{-1}\boldsymbol{X}_i]_j$ is zero-mean sub-Gaussian with variance proxy $\rho_{\min}^{-2}\sigma_x^2$. Hence, for any $\lambda$, we have

$$\begin{aligned}\mathbb{E}\big[\exp(|\lambda\Xi_{ij}|) &\le C_-^{-1}\mathbb{E}\big[\exp(\lambda[\boldsymbol{\Sigma}^{-1}\boldsymbol{X}_i]_j) + \exp(-\lambda[\boldsymbol{\Sigma}^{-1}\boldsymbol{X}_i]_j)\big]\\ &\le 2C_-^{-1}\mathbb{E}\big[\exp(\lambda[\boldsymbol{\Sigma}^{-1}\boldsymbol{X}_i]_j)\big]\\ &\le C\lambda^2\rho_{\min}^{-2}\sigma_x^2,\end{aligned}$$

for some constant $C$. Moreover, the third and fourth moments of the sub-Gaussian random variable are bounded. Therefore, there exist constants $c_2$, $C_2$ and $B$ such that $c_2 \le \mathbb{E}[\Xi_{ij}^2] \le C_2$, and

$$\max_{k=1,2}\mathbb{E}\big[|\Xi_{ij}|^{2+k}/B^k\big] + \mathbb{E}\big[\exp(|\xi_{ij}|/B)\big] \le 4.$$

Therefore, when $B^2(\log(dn))^7/n \le C_1 n^{-c_1}$, the conclusion in the lemma is true. $\square$

## E.3 Proof of Lemma B.3

*Proof.* By the definition of $\Psi_{i,K}$, for any $j \in \mathcal{G}$, we have

$$\begin{aligned}&\max_{j\in\mathcal{G}}\frac{1}{\sqrt{n}}\left|\sum_{i=1}^{n}\theta_K^{-1}\Psi_{i,K}\widehat{\boldsymbol{\mu}}_j'^T\boldsymbol{X}_i - \sum_{i=1}^{n}\theta_K^{-1}\Psi_{i,K}[\boldsymbol{\Sigma}^{-1}\boldsymbol{X}_i]_j\right|\\ &\quad\le \theta_K^{-1}\max_{j\in\mathcal{G}}\left\|\widehat{\boldsymbol{\mu}}_j' - [\boldsymbol{\Sigma}^{-1}]_{\cdot,j}\right\|_1\sum_{k=1}^{K}\frac{1}{\sqrt{n}}\left\|\sum_{i=1}^{n}(\mathbb{1}\{\epsilon_i \le b_k^*\} - \tau_k)\boldsymbol{X}_i\right\|_\infty,\end{aligned} \tag{E.1}$$

where $[\boldsymbol{\Sigma}^{-1}]_{\cdot,j}$ denotes the $j$-th column of $\boldsymbol{\Sigma}^{-1}$. From proof of Theorem 3.13, we have that $\max_{j\in\mathcal{G}}\|\widehat{\boldsymbol{\mu}}_j' - [\boldsymbol{\Sigma}^{-1}]_{\cdot,j}\|_1 \lesssim s_1\sqrt{(\log p)/n}$ with probability at least $1-4p^{-3}$. Moreover, $|\mathbb{1}\{\epsilon_i \le b_k^*\} - \tau_k|$ is uniformly bounded by 1, thus by sub-Gaussianity of $\boldsymbol{X}_i$ and the union bound, we have for any $t > 0$ that

$$\mathbb{P}\left(\frac{1}{\sqrt{n}}\left\|\sum_{i=1}^{n}(\mathbb{1}\{\epsilon_i \le b_k^*\} - \tau_k)\boldsymbol{X}_i\right\|_\infty > t\right) \le 2p\exp\left(-\frac{t^2}{2\sigma_x^2}\right).$$

Further applying union bound, we have

$$\mathbb{P}\left(\sum_{k=1}^{K}\frac{1}{\sqrt{n}}\left\|\sum_{i=1}^{n}(\mathbb{1}\{\epsilon_i \le b_k^*\} - \tau_k)\boldsymbol{X}_i\right\|_\infty \ge Kt\right) \le 2Kp\exp\left(-\frac{t^2}{2\sigma_x^2}\right).$$

Taking $t = 2\sqrt{2}\sigma_x\sqrt{\log p}$ and by (E.1), we have

$$\mathbb{P}\left(\max_{j\in\mathcal{G}}\frac{1}{\sqrt{n}}\Big|\sum_{i=1}^n \theta_K^{-1}\Psi_{i,K}\widehat{\boldsymbol{\mu}}_j'^T\boldsymbol{X}_i - \sum_{i=1}^n \theta_K^{-1}\Psi_{i,K}[\boldsymbol{\Sigma}^{-1}\boldsymbol{X}_i]_j\Big| \geq Cs_1(\log p)/\sqrt{n}\right) \leq (2K+4)p^{-3}.$$

We take $\zeta_1 = s_1\log p/\sqrt{n}$ and $\zeta_2 = (2K+4)p^{-3}$. Then by the above equation, it holds that

$$\mathbb{P}\left(\max_{j\in\mathcal{G}}\frac{1}{\sqrt{n}}\Big|\sum_{i=1}^n \theta_K^{-1}\Psi_{i,K}\widehat{\boldsymbol{\mu}}_j'^T\boldsymbol{X}_i - \sum_{i=1}^n \theta_K^{-1}\Psi_{i,K}[\boldsymbol{\Sigma}^{-1}\boldsymbol{X}_i]_j\Big| > \zeta_1\right) \leq \zeta_2.$$

By condition in the lemma, we have $s_1\log p\sqrt{\log(d\vee n)}/\sqrt{n} = o(1)$. Hence $\zeta_1\sqrt{1\vee\log(d/\zeta_1)} = o(1)$ and $\zeta_2 = o(1)$. □

## E.4 Proof of Lemma C.1

*Proof.* Recall the Knight's identity (see proof of Theorem 1 in Knight, 1998)

$$|x-y| - |x| = -y\,\mathrm{sign}(x) + 2\int_0^y (I\{x\leq s\} - I\{x\leq 0\})\mathrm{d}s. \tag{E.2}$$

By definition of $\phi_\tau$, we have $\phi_{\tau_k}(x-y) - \phi_\tau(x) = (\tau - \mathbb{1}\{x\leq 0\})y + \int_0^y (I\{x\leq s\} - I\{x\leq 0\})\mathrm{d}s$. By definitions of $\mathcal{H}$ and $\mathcal{L}_K$, we have

$$\begin{aligned}\mathcal{H}(\underset{\sim}{\boldsymbol{\Delta}}) &= \sum_{k=1}^K \mathbb{E}\big[\phi_{\tau_k}(Y - \underset{\sim}{\boldsymbol{X}}^T\underset{\sim}{\boldsymbol{\beta}}_k^* - \underset{\sim}{\boldsymbol{X}}^T\underset{\sim}{\boldsymbol{\Delta}}_k) - \phi_{\tau_k}(Y - \underset{\sim}{\boldsymbol{X}}^T\underset{\sim}{\boldsymbol{\beta}}_k^*)\big] \\ &= \sum_{k=1}^K \mathbb{E}\Big[\big(\tau_k - \mathbb{1}\{Y\leq \underset{\sim}{\boldsymbol{X}}^T\underset{\sim}{\boldsymbol{\beta}}_k^*\}\big)\underset{\sim}{\boldsymbol{X}}^T\underset{\sim}{\boldsymbol{\Delta}}_k \\ &\qquad + \int_0^{\underset{\sim}{\boldsymbol{X}}^T\underset{\sim}{\boldsymbol{\Delta}}_k}\Big(\mathbb{1}\{Y\leq \underset{\sim}{\boldsymbol{X}}^T\underset{\sim}{\boldsymbol{\beta}}_k^* + t\} - \mathbb{1}\{Y\leq \underset{\sim}{\boldsymbol{X}}^T\underset{\sim}{\boldsymbol{\beta}}_k^*\}\Big)dt\Big].\end{aligned}$$

As $\mathbb{E}\big[\tau_k - \mathbb{1}\{Y\leq \underset{\sim}{\boldsymbol{X}}^T\underset{\sim}{\boldsymbol{\beta}}_k^*\}\,|\,\boldsymbol{X}\big] = \tau_k - \mathbb{P}(\epsilon\leq b_k^*\,|\,\boldsymbol{X}) = 0$, and

$$\mathbb{P}\big(Y\leq \underset{\sim}{\boldsymbol{X}}^T\boldsymbol{\beta}_k^* + t\,|\,\boldsymbol{X}\big) = \mathbb{P}\big(\epsilon\leq b_k^* + t\big) = F_\epsilon(b_k^*+t),$$

it follows that

$$\begin{aligned}\mathcal{H}(\underset{\sim}{\boldsymbol{\Delta}}) &= \sum_{k=1}^K \mathbb{E}\Big[\int_0^{\underset{\sim}{\boldsymbol{X}}^T\underset{\sim}{\boldsymbol{\Delta}}_k} F_\epsilon(b_k^*+t) - F_\epsilon(b_k^*)dt\Big] \\ &= \sum_{k=1}^K \mathbb{E}\Big[\int_o^{\underset{\sim}{\boldsymbol{X}}^T\underset{\sim}{\boldsymbol{\Delta}}_k} tf_k^* + \frac{t^2}{2}f_\epsilon'(b_k^*+\widetilde{t})dt\Big] \\ &\geq \sum_{k=1}^K \mathbb{E}\Big[(\underset{\sim}{\boldsymbol{X}}^T\underset{\sim}{\boldsymbol{\Delta}}_k)^2 f_k^*/2 - |\underset{\sim}{\boldsymbol{X}}^T\underset{\sim}{\boldsymbol{\Delta}}_k|^3 C_+'/6\Big] \\ &= \frac{\|\underset{\sim}{\boldsymbol{\Delta}}\|_{\mathbf{S}}^2}{2} - \frac{C_+'}{6}\sum_{k=1}^K \mathbb{E}\big[|\underset{\sim}{\boldsymbol{X}}^T\underset{\sim}{\boldsymbol{\Delta}}_k|^3\big].\end{aligned} \tag{E.3}$$

where the second equality is by mean value theorem. By Assumption 4.1, we have

$$\sum_{k=1}^{K}\mathbb{E}\big[|\underset{\sim}{\boldsymbol{X}}^T\underset{\sim}{\boldsymbol{\Delta}}_k|^3\big] \le m_0\|\underset{\sim}{\boldsymbol{\Delta}}\|_{\mathbf{S}}^3.$$

Define $\eta = 3/(2C'_+m_0)$. We consider the following two cases:

(i) when $\|\underset{\sim}{\boldsymbol{\Delta}}\|_{\mathbf{S}} \le \eta$, we have $m_0\|\underset{\sim}{\boldsymbol{\Delta}}\|_{\mathbf{S}} \le 3/(2C'_+)$, and so

$$\sum_{k=1}^{K}\mathbb{E}\big[|\underset{\sim}{\boldsymbol{X}}^T\underset{\sim}{\boldsymbol{\Delta}}_k|^3\big] \le m_0\|\underset{\sim}{\boldsymbol{\Delta}}\|_{\mathbf{S}}^3 \le \frac{3}{2C'_+}\|\underset{\sim}{\boldsymbol{\Delta}}\|_{\mathbf{S}}^2,$$

which implies

$$\frac{\|\underset{\sim}{\boldsymbol{\Delta}}\|_{\mathbf{S}}^2}{4} \ge \frac{C'_+}{6}\sum_{k=1}^{K}\mathbb{E}\big[|\underset{\sim}{\boldsymbol{X}}^T\underset{\sim}{\boldsymbol{\Delta}}_k|^3\big].$$

Then by (E.3), when $\|\underset{\sim}{\boldsymbol{\Delta}}\|_{\mathbf{S}} \le \eta$, we have $\mathcal{H}(\underset{\sim}{\boldsymbol{\Delta}}) \ge \|\underset{\sim}{\boldsymbol{\Delta}}\|_{\mathbf{S}}^2/4$.

(ii) When $\|\underset{\sim}{\boldsymbol{\Delta}}\|_{\mathbf{S}} > \eta$, observe that $\mathcal{H}$ is convex with respect to $\underset{\sim}{\boldsymbol{\Delta}}$ and that $\mathcal{H}(0) = 0$. Moreover, $\eta/\|\underset{\sim}{\boldsymbol{\Delta}}\|_{\mathbf{S}} \in (0,1)$. Therefore, by convexity it holds that

$$\mathcal{H}\Big(\frac{\eta}{\|\underset{\sim}{\boldsymbol{\Delta}}\|_{\mathbf{S}}}\underset{\sim}{\boldsymbol{\Delta}}\Big) \le \Big(1-\frac{\eta}{\|\underset{\sim}{\boldsymbol{\Delta}}\|_{\mathbf{S}}}\Big)\mathcal{H}(0) + \frac{\eta}{\|\underset{\sim}{\boldsymbol{\Delta}}\|_{\mathbf{S}}}\mathcal{H}(\underset{\sim}{\boldsymbol{\Delta}}) = \frac{\eta}{\|\underset{\sim}{\boldsymbol{\Delta}}\|_{\mathbf{S}}}\mathcal{H}(\underset{\sim}{\boldsymbol{\Delta}}). \tag{E.4}$$

If we let $\underset{\sim}{\boldsymbol{\Delta}}_0 = (\eta/\|\underset{\sim}{\boldsymbol{\Delta}}\|_{\mathbf{S}})\underset{\sim}{\boldsymbol{\Delta}}$, then $\|\underset{\sim}{\boldsymbol{\Delta}}_0\|_{\mathbf{S}} = \eta$. Then by (i), we have

$$\mathcal{H}(\underset{\sim}{\boldsymbol{\Delta}}_0) \ge \frac{\|\underset{\sim}{\boldsymbol{\Delta}}_0\|_{\mathbf{S}}^2}{4} = \frac{\eta^2}{4}. \tag{E.5}$$

By (E.4) and (E.5), we have $\eta\mathcal{H}(\underset{\sim}{\boldsymbol{\Delta}})/\|\underset{\sim}{\boldsymbol{\Delta}}\|_{\mathbf{S}} \ge \frac{\eta^2}{4}$, and so $\mathcal{H}(\underset{\sim}{\boldsymbol{\Delta}}) \ge \eta\|\underset{\sim}{\boldsymbol{\Delta}}\|_{\mathbf{S}}/4$.

Combining (i) and (ii), we get

$$\mathcal{H}(\underset{\sim}{\boldsymbol{\Delta}}) \ge \min\Big\{\frac{\|\underset{\sim}{\boldsymbol{\Delta}}\|_{\mathbf{S}}^2}{4}, \frac{\eta}{4}\|\underset{\sim}{\boldsymbol{\Delta}}\|_{\mathbf{S}}\Big\},$$

which concludes the proof. □

### E.5 Proof of Lemma C.2

*Proof.* By Knight identity, for any $\underset{\sim}{\boldsymbol{\beta}} = (\boldsymbol{\beta}^T, b_1, \ldots, b_k)^T \in \mathbb{R}^{p+K}$, we have

$$\begin{aligned}
\phi_{\tau_k}\big(Y_i - \underset{\sim}{\boldsymbol{X}}_i^T\underset{\sim}{\boldsymbol{\beta}}_k\big) &= \phi_{\tau_k}\big(Y_i - \underset{\sim}{\boldsymbol{X}}_i^T\underset{\sim}{\boldsymbol{\beta}}_k^* - \underset{\sim}{\boldsymbol{X}}_i^T(\underset{\sim}{\boldsymbol{\beta}}_k - \underset{\sim}{\boldsymbol{\beta}}_k^*)\big) \\
&= \phi_{\tau_k}\big(Y_i - \underset{\sim}{\boldsymbol{X}}_i^T\underset{\sim}{\boldsymbol{\beta}}_k^*\big) + \underset{\sim}{\boldsymbol{X}}_i^T(\underset{\sim}{\boldsymbol{\beta}}_k - \underset{\sim}{\boldsymbol{\beta}}_k^*)\big(\tau_k - \mathbb{1}\{Y_i \le \underset{\sim}{\boldsymbol{X}}_i^T\underset{\sim}{\boldsymbol{\beta}}_k^*\}\big) \\
&\quad + \int_0^{\underset{\sim}{\boldsymbol{X}}_i^T(\underset{\sim}{\boldsymbol{\beta}}_k - \underset{\sim}{\boldsymbol{\beta}}_k^*)} \big(\mathbb{1}\{Y_i \le \underset{\sim}{\boldsymbol{X}}_i^T\underset{\sim}{\boldsymbol{\beta}}_k^* + t\} - \mathbb{1}\{Y_i \le \underset{\sim}{\boldsymbol{X}}_i^T\underset{\sim}{\boldsymbol{\beta}}_k^*\}\big)dt.
\end{aligned} \tag{E.6}$$

Let $a_{ik}^* = \tau_k - \mathbb{1}\{\epsilon_i \le b_k^*\} = \tau_k - \mathbb{1}\{Y_i \le \underset{\sim}{\boldsymbol{X}}_i^T \underset{\sim}{\boldsymbol{\beta}}_k^*\}$ and $\boldsymbol{a}_i^* = (a_{i1}^*, \ldots, a_{iK}^*) \in \mathbb{R}^K$. By (E.6), we have

$$\begin{cases} \frac{\partial \phi_{\tau_k}\left(Y_i - \underset{\sim}{\boldsymbol{X}}_i^T \underset{\sim}{\boldsymbol{\beta}}_k\right)}{\partial \boldsymbol{\beta}}\Big|_{\underset{\sim}{\boldsymbol{\beta}}_k = \underset{\sim}{\boldsymbol{\beta}}_k^*} = \boldsymbol{X}_i\big(\tau_k - \mathbb{1}\{\epsilon_i \le b_k^*\}\big) = a_{ik}^* \boldsymbol{X}_i, \\ \frac{\partial \phi_{\tau_k}\left(Y_i - \underset{\sim}{\boldsymbol{X}}_i^T \underset{\sim}{\boldsymbol{\beta}}_k\right)}{\partial b_k}\Big|_{\underset{\sim}{\boldsymbol{\beta}}_k = \underset{\sim}{\boldsymbol{\beta}}_k^*} = \tau_k - \mathbb{1}\{\epsilon \le b_k^*\} = a_{ik}^*. \end{cases}$$

Recall the definition of $\mathcal{L}_{n,k}(\underset{\sim}{\boldsymbol{\beta}}) = \sum_{k=1}^K \frac{1}{n}\sum_{i=1}^n \phi_{\tau_k}\big(Y_i - \underset{\sim}{\boldsymbol{X}}_i^T \underset{\sim}{\boldsymbol{\beta}}_k\big)$. Hence

$$D\mathcal{L}(\underset{\sim}{\boldsymbol{\beta}}^*) := \frac{1}{n}\sum_{i=1}^n \Big(\sum_{k=1}^K a_{ik}^* \boldsymbol{X}_i^T, \boldsymbol{a}_i^{*T}\Big)^T \in \partial \mathcal{L}_{n,k}(\underset{\sim}{\boldsymbol{\beta}}^*).$$

Also, let $D_1\mathcal{L}(\underset{\sim}{\boldsymbol{\beta}}^*) = n^{-1}\sum_{i=1}^n \sum_{k=1}^K a_{ik}^* \boldsymbol{X}_i$ and $D_2\mathcal{L}(\underset{\sim}{\boldsymbol{\beta}}^*) = n^{-1}\sum_{i=1}^n \boldsymbol{a}_i^*$. Then

$$D\mathcal{L}(\underset{\sim}{\boldsymbol{\beta}}^*) = \big(D_1\mathcal{L}(\underset{\sim}{\boldsymbol{\beta}}^*)^T, D_2\mathcal{L}(\underset{\sim}{\boldsymbol{\beta}}^*)^T\big)^T \in \mathbb{R}^{p+K}.$$

In the following, we show that $\|D\mathcal{L}(\underset{\sim}{\boldsymbol{\beta}}^*)\|_\infty$ is bounded by $2K\max\{\widetilde{\sigma}_x, 1\}\sqrt{(\log p)/n}$ with probability tending to 1. Define $\mathcal{E} = \big\{n^{-1}\sum_{i=1}^n X_{ij}^2 \le \widetilde{\sigma}_x^2 \text{ for all } j = 1, \ldots, p\big\}$. By Assumption 4.2, we have $\mathbb{P}(\mathcal{E}^c) \le \delta_n$. Note that $a_{ik}^*$ for $i = 1, \ldots, n$ are zero-mean i.i.d. random variables independent of $\mathbb{X}$. Moreover, $a_{ik}^*$ are uniformly bounded as $\tau_k - 1 \le a_{ik}^* \le \tau$. By Hoeffding's inequality, we have

$$\mathbb{P}\left(\left|\frac{1}{n}\sum_{i=1}^n a_{ik}^*\right| \ge t\right) \le 2\exp(-2nt^2), \tag{E.7}$$

and

$$\mathbb{P}\left(\left|\frac{1}{n}\sum_{i=1}^n a_{ik}^* X_{ij}\right| \ge t \,|\, \mathbb{X}\right) \le 2\exp\left(-\frac{2n^2t^2}{\sum_{i=1}^n X_{ij}^2}\right). \tag{E.8}$$

By (E.7), we apply union bound to get

$$\mathbb{P}\left(\max_{k\in[K]}\left|\frac{1}{n}\sum_{i=1}^n a_{ik}^*\right| \ge t\right) \le 2K\exp(-2nt^2).$$

Letting $t = 2\max\{\widetilde{\sigma}_x, 1\}\sqrt{\frac{\log p}{n}}$, we have

$$\mathbb{P}\left(\max_{k\in[K]}\left|\frac{1}{n}\sum_{i=1}^n a_{ik}^*\right| \ge 2\max\{\widetilde{\sigma}_x, 1\}\sqrt{\frac{\log p}{n}}\right) \le 2Kp^{-8}. \tag{E.9}$$

By (E.8), conditioned on the event $\mathcal{E}$, we have

$$\mathbb{P}\Big(\Big|\frac{1}{n}\sum_{i=1}^n a_{ik}^* X_{ij}\Big| \ge t \,|\, \mathcal{E}\Big) \le 2\exp\Big(-\frac{2nt^2}{\widetilde{\sigma}_x^2}\Big),$$

and so by union bound,

$$\mathbb{P}\Big(\Big\|\sum_{k=1}^K \frac{1}{n}\sum_{i=1}^n a_{ik}^* \boldsymbol{X}_i\Big\|_\infty \ge Kt \,|\, \mathcal{E}\Big) \le 2Kp\exp\Big(-\frac{2nt^2}{\widetilde{\sigma}_x^2}\Big).$$

Let $t = 2\max\{\widetilde{\sigma}_x, 1\}\sqrt{\frac{\log p}{n}}$, then we have

$$\mathbb{P}\Big(\Big\|\sum_{k=1}^{K}\frac{1}{n}\sum_{i=1}^{n}a_{ik}^{*}\boldsymbol{X}_i\Big\|_{\infty} \geq 2K\max\{\widetilde{\sigma}_x, 1\}\sqrt{\frac{\log p}{n}}\,|\,\mathcal{E}\Big) \leq 2Kp^{-7}.$$

Therefore, it follows that

$$\begin{aligned}
&\mathbb{P}\Big(\Big\|\sum_{k=1}^{K}\frac{1}{n}\sum_{i=1}^{n}a_{ik}^{*}\boldsymbol{X}_i\Big\|_{\infty} \geq 2K\max\{\widetilde{\sigma}_x, 1\}\sqrt{\frac{\log p}{n}}\Big)\\
&\qquad \leq \mathbb{P}\Big(\Big\|\sum_{k=1}^{K}\frac{1}{n}\sum_{i=1}^{n}a_{ik}^{*}\boldsymbol{X}_i\Big\|_{\infty} \geq 2K\max\{\widetilde{\sigma}_x, 1\}\sqrt{\frac{\log p}{n}}\,|\,\mathcal{E}\Big) + \mathbb{P}(\mathcal{E}^c)\\
&\qquad \leq 2Kp^{-7} + \delta_n. \qquad\qquad (E.10)
\end{aligned}$$

Combining (E.9) and (E.10), and letting $\lambda \geq 4K\max\{\widetilde{\sigma}_x, 1\}\sqrt{\frac{\log p}{n}}$, we have with probability at least $1 - 4Kp^{-7} - \delta_n$ that

$$\|D_1\mathcal{L}(\underset{\sim}{\boldsymbol{\beta}}^*)\|_{\infty} \leq \frac{\lambda}{2},$$

and

$$\|D_2\mathcal{L}(\underset{\sim}{\boldsymbol{\beta}}^*)\|_{\infty} \leq \frac{\lambda}{2K}.$$

Hence by convexity of the loss function, we have

$$\mathcal{L}_{n,K}(\underset{\sim}{\widehat{\boldsymbol{\beta}}}) - \mathcal{L}_{n,K}(\underset{\sim}{\boldsymbol{\beta}}^*) \geq D\mathcal{L}(\underset{\sim}{\boldsymbol{\beta}}^*)^T(\underset{\sim}{\widehat{\boldsymbol{\beta}}} - \underset{\sim}{\boldsymbol{\beta}}^*) = D_1\mathcal{L}(\underset{\sim}{\boldsymbol{\beta}}^*)^T(\widehat{\boldsymbol{\beta}} - \boldsymbol{\beta}^*) + D_2\mathcal{L}(\underset{\sim}{\boldsymbol{\beta}}^*)^T(\widehat{\boldsymbol{b}} - \boldsymbol{b}^*).$$

By optimality condition, we have with probability at least $1 - 4Kp^{-7} - \delta_n$ that

$$\begin{aligned}
0 &\leq \mathcal{L}_{n,K}(\underset{\sim}{\boldsymbol{\beta}}^*) - \mathcal{L}_{n,K}(\underset{\sim}{\widehat{\boldsymbol{\beta}}}) + \lambda(\|\boldsymbol{\beta}^*\|_1 - \|\widehat{\boldsymbol{\beta}}\|_1)\\
&\leq \|D_1\mathcal{L}(\underset{\sim}{\boldsymbol{\beta}}^*)\|_{\infty}\|\widehat{\boldsymbol{\beta}} - \boldsymbol{\beta}^*\|_1 + \|D_2\mathcal{L}(\underset{\sim}{\boldsymbol{\beta}}^*)\|_{\infty}\|\widehat{\boldsymbol{b}} - \boldsymbol{b}^*\|_1 + \lambda(\|\boldsymbol{\beta}^*\|_1 - \|\widehat{\boldsymbol{\beta}}\|_1)\\
&\leq \frac{\lambda}{2}\|\widehat{\boldsymbol{\beta}} - \boldsymbol{\beta}^*\|_1 + \frac{\lambda}{2K}\|\widehat{\boldsymbol{b}} - \boldsymbol{b}^*\|_1 + \lambda(\|\boldsymbol{\beta}^*\|_1 - \|\widehat{\boldsymbol{\beta}}\|_1).
\end{aligned}$$

Canceling out $\lambda$, we have

$$\frac{1}{2}\|\widehat{\boldsymbol{\beta}} - \boldsymbol{\beta}^*\|_1 + \frac{1}{2K}\|\widehat{\boldsymbol{b}} - \boldsymbol{b}^*\|_1 + (\|\boldsymbol{\beta}^*\|_1 - \|\widehat{\boldsymbol{\beta}}\|_1) \geq 0.$$

By the relation that $\|\widehat{\boldsymbol{\beta}} - \boldsymbol{\beta}^*\|_1 = \|(\widehat{\boldsymbol{\beta}} - \boldsymbol{\beta}^*)_{\mathcal{T}}\|_1 + \|\widehat{\boldsymbol{\beta}}_{\mathcal{T}^c}\|_1$ and

$$\|\boldsymbol{\beta}^*\|_1 - \|\widehat{\boldsymbol{\beta}}\|_1 = \|\boldsymbol{\beta}^*_{\mathcal{T}}\|_1 - \|\widehat{\boldsymbol{\beta}}_{\mathcal{T}}\|_1 - \|\widehat{\boldsymbol{\beta}}_{\mathcal{T}^c}\|_1 \leq \|(\widehat{\boldsymbol{\beta}} - \boldsymbol{\beta}^*)_{\mathcal{T}}\|_1 - \|\widehat{\boldsymbol{\beta}}_{\mathcal{T}^c}\|_1,$$

we get with probability at least $1 - 4Kp^{-7} - \delta_n$ that

$$\|\widehat{\boldsymbol{\beta}}_{\mathcal{T}^c}\|_1 \leq 3\|(\widehat{\boldsymbol{\beta}} - \boldsymbol{\beta}^*)_{\mathcal{T}}\|_1 + \|\widehat{\boldsymbol{b}} - \boldsymbol{b}^*\|_1/K.$$

This concludes the proof. □

### E.6 Proof of Lemma C.3

*Proof.* Define the following quantity

$$
\begin{aligned}
\mathbb{G}(\underset{\sim}{\boldsymbol{\Delta}}) &:= \sqrt{n}\Big\{\widehat{\mathcal{L}}_{n,K}(\underset{\sim}{\boldsymbol{\beta}}^* + \underset{\sim}{\boldsymbol{\Delta}}) - \widehat{\mathcal{L}}_{n,K}(\underset{\sim}{\boldsymbol{\beta}}^*) - \big(\mathcal{L}_K(\underset{\sim}{\boldsymbol{\beta}}^* + \underset{\sim}{\boldsymbol{\Delta}}) - \mathcal{L}_K(\underset{\sim}{\boldsymbol{\beta}}^*)\big)\Big\} \\
&= \frac{1}{\sqrt{n}} \sum_{i=1}^n \sum_{k=1}^K \Big\{\phi_{\tau_k}(Y_i - \underset{\sim}{\boldsymbol{X}}_i^T \underset{\sim}{\boldsymbol{\beta}}_k^* - \underset{\sim}{\boldsymbol{X}}_i^T \underset{\sim}{\boldsymbol{\Delta}}_k) - \phi_{\tau_k}(Y_i - \underset{\sim}{\boldsymbol{X}}_i^T \underset{\sim}{\boldsymbol{\beta}}_k^*) \\
&\qquad\qquad - \mathbb{E}\big[\phi_{\tau_k}(Y - \underset{\sim}{\boldsymbol{X}}^T \underset{\sim}{\boldsymbol{\beta}}_k^* - \underset{\sim}{\boldsymbol{X}}^T \underset{\sim}{\boldsymbol{\Delta}}_k) - \phi_{\tau_k}(Y - \underset{\sim}{\boldsymbol{X}}^T \underset{\sim}{\boldsymbol{\beta}}_k^*)\big]\Big\}.
\end{aligned}
$$

By the definition of $\phi_{\tau_k}$, we have $|\phi_{\tau_k}(x) - \phi_{\tau_k}(y)| \le |x - y|$, hence

$$
\begin{aligned}
\mathrm{Var}[\mathbb{G}(\underset{\sim}{\boldsymbol{\Delta}})] &= \mathrm{Var}\Big[\sum_{k=1}^K \{\phi_{\tau_k}(Y - \underset{\sim}{\boldsymbol{X}}^T \underset{\sim}{\boldsymbol{\beta}}_k^* - \underset{\sim}{\boldsymbol{X}}^T \underset{\sim}{\boldsymbol{\Delta}}_k) - \phi_{\tau_k}(Y - \underset{\sim}{\boldsymbol{X}}^T \underset{\sim}{\boldsymbol{\beta}}_k^*)\}\Big] \\
&\le K \sum_{k=1}^K \mathbb{E}\big[(\underset{\sim}{\boldsymbol{X}}^T \underset{\sim}{\boldsymbol{\Delta}}_k)^2\big] \le K C_-^{-1} \|\underset{\sim}{\boldsymbol{\Delta}}\|_{\mathbf{S}}^2 \le K C_-^{-1} \xi^2.
\end{aligned}
$$

Applying symmetrization method (Lemma 2.3.7 of van der Vaart and Wellner (1996)), we have

$$
\mathbb{P}\Bigg[\sup_{\underset{\sim}{\boldsymbol{\Delta}} \in \mathcal{A}, \|\underset{\sim}{\boldsymbol{\Delta}}\|_{\mathbf{S}} \le \xi} \frac{|\mathbb{G}(\underset{\sim}{\boldsymbol{\Delta}})|}{\sqrt{n}} > t\Bigg] \le \frac{2\mathbb{P}\Big[\sup_{\underset{\sim}{\boldsymbol{\Delta}} \in \mathcal{A}, \|\underset{\sim}{\boldsymbol{\Delta}}\|_{\mathbf{S}} \le \xi} |\mathbb{G}^0(\underset{\sim}{\boldsymbol{\Delta}})|/\sqrt{n} > t/4\Big]}{1 - 4K\xi^2/(C_- n t^2)}, \tag{E.11}
$$

where

$$
\mathbb{G}^0(\underset{\sim}{\boldsymbol{\Delta}}) = \frac{1}{\sqrt{n}} \sum_{i=1}^n \varepsilon_i \sum_{k=1}^K \Big\{\phi_{\tau_k}(Y_i - \underset{\sim}{\boldsymbol{X}}_i^T \underset{\sim}{\boldsymbol{\beta}}_k^* - \underset{\sim}{\boldsymbol{X}}_i^T \underset{\sim}{\boldsymbol{\Delta}}_k) - \phi_{\tau_k}(Y_i - \underset{\sim}{\boldsymbol{X}}_i^T \underset{\sim}{\boldsymbol{\beta}}_k^*)\Big\},
$$

and $\{\varepsilon_i\}_{i=1}^n$ is a sequence of *i.i.d.* Rademacher random variables.

Let $\mathcal{E} = \Big\{\frac{1}{n}\sum_{i=1}^n X_{ij}^2 \le \widetilde{\sigma}_x^2 \text{ for all } j = 1, \dots, p\Big\}$. By Assumption 4.2, we have $\mathbb{P}(\mathcal{E}^c) \le \delta_n$. By Markov's inequality, we have for any $\lambda > 0$,

$$
\mathbb{P}\Big(\sup_{\underset{\sim}{\boldsymbol{\Delta}} \in \mathcal{A}, \|\underset{\sim}{\boldsymbol{\Delta}}\|_{\mathbf{S}} \le \xi} |\mathbb{G}^0(\underset{\sim}{\boldsymbol{\Delta}})| > t \,|\, \mathcal{E}\Big) \le e^{-\lambda t} \mathbb{E}\Big[\exp\Big(\sup_{\underset{\sim}{\boldsymbol{\Delta}} \in \mathcal{A}, \|\underset{\sim}{\boldsymbol{\Delta}}\|_{\mathbf{S}} \le \xi} \lambda |\mathbb{G}^0(\underset{\sim}{\boldsymbol{\Delta}})|\Big) \,|\, \mathcal{E}\Big]. \tag{E.12}
$$

Observe that the function $t \mapsto \phi_{\tau_k}(Y_i - \underset{\sim}{\boldsymbol{X}}_i^T \underset{\sim}{\boldsymbol{\beta}}_k^* + t) - \phi_{\tau_k}(Y_i - \underset{\sim}{\boldsymbol{X}}_i^T \underset{\sim}{\boldsymbol{\beta}}_k^*)$ is 1-Lipschitz. Therefore by the contraction principle (Theorem 4.12 of Ledoux and Talagrand (1991)), we have

$$
\begin{aligned}
&\mathbb{E}\Big[\exp\Big(\sup_{\underset{\sim}{\boldsymbol{\Delta}} \in \mathcal{A}, \|\underset{\sim}{\boldsymbol{\Delta}}\|_{\mathbf{S}} \le \xi} \lambda |\mathbb{G}^0(\underset{\sim}{\boldsymbol{\Delta}})|\Big) \,|\, \mathcal{E}\Big] \\
&\qquad \le \mathbb{E}\Big[\exp\Big(\sup_{\underset{\sim}{\boldsymbol{\Delta}} \in \mathcal{A}, \|\underset{\sim}{\boldsymbol{\Delta}}\|_{\mathbf{S}} \le \xi} 2\lambda \Big|\frac{1}{\sqrt{n}} \sum_{i=1}^n \varepsilon_i \sum_{k=1}^K \underset{\sim}{\boldsymbol{X}}_i^T \underset{\sim}{\boldsymbol{\Delta}}_k\Big|\Big) \,|\, \mathcal{E}\Big] \\
&\qquad \le \mathbb{E}\Big[\exp\Big(\sup_{\underset{\sim}{\boldsymbol{\Delta}} \in \mathcal{A}, \|\underset{\sim}{\boldsymbol{\Delta}}\|_{\mathbf{S}} \le \xi} 2\lambda \Big\|\frac{1}{\sqrt{n}} \sum_{i=1}^n \varepsilon_i \underset{\sim}{\boldsymbol{X}}_i\Big\|_\infty \sum_{k=1}^K \|\underset{\sim}{\boldsymbol{\Delta}}_k\|_1\Big) \,|\, \mathcal{E}\Big].
\end{aligned} \tag{E.13}
$$

As $\xi^2 \geq \|\underset{\sim}{\mathbf{\Delta}}\|_{\mathbf{S}}^2 = \sum_{k=1}^K f_k^*(\mathbf{\Delta}^T\mathbf{\Sigma}\mathbf{\Delta} + \delta_k^2) \geq C_-(K\rho_{\min}\|\mathbf{\Delta}\|_2^2 + \|\boldsymbol{\delta}\|_2^2)$, we have

$$\|\mathbf{\Delta}\|_2 \leq C_-^{-1/2}\rho_{\min}^{-1/2}K^{-1/2}\xi \quad \text{and} \quad \|\boldsymbol{\delta}\|_2 \leq C_-^{-1/2}\xi.$$

Therefore, when $\underset{\sim}{\mathbf{\Delta}} \in \mathcal{A}, \|\underset{\sim}{\mathbf{\Delta}}\|_{\mathbf{S}} \leq \xi$, we have

$$\begin{aligned}\sum_{k=1}^K \|\underset{\sim}{\mathbf{\Delta}}_k\|_1 = K\|\mathbf{\Delta}\|_1 + \|\boldsymbol{\delta}\|_1 &\leq 4K\|\mathbf{\Delta}_{\mathcal{T}}\|_1 + 2\|\boldsymbol{\delta}\|_1 \\ &\leq 4\sqrt{s}K\|\mathbf{\Delta}\|_2 + 2\sqrt{K}\|\boldsymbol{\delta}\|_2 \\ &\leq 4C_-^{-1/2}\rho_{\min}^{-1/2}\sqrt{sK}\xi + 2C_-^{-1/2}\sqrt{K}\xi.\end{aligned}$$

Let $\mathcal{K} := \sqrt{K}C_-^{-1/2}(4\rho_{\min}^{-1/2} + 2)\sqrt{s}$, then $\sup_{\underset{\sim}{\mathbf{\Delta}} \in \mathcal{A}, \|\underset{\sim}{\mathbf{\Delta}}\|_{\mathbf{S}} \leq \xi} \sum_{k=1}^K \|\underset{\sim}{\mathbf{\Delta}}_k\|_1 \leq \mathcal{K}\xi$. It follows from (E.13) that

$$\begin{aligned}\mathbb{E}\Big[\exp\Big(\sup_{\underset{\sim}{\mathbf{\Delta}} \in \mathcal{A}, \|\underset{\sim}{\mathbf{\Delta}}\|_{\mathbf{S}} \leq \xi} \lambda|\mathbb{G}^0(\underset{\sim}{\mathbf{\Delta}})|\Big) \,|\, \mathcal{E}\Big] &\leq \mathbb{E}\Big[\max_{j=[p+1]} \exp\Big(2\lambda\mathcal{K}\xi\Big|\frac{1}{\sqrt{n}}\sum_{i=1}^n \varepsilon_i X_{ij}\Big|\Big) \,|\, \mathcal{E}\Big] \\ &\leq p \max_{j=[p+1]} \mathbb{E}\Big[\exp\Big(2\lambda\mathcal{K}\xi\Big|\frac{1}{\sqrt{n}}\sum_{i=1}^n \varepsilon_i X_{ij}\Big|\Big) \,|\, \mathcal{E}\Big]. \qquad \text{(E.14)}\end{aligned}$$

For any symmetric random variable $Z$ such that $Z \overset{d}{=} -Z$, we have

$$\mathbb{E}[e^{|Z|}] = \mathbb{E}[\max\{e^Z, e^{-Z}\}] \leq \mathbb{E}[e^Z] + \mathbb{E}[e^{-Z}] = 2\mathbb{E}[e^Z].$$

Moreover, conditioned on $\mathbb{X}$, the random variables $\epsilon_i X_{ij}$ are independent and bounded between $[-|X_{ij}|, |X_{ij}|]$. By Hoeffding's Lemma (Lemma 3.6 in van Handel (2014)), we have $\mathbb{E}[e^{\lambda Z}] \leq \exp(\lambda^2(b-a)^2/2)$ for any random variable $Z$ bounded between $[a, b]$. Therefore, for all $j = [p+1]$, we have

$$\begin{aligned}\mathbb{E}\Big[\exp\Big(2\lambda\mathcal{K}\xi\Big|\frac{1}{\sqrt{n}}\sum_{i=1}^n \varepsilon_i X_{ij}\Big|\Big) \,|\, \mathcal{E}, \mathbb{X}\Big] &\leq 2\mathbb{E}\Big[\exp\Big(2\lambda\mathcal{K}\xi\frac{1}{\sqrt{n}}\sum_{i=1}^n \varepsilon_i X_{ij}\Big) \,|\, \mathcal{E}, \mathbb{X}\Big] \\ &\leq 2\mathbb{E}\Big[\exp\Big(4\lambda^2\mathcal{K}^2\xi^2 n^{-1}\sum_{i=1}^n X_{ij}^2\Big) \,|\, \mathcal{E}\Big] \\ &\leq 2\exp\Big(4\lambda^2\mathcal{K}^2\widetilde{\sigma}_x^2\xi^2\Big). \qquad \text{(E.15)}\end{aligned}$$

Hence, combining (E.12), (E.14) and (E.15), we obtain

$$\mathbb{P}\Big(\sup_{\underset{\sim}{\mathbf{\Delta}} \in \mathcal{A}, \|\underset{\sim}{\mathbf{\Delta}}\|_{\mathbf{S}} \leq \xi} |\mathbb{G}^0(\underset{\sim}{\mathbf{\Delta}})| > t \,|\, \mathcal{E}\Big) \leq 2p\exp\Big(-\lambda t + 4\lambda^2\mathcal{K}^2\widetilde{\sigma}_x^2\xi^2\Big)$$

for any $\lambda$. To minimize the R.H.S, we take $\lambda = t/(8\mathcal{K}^2\xi^2\widetilde{\sigma}_x^2)$. Then

$$\mathbb{P}\Big(\sup_{\underset{\sim}{\mathbf{\Delta}} \in \mathcal{A}, \|\underset{\sim}{\mathbf{\Delta}}\|_{\mathbf{S}} \leq \xi} |\mathbb{G}^0(\underset{\sim}{\mathbf{\Delta}})| > t \,|\, \mathcal{E}\Big) \leq 2p\exp\Big(-\frac{t^2}{16\mathcal{K}^2\xi^2\widetilde{\sigma}_x^2}\Big). \qquad \text{(E.16)}$$

Therefore, we have when $t > \max\{\sqrt{6K}\xi(nC_-)^{-1/2}, 32\widetilde{\sigma}_x\mathcal{K}\xi\sqrt{(\log p)/n}\}$,

$$\begin{aligned}\mathbb{P}\Big(\sup_{\underset{\sim}{\boldsymbol{\Delta}}\in\mathcal{A},\|\underset{\sim}{\boldsymbol{\Delta}}\|_{\mathbf{S}}\leq\xi}\big|\mathbb{G}(\underset{\sim}{\boldsymbol{\Delta}})/\sqrt{n}\big| > t\Big) &\leq 3\mathbb{P}\Big(\sup_{\underset{\sim}{\boldsymbol{\Delta}}\in\mathcal{A},\|\underset{\sim}{\boldsymbol{\Delta}}\|_{\mathbf{S}}\leq\xi}|\mathbb{G}^0(\underset{\sim}{\boldsymbol{\Delta}})| > \frac{\sqrt{n}t}{4}\Big)\\ &\leq 3\mathbb{P}\Big(\sup_{\underset{\sim}{\boldsymbol{\Delta}}\in\mathcal{A},\|\underset{\sim}{\boldsymbol{\Delta}}\|_{\mathbf{S}}\leq\xi}|\mathbb{G}^0(\underset{\sim}{\boldsymbol{\Delta}})| > \frac{\sqrt{n}t}{4}\,|\,\mathcal{E}\Big) + 3\mathbb{P}(\mathcal{E}^c)\\ &\leq 6p\exp\Big(-\frac{nt^2}{256\mathcal{K}^2\xi^2\widetilde{\sigma}_x^2}\Big) + 3\mathbb{P}(\mathcal{E}^c)\\ &\leq 6p^{-3} + 3\delta_n,\end{aligned}$$

where the first inequality is by (E.11) and the fact that $t > \sqrt{6K}\xi(nC_-)^{-1/2}$ implies $4K\xi^2/(C_-nt^2) < 2/3$. The penultimate inequality is by (E.16) and the last inequality is by the choice that $t > 32\widetilde{\sigma}_x\mathcal{K}\xi\sqrt{(\log p)/n}$. When $p > 3$, by the definition of $\mathcal{K}$ and the bound for $t$, it suffices to take $t = C_E\sqrt{K}\xi\sqrt{s}\sqrt{(\log p)/n}$, where $C_E$ is as stated in the lemma. □

## E.7 Proof of Lemma C.4

*Proof.* By (C.9), we have

$$\begin{aligned}\widehat{s}\lambda = \operatorname{sign}(\widehat{\boldsymbol{\beta}})^T\operatorname{sign}(\widehat{\boldsymbol{\beta}})\lambda = \operatorname{sign}(\widehat{\boldsymbol{\beta}})^T\frac{1}{n}\sum_{k=1}^K\mathbb{X}^T\widetilde{\boldsymbol{a}}_k\\ &\leq \big(\mathbb{X}\operatorname{sign}(\widehat{\boldsymbol{\beta}})\big)^T\frac{1}{n}\sum_{k=1}^K\widetilde{\boldsymbol{a}}_k\\ &\leq n^{-1}\big\|\mathbb{X}\operatorname{sign}(\widehat{\boldsymbol{\beta}})\big\|_2\Big\|\sum_{k=1}^K\widetilde{\boldsymbol{a}}_k\Big\|_2\\ &\leq K\sqrt{\widehat{s}\psi(\widehat{s})},\end{aligned}$$

where the last inequality is by the fact that $\|\widetilde{\boldsymbol{a}}_k\|_2 \leq \sqrt{n}$ for any $k$ and the definition of $\psi(\cdot)$ as in (4.2). Therefore $\widehat{s} \leq K^2\lambda^{-2}\psi(\widehat{s})$ for any $\lambda > 0$.

Now we choose $\lambda \geq K\sqrt{2\psi(n/\log(p\vee n))\log(p\vee n)/n}$. Let $t_1 = K^2\lambda^{-2}\psi(\widehat{s})$ and $t_2 = n/\log(p\vee n)$. Note that $\widehat{s} \leq t_1$ so $\psi(\widehat{s}) \leq \psi(t_1)$. Suppose $t_1 > t_2$. Then we have

$$t_1 = K^2\lambda^{-2}\psi(\widehat{s}) \leq \frac{n}{2\log(p\vee n)}\frac{\psi(\widehat{s})}{\psi(t_2)} \leq \frac{t_2\psi(t_1)}{2\psi(t_2)} \leq \frac{t_2}{2}\Big\lceil\frac{t_1}{t_2}\Big\rceil < t_1,$$

where the penultimate inequality is by Lemma 13 in Belloni and Chernozhukov (2011). This is a contradiction. Therefore $\widehat{s} \leq t_1 \leq t_2 = n/\log(p\vee n)$. □

## E.8 Proof of Lemma C.6

*Proof.* We have by definition that

$$
\begin{aligned}
\mathcal{E}_2(q,\xi) &= \sup_{\substack{\boldsymbol{\theta}\in\mathbb{S}(q)\\ \underset{\sim}{\boldsymbol{\Delta}}\in R(q,\xi)}} \sum_{k=1}^{K} \Big|\mathbb{E}\Big[\boldsymbol{\theta}^T\boldsymbol{X}\big(\mathbb{1}\{Y \le \underset{\sim}{\boldsymbol{X}}^T(\underset{\sim}{\boldsymbol{\beta}}_k^* + \underset{\sim}{\boldsymbol{\Delta}}_k) + b_k^*\} - \mathbb{1}\{\epsilon \le b_k^*\}\big)\Big]\Big| \\
&\le \sup_{\substack{\boldsymbol{\theta}\in\mathbb{S}(q)\\ \underset{\sim}{\boldsymbol{\Delta}}\in R(q,\xi)}} \sum_{k=1}^{K} \sqrt{\mathbb{E}\big[(\boldsymbol{\theta}^T\boldsymbol{X})^2\big]}\sqrt{\mathbb{E}\big[(\mathbb{1}\{Y \le \underset{\sim}{\boldsymbol{X}}^T(\underset{\sim}{\boldsymbol{\beta}}_k^* + \underset{\sim}{\boldsymbol{\Delta}}_k) + b_k^*\} - \mathbb{1}\{\epsilon \le b_k^*\})^2\big]} \\
&\le \rho_{\max}^{1/2} \sup_{\underset{\sim}{\boldsymbol{\Delta}}\in R(q,\xi)} \sum_{k=1}^{K} \sqrt{\mathbb{E}\big[(\mathbb{1}\{Y \le \underset{\sim}{\boldsymbol{X}}^T(\underset{\sim}{\boldsymbol{\beta}}_k^* + \underset{\sim}{\boldsymbol{\Delta}}_k) + b_k^*\} - \mathbb{1}\{\epsilon \le b_k^*\})^2\big]}. \qquad \text{(E.17)}
\end{aligned}
$$

where in the first inequality we used the Cauchy-Schwarz inequality. We next bound the expectation in (E.17).

$$
\begin{aligned}
&\mathbb{E}\big[(\mathbb{1}\{Y \le \underset{\sim}{\boldsymbol{X}}^T(\underset{\sim}{\boldsymbol{\beta}}_k^* + \underset{\sim}{\boldsymbol{\Delta}}_k) + b_k^*\} - \mathbb{1}\{\epsilon \le b_k^*\})^2 \,|\, \boldsymbol{X}\big] \\
&= \mathbb{E}\big[(\mathbb{1}\{\epsilon \le \underset{\sim}{\boldsymbol{X}}^T\underset{\sim}{\boldsymbol{\Delta}}_k + b_k^*\} - \mathbb{1}\{\epsilon \le b_k^*\})^2 \,|\, \boldsymbol{X}\big] \\
&= F_\epsilon(\underset{\sim}{\boldsymbol{X}}^T\underset{\sim}{\boldsymbol{\Delta}}_k + b_k^*) + F_\epsilon(b_k^*) - 2F_\epsilon(\min\{\underset{\sim}{\boldsymbol{X}}^T\underset{\sim}{\boldsymbol{\Delta}}_k + b_k^*, b_k^*\}) \\
&= F_\epsilon(\underset{\sim}{\boldsymbol{X}}^T\underset{\sim}{\boldsymbol{\Delta}}_k + b_k^*) - F_\epsilon(b_k^*) + 2\big(F_\epsilon(b_k^*) - F_\epsilon(\min\{\underset{\sim}{\boldsymbol{X}}^T\underset{\sim}{\boldsymbol{\Delta}}_k + b_k^*, b_k^*\})\big) \\
&\le 3C_+|\underset{\sim}{\boldsymbol{X}}^T\underset{\sim}{\boldsymbol{\Delta}}_k|. \qquad \text{(E.18)}
\end{aligned}
$$

Therefore, it holds that

$$
\begin{aligned}
&\sup_{\underset{\sim}{\boldsymbol{\Delta}}\in R(q,\xi)} \sum_{k=1}^{K} \sqrt{\mathbb{E}\big[(\mathbb{1}\{Y \le \underset{\sim}{\boldsymbol{X}}^T(\underset{\sim}{\boldsymbol{\beta}}_k^* + \underset{\sim}{\boldsymbol{\Delta}}_k) + b_k^*\} - \mathbb{1}\{\epsilon \le b_k^*\})^2\big]} \\
&\le \sup_{\underset{\sim}{\boldsymbol{\Delta}}\in R(q,\xi)} \sum_{k=1}^{K} \sqrt{3C_+\mathbb{E}[|\underset{\sim}{\boldsymbol{X}}^T\underset{\sim}{\boldsymbol{\Delta}}_k|]} \\
&\le \sup_{\underset{\sim}{\boldsymbol{\Delta}}\in R(q,\xi)} \sum_{k=1}^{K} \sqrt{3C_+}\sqrt{\mathbb{E}[|\underset{\sim}{\boldsymbol{X}}^T\underset{\sim}{\boldsymbol{\Delta}}_k|^2]}. \qquad \text{(E.19)}
\end{aligned}
$$

By Jensen's inequality, we have $\xi^2 = \sum_{k=1}^K f_k^*\mathbb{E}\big[(\underset{\sim}{\boldsymbol{X}}^T\underset{\sim}{\boldsymbol{\Delta}}_k)^2\big] \ge C_-\Big(\sum_{k=1}^K \sqrt{\mathbb{E}\big[(\underset{\sim}{\boldsymbol{X}}^T\underset{\sim}{\boldsymbol{\Delta}}_k)^2\big]}\Big)^2/K$. Therefore, (E.19) implies that

$$
\sup_{\underset{\sim}{\boldsymbol{\Delta}}\in R(q,\xi)} \sum_{k=1}^{K} \sqrt{\mathbb{E}\big[(\mathbb{1}\{Y \le \underset{\sim}{\boldsymbol{X}}^T(\underset{\sim}{\boldsymbol{\beta}}_k^* + \underset{\sim}{\boldsymbol{\Delta}}_k) + b_k^*\} - \mathbb{1}\{\epsilon \le b_k^*\})^2\big]} \le \sqrt{3KC_+/C_-}\,\xi,
$$

which concludes the proof. □

# F Auxiliary Lemmas

**Lemma F.1.** For any matrix $\mathbf{A} \in \mathbb{R}^{q\times p}$ and $\boldsymbol{v} \in \mathbb{R}^p$, we have

$$\|\mathbf{A}\boldsymbol{v}\|_2 \le \|\mathbf{A}\|_{1,1}\|\boldsymbol{v}\|_\infty.$$

*Proof.* Let $\mathbf{A} = [\boldsymbol{a}_1, \dots, \boldsymbol{a}_q]^T$, where $\boldsymbol{a}_i \in \mathbb{R}^p$ for all $j = 1, \dots, q$.

$$\begin{aligned}\|\mathbf{A}\boldsymbol{v}\|_2^2 = \sum_{j=1}^q (\boldsymbol{a}_j^T\boldsymbol{v})^2 &\le \sum_{j=1}^q \|\boldsymbol{a}_j\|_1^2\|\boldsymbol{v}\|_\infty^2 \\ &\le \Big(\sum_{j=1}^q \|\boldsymbol{a}_j\|_1\Big)^2\|\boldsymbol{v}\|_\infty^2 = \|\mathbf{A}\|_{1,1}^2\|\boldsymbol{v}\|_\infty^2,\end{aligned}$$

as desired. □

**Lemma F.2.** For any matrix $\mathbf{Q} \in \mathbb{R}^{q\times p}$ and symmetric positive semidefinite matrix $\mathbf{A} \in \mathbb{R}^{p\times p}$, we have

$$\|\mathbf{Q}\mathbf{A}\mathbf{Q}^T\|_2 \ge \min_{1\le j\le p} \mathbf{A}_{jj} S_{\min}^2(\mathbf{Q}),$$

where $S_{\min}(\mathbf{Q})$ is the minimum singular value of $\mathbf{Q}$.

*Proof.* We have

$$\|\mathbf{Q}\mathbf{A}\mathbf{Q}^T\|_2 = \max_{\boldsymbol{x}\in\mathbb{R}^q} \frac{\boldsymbol{x}^T\mathbf{Q}\mathbf{A}\mathbf{Q}^T\boldsymbol{x}}{\|\boldsymbol{x}\|_2^2}.$$

For any $\boldsymbol{x} \in \mathbb{R}^q$, let $\boldsymbol{y} = \mathbf{Q}^T\boldsymbol{x}$. Then $\|\boldsymbol{y}\|_2^2 = \boldsymbol{x}^T\mathbf{Q}\mathbf{Q}^T\boldsymbol{x}^T \ge \|\boldsymbol{x}\|_2^2 S_{\min}^2(\mathbf{Q})$. Therefore,

$$\|\mathbf{Q}\mathbf{A}\mathbf{Q}^T\|_2 \ge \max_{\boldsymbol{y}\in\mathrm{Range}(\mathbf{Q}^T)} \frac{S_{\min}^2(\mathbf{Q})\boldsymbol{y}^T\mathbf{A}\boldsymbol{y}}{\|\boldsymbol{y}\|_2^2}.$$

$\mathrm{Range}(\mathbf{Q}^T)$ contains the vector $e_j \in \mathbb{R}^p$ for some $1 \le j \le p$ (otherwise $\mathrm{Range}(\mathbf{Q}^T) = \{\mathbf{0}\}$). Therefore,

$$\|\mathbf{Q}\mathbf{A}\mathbf{Q}^T\|_2 \ge S_{\min}^2(\mathbf{Q}) \min_{1\le j\le p} e_j^T\mathbf{A}e_j = S_{\min}^2(\mathbf{Q}) \min_{1\le j\le p} \mathbf{A}_{jj},$$

which concludes the proof. □